\documentclass{article}

\usepackage{arxiv}

\usepackage[utf8]{inputenc} 
\usepackage[T1]{fontenc}    
\usepackage{hyperref}       
\usepackage{url}            
\usepackage{booktabs}       
\usepackage{amsfonts}       
\usepackage{nicefrac}       
\usepackage{microtype}      
\usepackage{lipsum}
\usepackage{color}
\usepackage{times}
\usepackage{epsfig}
\usepackage{graphicx}
\usepackage{amsmath}
\usepackage{amssymb}
\usepackage{subfig}
\usepackage{array}
\usepackage{textcomp}
\usepackage{pdfpages}
\title{SDF-MAN: Semi-supervised Disparity Fusion with Multi-scale Adversarial Networks}

\author{
  Can Pu \\
   \And
 Runzi Song \\
   \And
Radim Tylecek \\
   \And
Nanbo Li \\ 
   \And
Robert B. Fisher \\
}

\begin{document}
\maketitle

\begin{abstract}
 Refining raw disparity maps from different algorithms to exploit their complementary advantages is still challenging. Uncertainty~estimation and complex disparity relationships among pixels limit the accuracy and robustness of existing methods and there is no standard method for fusion of different kinds of depth data. In~this paper, we~introduce a new method to fuse disparity maps from different sources, while incorporating supplementary information (intensity,~gradient,~etc.) into a refiner network to better refine raw disparity inputs. A~discriminator network classifies disparities at different receptive fields and scales. Assuming~a Markov Random Field for the refined disparity map produces better estimates of the true disparity distribution. Both~fully supervised and semi-supervised versions of the algorithm are proposed. The~approach includes a more robust loss function to inpaint invalid disparity values and requires much less labeled data to train in the semi-supervised learning mode. The~algorithm can be generalized to fuse depths from different kinds of depth sources. Experiments~explored different fusion opportunities: stereo-monocular fusion, stereo-ToF fusion and stereo-stereo fusion. The~experiments show the superiority of the proposed algorithm compared with the most recent algorithms on public synthetic datasets (Scene~Flow, SYNTH3, our~synthetic garden dataset ) and real datasets (Kitti2015 dataset and Trimbot2020 \mbox{Garden dataset}).
\end{abstract}


\section{Introduction \label{s1}}
With recent improvements in depth sensing devices, depth information is now easily accessible ({In a stereo camera pair depth and disparity are interchangeable measures: \textit{depth = focal \_ length $\times$ baseline / disparity}. When~data is from a sensor like time of flight sensor, the~depths can be converted into disparities using a constant focal length and baseline}). 
However,~each sensor has its own advantages and disadvantages, with the result that no algorithm can perform accurately and robustly in all general scenes. For~example, active illumination devices such as ToF (Time of Flight) sensors and structured light cameras~\cite{Principle_ToF_structure_light_devices} estimate the depth information accurately regardless of the scene content but struggle on low reflective surfaces or outdoors. Stereo~vision algorithms~\cite{PSMNet,FPGA_stereo,Dispnet,SGM_Luis,SGM} 
work better outdoors and perform accurately on high texture areas but behave poorly in repetitive or textureless regions. Monocular~vision algorithms~\cite{Monodepth} work robustly in textureless areas but tend to produce blurry depth edges. Thus,~fusing multiple depth maps from different kinds of algorithms or devices and utilizing their complementary strengths to get more accurate depth information is a valuable technique for various applications. 

The traditional pipeline for the majority of the fusion algorithms~\cite{Stereo-ToF_Survey,Probablistic_Fusion,Reliable_Fusion,DL_Fusion,SM_Fusion} is: (1) estimate disparities from the different sensors, (2) estimate associated confidence maps, and~(3) apply a specific fusion algorithm based on the confidence maps to get a refined disparity map. This~approach has three potential problems. Primarily,~estimating the confidence maps for different sensors is a hard task with limited robustness and accuracy. Second,~estimating the disparity relationship among pixels in a general scene is hard without prior knowledge. Finally,~there is no common methodology for different kinds of depth fusion, such as stereo-stereo fusion, monocular-stereo fusion and stereo-ToF fusion. Thus,~researchers have designed different methods for different fusion tasks. The~recent fusion method~\cite{Deep_stere_fusion} based on end to end deep learning has provided a general solution to different kinds of fusion but has limited accuracy and robustness, in~part due to 
not exploiting other associated information to help the network make judgments. It~also did not exploit the disparity relationship among pixels.

In this paper, an~architecture similar to a Generative Adversarial Network (GAN)~\cite{Standard_GAN} (generator is replaced by a refiner network without random noise input) is proposed to solve the three problems listed above, by~designing an efficient network structure and a robust object function. In~addition to the raw disparity maps the network input also includes other image information, i.e., the original intensity and gradient images (see Figure~\ref{fig:Whole_diagram}), in~order to facilitate the selection of a more accurate disparity value from the input disparity images. This~avoids having to design a manual confidence measure for different sensors and allows a common methodology for different kinds of sensor.
To~preserve and exploit the local information better, some successful ideas about local structure from Unet~\cite{Unet} and Densenet~\cite{Densenet} have been used. To~help the network refine the disparity maps accurately and robustly a novel objective function was designed. Gradient~information is incorporated as a weight into the $L_1$ distance to force the disparity values at the edges to get closer to the ground truth. A~smoothness term helps the network propagate the accurate disparity values at edges to adjacent areas, which inpaints regions with invalid disparity values. The~Wasserstein distance~\cite{WGAN,Improved-WGAN} replaced the Jensen-Shannon divergence~\cite{Standard_GAN} for GAN loss to reduce training difficulties and avoid mode collapse. With~the discriminator network classifying input samples in different receptive fields and scales, the~disparity Markov Random Field in the refined disparity map gives a better estimate of the real distribution.

Our semi-supervised approach trains the discriminator network to produce the refined disparity map using not only the labeled data but also the unlabeled data along with the ground truth of the labeled data. It~requires less labeled training data but still achieves accuracy similar to the proposed fully-supervised method or better performance when using the same amount of labeled data with additional unlabeled data compared with the supervised method, as~shown in the experimental results. 

Section~\ref{s2} reviews key previous disparity fusion algorithms and also recent advances in GAN networks. Section~\ref{s3} presents the new fusion model including the objective function and network structure. Section~\ref{s4} presents the results of experiments conducted with synthetic and real data (Table~\ref{Dataset}) for stereo-monocular fusion, stereo-ToF fusion and stereo-stereo fusion. 

Contributions: 
We have:
\begin{enumerate}
	\item Improved fusion accuracy by using a network that learns the disparity relationships among pixels without any prior knowledge.
	\item Reduced the labeled data requirement drastically by using the proposed semi-supervised strategy.
	\item Increased robustness by fusing intensity and gradient information as well as depth data.
	\item Proposed a common network methodology allowing different kinds of sensor fusion without requiring detailed knowledge of the performance of each sensor.
\end{enumerate}

\section{Related Work \label{s2}}
The approach of fusing depth maps from different sensors (e.g., stereo-ToF depth fusion) has become popular. 
The majority of the fusion work~\cite{Probablistic_Fusion,Reliable_Fusion,DL_Fusion,SM_Fusion} shares the same pipeline architecture, which estimates the uncertainty of each pixel first and then refines the depth map based on those confidence maps. 
A recent survey is in~\cite{Stereo-ToF_Survey}. More~recently, Dal Mutto et al.~\cite{Probablistic_Fusion} used the IR frequency, etc., of a ToF sensor to estimate the depth map uncertainty and used the similarity of image patches in the stereo images to estimate the confidence of pixels in the stereo depth map. Then~a MAP-MRF framework refined the depth map. Later,~Marin et al.~\cite{Reliable_Fusion} also utilized sensor physical properties to estimate the confidence for the ToF depth map and used an empirical model based on the global and local cost of stereo matching to calculate the confidence map for the stereo vision sensor. The~extended LC (Locally Consistent) technique was used to fuse the depth maps based on each confidence map. To~get a more accurate confidence map for fusion, Agresti et al.~\cite{DL_Fusion} used a simple convolution neural network for uncertainty estimation and then used the LC technique from~\cite{Reliable_Fusion} for the fusion. In~addition to the work in stereo-ToF fusion above, Facil et al.~\cite{SM_Fusion} used a weighted interpolation of depths from a monocular vision sensor and a multi-view vision sensor based on the likelihood of each pixel's contribution to the depth value. 

The above-mentioned approaches have two issues limiting the accuracy of the refined disparity map: (1) Estimating the confidence map for each type of sensor accurately is hard and makes the system unstable. (2) Accurately modeling the complex disparity relationship among neighboring pixels in random scenes is challenging.

The other class of depth fusion methods is based on deep learning. The~method proposed here belongs to this class and we believe that it is the first to solve the two critical problems above simultaneously. Some~researchers~\cite{DL_Fusion,Patch_based_stereo_confidence} have estimated the confidence maps for different sensors with deep learning methods and then incorporated the confidence as weights into the classical pipeline to refine the disparity map. However,~these methods treat the confidence maps as an intermediate result and no one has trained the neural network to do the fusion from end to end directly and taken both the depth and confidence information into account simultaneously. For~example, Poggi and Mattoccia~\cite{Deep_stere_fusion} selected the best disparity value for each pixel from the several algorithms by formulating depth fusion as a multi-labeling deep network classification problem. However,~the~method only used the disparity maps from the sensors and neglected other associated image information (e.g., intensities, gradients). The~approach did not exploit the real disparity relationship among neighbouring pixels.

The recent development of the GAN methodology led to the foundation of the approach proposed here. The~GAN was first proposed by Goodfellow et al.~\cite{Standard_GAN}, who~trained two neural networks (generator and discriminator) simultaneously to make the distribution of the output from the generator approximate the real data distribution by a minimax two-player strategy. To~control the data generation process, Mirza and Osindero~\cite{Condition_GAN} conditioned the model on additional information. There~are many variants based on the initial GAN model as seen in the survey~\cite{Survey_GAN}. Some~researchers~\cite{WGAN,Improved-WGAN} used the Wasserstein distance to measure the distance between the model distribution and the real distribution, which reduced the difficulty of training the GAN drastically. It~also reduced mode collapse to some extent. 
GANs have been applied to problems other than disparity fusion. 
For example, Isola et al.~\cite{Pix2pix} trained a GAN to translate between image domains which can be also used to transfer the initial disparity maps from several sensors into a refined disparity map. 
However, the~design proposed in~\cite{Pix2pix} neglects information useful for disparity fusion, which limits the accuracy of the refined disparity map. 

In summary, there are previously developed methods for depth fusion based on both the algorithmic pipeline and emerging deep network techniques. 
In this paper, we~combine image evidence as well as raw depth to give a more robust objective function. This~is implemented in an end-to-end architecture similar to a GAN.
We are the first to our knowledge to use such structure to learn the complex disparity relationship among pixels to improve depth fusion accuracy.

\section{Methodology \label{s3}}
First we introduce the proposed general framework (Figure~\ref{fig:Whole_diagram}) for disparity fusion and then the new loss functions in the supervised and semi-supervised methods. These~functions will make adversarial training simple and the refined disparity more accurate and robust. Finally,~the~end-to-end refiner (Figure~\ref{fig:Network_generator}) and discriminator (Figure~\ref{fig:Network_discriminator}) network structure are presented.

\subsection{Framework}
\begin{figure}[h!]
	\centering
	\subfloat[]{\includegraphics[width = 7cm,height=5.5cm]{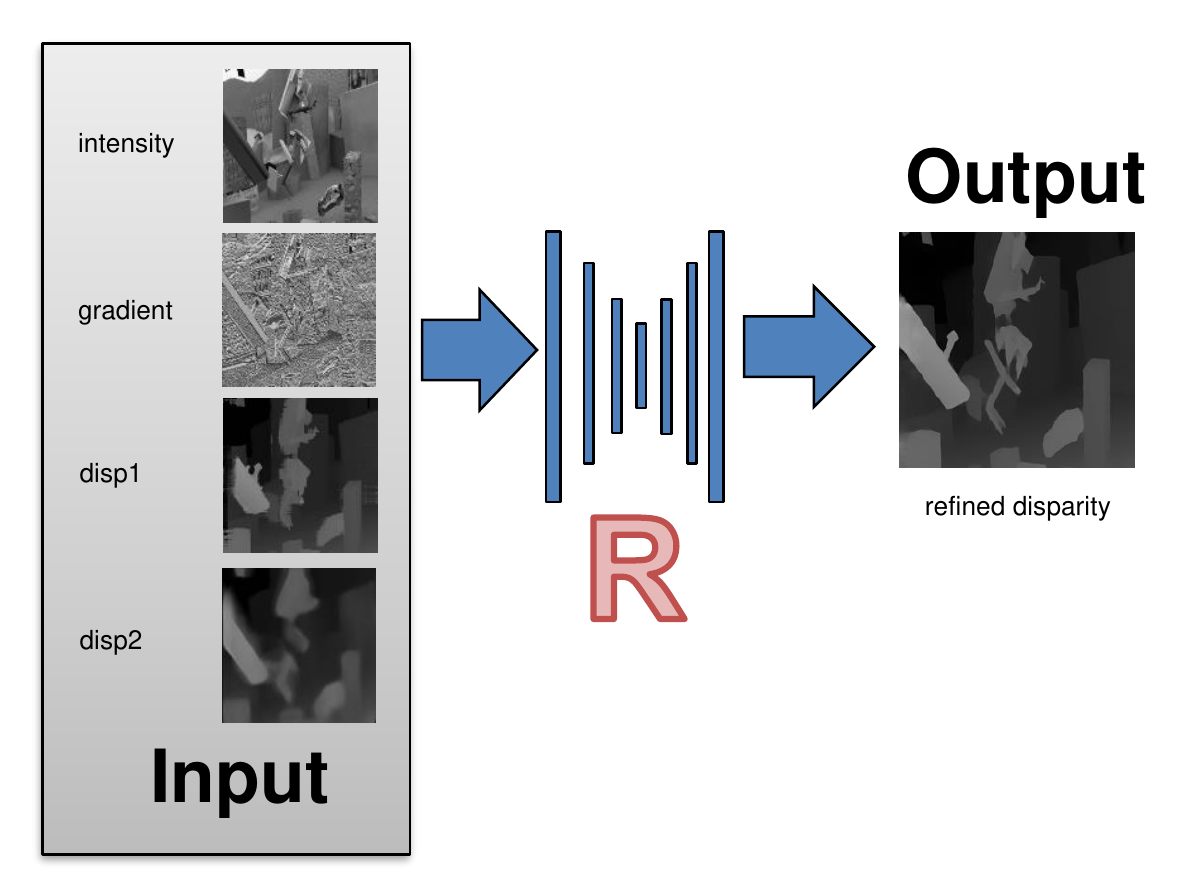}}
	
	\subfloat[]{\includegraphics[width = 7cm,height=5cm]{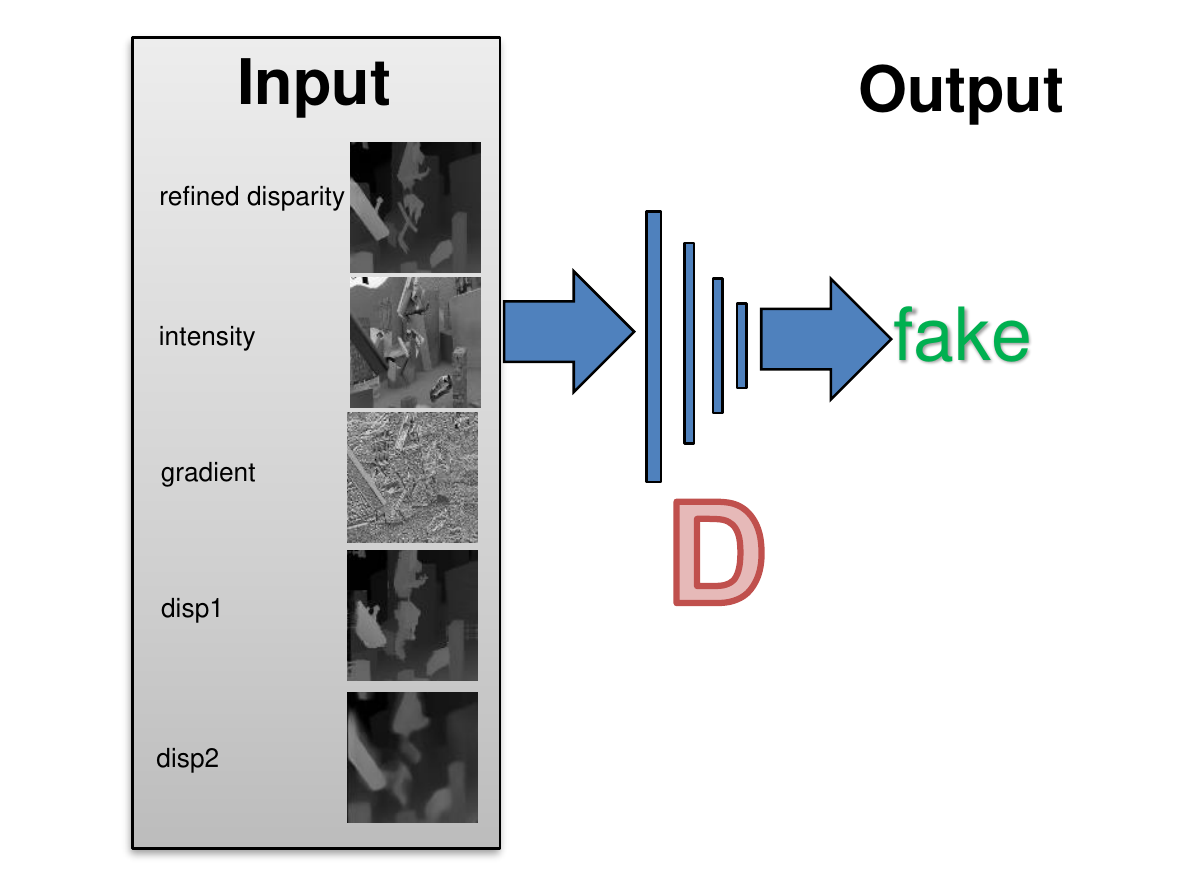}} 
	\hspace{1cm}
	\subfloat[]{\includegraphics[width = 7cm,height=5cm]{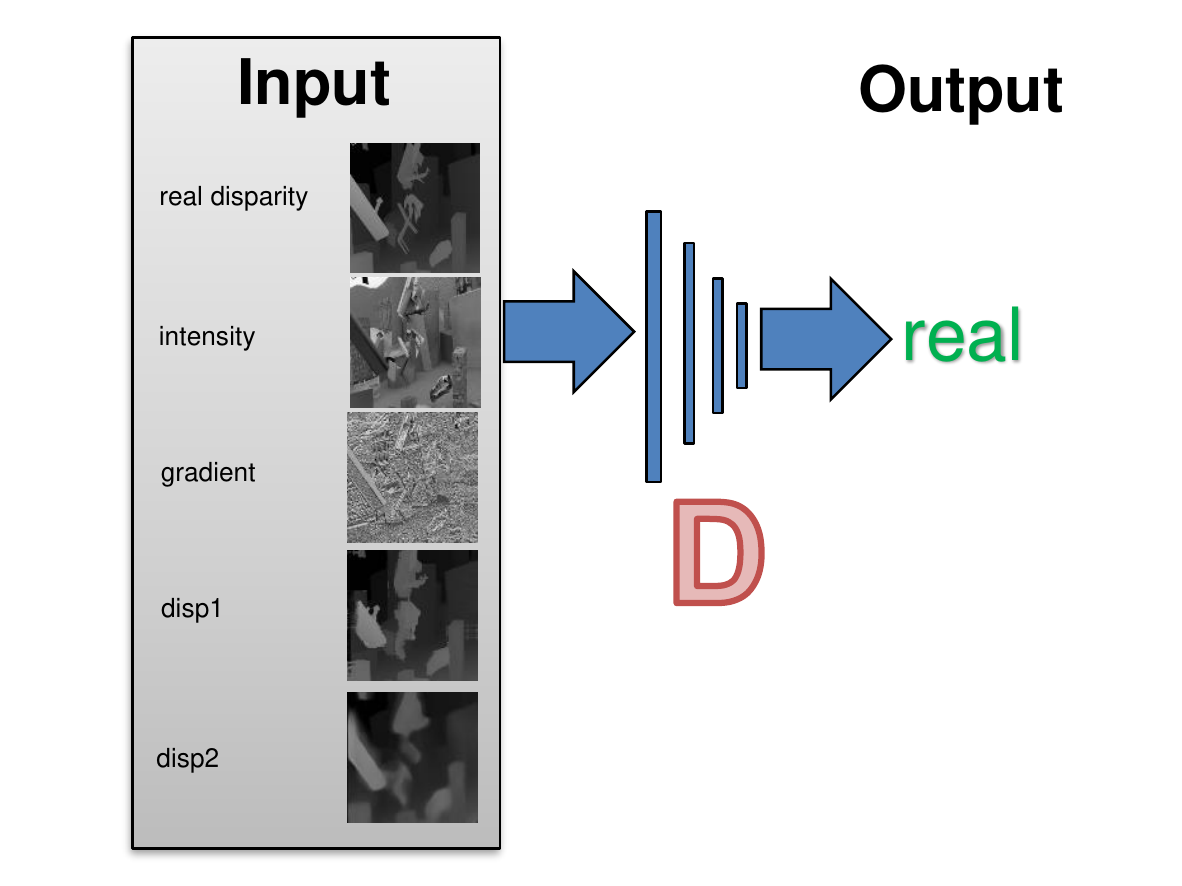}} 
	\caption{Overview 
		of Sdf-MAN. We~train a refiner network~\textit{R} to map raw disparity maps (\textit{disp1}, \textit{disp2}) from two input algorithms to the ground truth based on associated image information (gradient, intensity). The~refiner~\textit{R} tries to predict a refined disparity map close to the ground truth. The~discriminator \textit{D} tries to discriminate whether its input is fake (\textcolor{black}{refined disparity} from~\textit{R}) or real (\textcolor{black}{real disparity} from the ground truth). The~refiner and discriminator can see both the supplementary information and initial disparity inputs simultaneously. We~can fuse any number of disparity inputs or different information cues by concatenating them together directly as inputs. The~two networks are updated alternately. (\textbf{a})~Refiner: a network to refine initial disparity maps; (\textbf{b})~Negative examples: a discriminator network with refined disparity inputs; (\textbf{c})~Positive examples: a discriminator network with real disparity inputs.} 
	\label{fig:Whole_diagram}
\end{figure}

We develop a method that uses an adversarial network, which is similar to a GAN~\cite{Standard_GAN} but with raw disparity maps, gradient and intensity information as inputs instead of random noise. The~refiner network $R$ (similar to the generator $G$ in~\cite{Standard_GAN}) is trained to produce a refined disparity which cannot be classified as ``fake'' by the discriminator $D$. Simultaneously,~the~discriminator $D$ is trained to become better at distinguishing that the input from refiner $R$ is fake and the input from the ground truth is real. By~adopting a minimax two-player game strategy, the~two neural networks \{$R$, $D$\} make the output distribution from the refiner network approximate the real data distribution. The~full system diagram is shown in Figure~\ref{fig:Whole_diagram}.

\unskip

\subsection{Objective Function \label{objective_function} }
To let the refiner produce a more accurate refined disparity map, the~objective function is designed as follows:

(1) To encourage the disparity value of each pixel to approximate the ground truth and to avoid blur at scene edges (such as occurs with the Monodepth method~\cite{Monodepth}), a~gradient-based $L_1$ distance training loss is used, which applies a larger weight to the disparity values at the scene edges:
\begin{equation}
\label{eq1}
\mathcal{L}_{L_1}(R) = \mathop{{}\mathbb{E}}_{x \sim P_{real}, \tilde{x} \sim P_{refiner} }
\Big[\exp(\alpha |\nabla (I_l)|) \; || x-\tilde{x}||_1\Big]
\end{equation} 
where $R$ 
represents the refiner network. $x$ is the ground truth and $\tilde{x}$ is the refined disparity map from the refiner. $P_{real}$ and $P_{refiner}$ represents the real disparity distribution from the ground truth and fake disparity distribution produced by the refiner. $\nabla (I_l)$ is the gradient of the left intensity image in the scene because all the inputs and refined disparity map are from the left view. $\alpha \geq 0$ weights the gradient. $||\bullet ||_1$ is the $L_1$ distance. The~goal is to encourage disparity estimates near image edges (larger gradients) to get closer to the ground truth.

(2) 
A gradient-based smoothness term is added to propagate more reliable disparity values from image edges to the other areas in the image under the assumption that the disparity of neighboring pixels should be similar if their intensities are similar:
\begin{equation}
\label{eq2}
\mathcal{L}_{sm}(R) = \mathop{{}\mathbb{E}}_{u \in \tilde{x}, v~\in N(u), \tilde{x} \sim P_{refiner} }
\Big[\exp(1-\beta | \nabla (I_l)_{uv}|) \; || \tilde{x}_u-\tilde{x}_v||_1 \Big]
\end{equation} 
where $\tilde{x}_u$ is the disparity value of a pixel $u$ in the refined disparity map $\tilde{x}$ from the refiner. $\tilde{x}_v$ is the disparity value of a pixel $v$ in the neighborhood $N(u)$ of pixel $u$. $\nabla (I_l)_{uv}$ is the gradient from pixel $u$ to $v$ in the left intensity image (the refined disparity map is produced on the left view). $\beta \geq 0$ is responsible for how close the disparities are if the intensities in the neighbourhood are similar.

(3) The underlying assumption in $\mathcal{L}_{L_1}(R)$ is that the disparity relationship among pixels is independent.
The disparity relationship in $\mathcal{L}_{sm}(R)$ is too simple to describe the real disparity relationship among neighbours in the real situation. 
To help the refiner produce a disparity map whose disparity Markov Random Field is closer to the real distribution, the~proposed method inputs disparity maps from the refiner and the ground truth into the discriminator, which outputs the probability of the input samples being from the same distribution as the ground truth. This~probability is then used to update the refiner through its loss function. Instead~of defining a global discriminator to classify the whole disparity map, we~define it to classify all local disparity patches separately because any local disparity patch sampled from the refined disparity map should have similar statistics to the real disparity patch. Thus,~by~making the discriminator output the probabilities in different receptive fields or scales (In Figure~\ref{fig:Network_discriminator}, 
please refer to $D1$, $D2$, ..., $D5$), the~refiner will be forced to make the disparity distribution in the refined disparity map close to the real. 
In Equations~\eqref{eq3} and~\eqref{eq4} below, $D_i$ is the probability at the $i${th} scale that the input patch to the discriminator is from the real distribution at the $i${th} scale:
\begin{equation}
\label{eq3}
\mathcal{L}_{JS-GAN}(R,D_i) = \mathop{{}\mathbb{E}}_{x \sim P_{real}}\Big[\log(D_{i}(x))\Big] + 
\mathop{{}\mathbb{E}}_{\tilde{x} \sim P_{refiner} }\Big[\log(1-D_{i}(\tilde{x}))\Big]
\end{equation} 

To avoid $JS-GAN$  
mode collapse during training and alleviate other training difficulties, we~have also investigated replacing $\mathcal{L}_{JS-GAN}(R,D_i)$ with the improved WGAN loss function~\cite{Improved-WGAN}. $\lambda$ is the penalty coefficient (We set it 0.0001 for all the experiments in this paper) and $\hat{x}$ are the random samples (For more details, please read~\cite{Improved-WGAN}):

\begin{equation}
\label{eq4}
\mathcal{L}_{WGAN}(R,D_{i}) = \mathop{{}\mathbb{E}}_{\tilde{x} \sim P_{refiner} }\Big[D_{i}(\tilde{x})\Big]-
\mathop{{}\mathbb{E}}_{x \sim P_{real}}\Big[D_{i}(x)\Big]+
\lambda \mathop{{}\mathbb{E}}_{\hat{x} \sim P_{\hat{x}} }\Big[(||\nabla _{\hat{x}} D_{i}(\hat{x}) ||_2 - 1)^2\Big] 
\end{equation}

The experiments explored the difference in performance of these two GAN loss functions. We~let $\mathcal{L}_{GAN}(R,D_i)$ be either $\mathcal{L}_{JS-GAN}(R,D_i)$ or $\mathcal{L}_{WGAN}(R,D_{i})$ in the following context. The~difference of performance with both the single scale and multiple scales will also be explored.

(4) By inputting only the refined disparity map and its corresponding ground truth into the discriminator simultaneously in each step during training, the~discriminator is trained in a fully supervised manner considering whether the input disparity maps are the same.
In semi-supervised mode, we~still feed the refined disparity map and its corresponding ground truth into the discriminator for the labeled data. But~for the unlabeled data, we~feed the refined disparity map of the unlabeled data and random samples from a small ground truth dataset simultaneously. By~doing this, the~discriminator will be taught to classify the input samples based on the disparity Markov Random Field. Then,~in~turn, the~refiner will be trained to produce a disparity Markov Random Field in the refined disparity map that is closer to the real case.

(5) The combined loss function in the fully supervised learning approach is:
\begin{equation}
\label{eq5}
\mathcal{L}(R,D) = \theta _{1}\mathcal{L}_{L_1}^{Ld}(R) + \theta _{2}\mathcal{L}_{sm}^{Ld}(R) + \theta _{3}\sum_{i=1}^{M}\mathcal{L}_{GAN}^{Ld}(R,D_{i}) 
\end{equation} 
where $M$ is the number of the scales. $\theta _{1}$, $\theta _{2}$, $\theta _{3}$ are the weights for the different loss terms. In~the fully supervised learning approach (See Equation~\eqref{eq5}), we~only feed the labeled data (denoted by $Ld$). In~the semi-supervised learning (See Equation~\eqref{eq6}), in~each iteration, we~feed one batch of labeled data (denoted by $Ld$) and one batch of unlabeled data (denoted by $Ud$) simultaneously. As~for the labeled data $Ld$, we~calculate its L1 loss (denoted by $\mathcal{L}_{L_1}^{Ld}$), smoothness loss (denoted by $\mathcal{L}_{sm}^{Ld}$), and~GAN loss (denoted by $\mathcal{L}_{GAN}^{Ld}$). The~input to the discriminator is the refined disparity map (denoted by $Fake_1$) and corresponding ground truth (denoted by $Real_1$). Thus,~the~GAN loss for the labeled data $Ld$ is calculated using $Fake_1$ and $Real_1$. As~for the unlabeled data $Ud$, we~only calculate its GAN loss ($\mathcal{L}_{GAN}^{Ud}$) and neglect the other loss terms. The~unlabeled data gets its refined disparity map (denoted by $Fake_2$) from the refiner. Then~feed $Real_1$ and $Fake_2$ into the discriminator to get the GAN loss for the unlabeled data. As~our experiment results show, this approach allows the use of much less labeled data (expensive) in a semi-supervised method (Equation~\eqref{eq6}) to achieve similar performance to the fully supervised method (Equation~\eqref{eq5}) or better performance when using the same amount of labeled data with additional unlabeled data compared with the supervised method. The~combined loss function in the semi-supervised method is:

\begin{equation}
\label{eq6}
\mathcal{L}(R,D) = \theta _{1}\mathcal{L}_{L_1}^{Ld}(R) + \theta _{2}\mathcal{L}_{sm}^{Ld}(R) + 
\dfrac{\theta _{3}}{2}\Big(\sum_{i=1}^{M}\mathcal{L}_{GAN}^{Ld}(R,D_{i}) + \sum_{i=1}^{M}\mathcal{L}_{GAN}^{Ud}(R,D_{i})\Big) 
\end{equation} 

\subsection{Network Architectures}
We adopt a fully convolutional neural network~\cite{FCN} and also the partial architectures from~\mbox{~\cite{Pix2pix,DCGAN,Densenet}} are adapted here for the refiner and discriminator. The~refiner and discriminator use dense blocks to increase local non-linearity.
Transition layers change the size of the feature maps to reduce the time and space complexity~\cite{Densenet}.
In each dense block and transition layer, modules of the form ReLu-BatchNorm-convolution are used. We~use two modules in the refiner and four modules in the discriminator in each dense block, where the filter size is 3~$\times$~3 and stride is 1. The~growth rate $k$ for each dense block is dynamic (unlike~\cite{Densenet}). In~each transition layer, we~only use one module, where the filter size is 4~$\times$~4 and the stride is 2 (except that in Tran.3 of the discriminator the stride is 1). 

\begin{figure}[h!]
	\includegraphics[width =\linewidth]{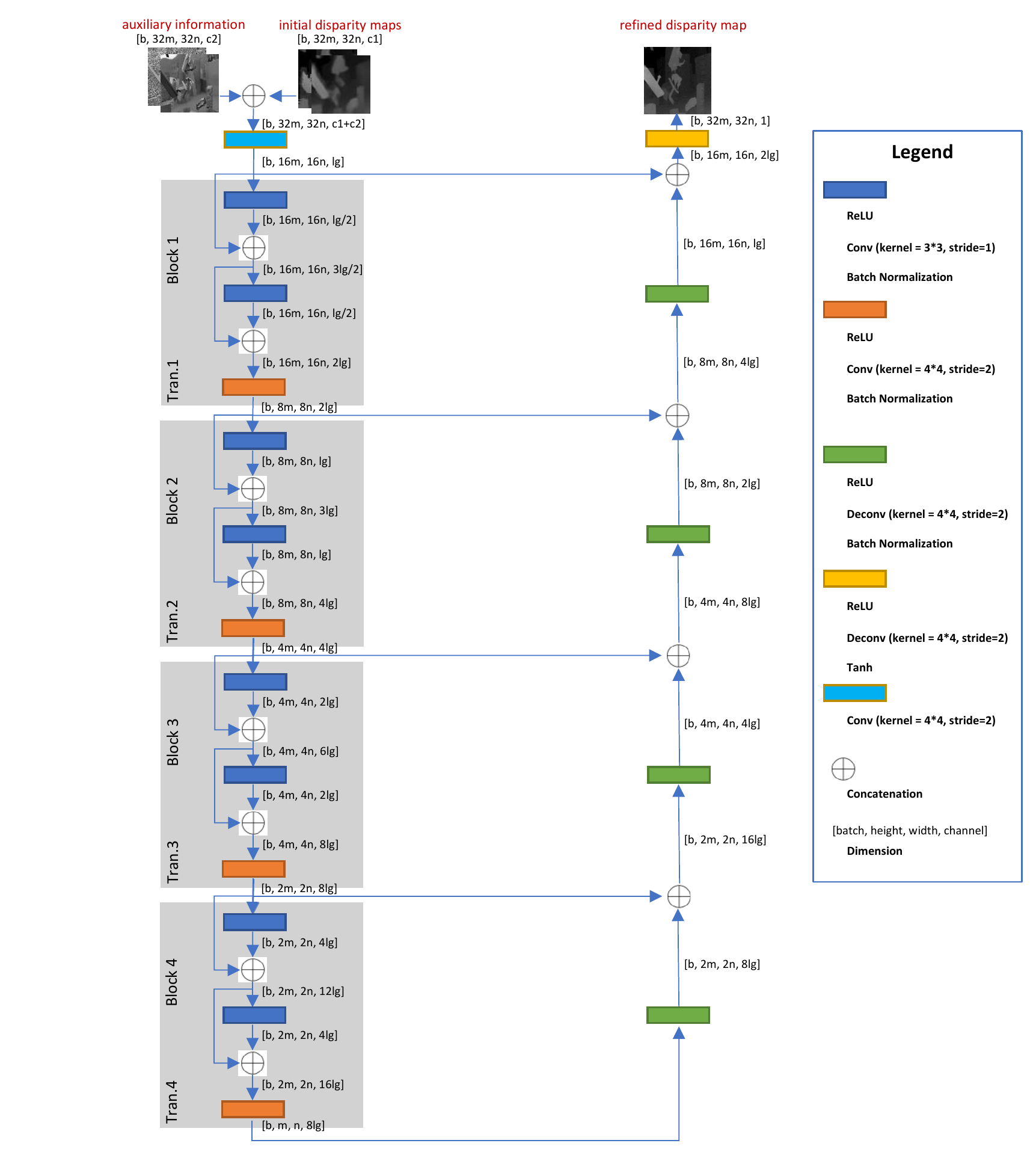}
	\caption{ This figure shows some important hyperparameters and the refiner architecture configuration. Please~refer to Table~\ref{Network_setting} for the specific values in each experiment. Tip: Readers can deepen the refiner by symmetrically adding more dense blocks and deconvolution layer by themselves according to their own needs. } 
	\label{fig:Network_generator}
\end{figure}

Figure~\ref{fig:Network_generator} shows the main architecture of the refiner, where $c1$ initial disparity inputs (the experiments below use $c1=2$ for 2 disparity maps) and $c2$ pieces of information (the experiments below use $c2=2$ for the left intensity image and a gradient of intensity image) are concatenated as input into the generator. The~batch size is $b$ and input image resolution is $32m \times 32n$ \textcolor{black}{($m$, $n$ are integers)}. 
$lg$ is the number of the feature map channels after the first convolution. To~reduce the computational complexity and increase the extraction ability of local details, each dense block contains only 2 internal layers (or modules above). Additionally,~the~skip connections~\cite{Unet} from the previous layers to the latter layers preserve the local details in order not to lose information after the network bottleneck. During~training, a~dropout strategy has been added into the layers in the refiner after the bottleneck to avoid overfitting and we cancel the dropout part during test to produce a determined result.

\begin{figure}[h!]
	\includegraphics[width =0.8\linewidth]{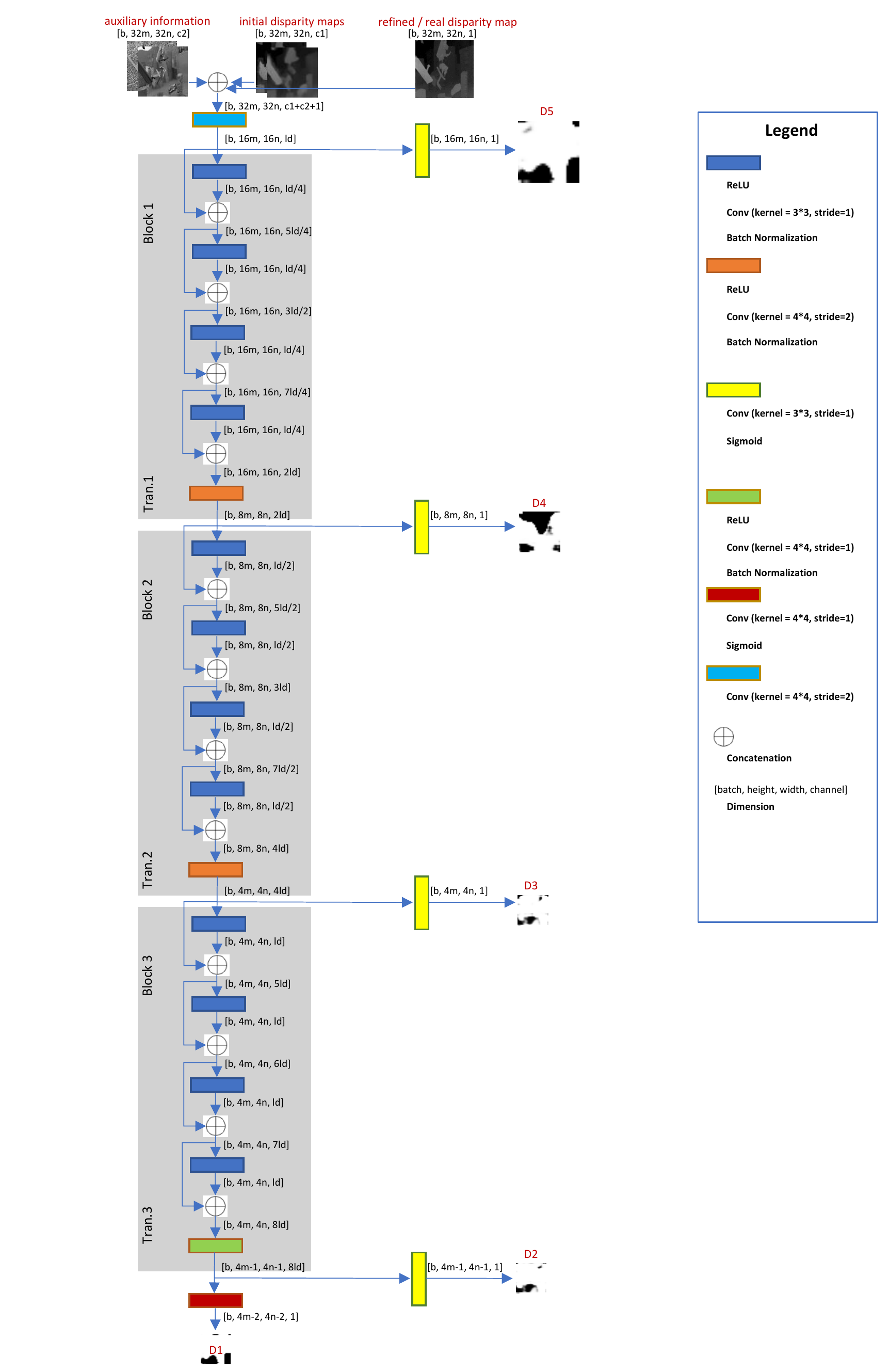}
	\caption{ This figure shows some important hyperparameters and the discriminator architecture configuration. Please~refer to Table~\ref{Network_setting} for the specific values in each experiment. }
	\label{fig:Network_discriminator}
\end{figure}

Figure~\ref{fig:Network_discriminator} is for the discriminator. The~discriminator will only be used during training and abandoned during testing. Thus,~the~architecture of the discriminator will only influence the computational costs during training. The~initial raw disparity maps, information and real or refined disparity maps are concatenated and fed into the discriminator. Each~dense block contains 4 internal layers (or modules above). The~sigmoid function outputs the probability map ($Di, i=1,2, ..., 5$) that the local disparity patch is real or fake at different scales to force the Markov Random Field of the refined disparity map to get closer to the real distribution at different receptive field sizes.

\section{Experimental Evaluation \label{s4}}

The network is implemented using TensorFlow~\cite{TensorFlow} and trained \& tested using an Intel Core i7-7820HK processor (quad-core, 8 MB cache, up~to 4.4 GHz) and Nvidia Geforce GTX 1080Ti. First,~an~ablation study with initial raw disparity inputs (\cite{Dispnet, FPGA_stereo}) is conducted using a synthetic garden dataset to analyze the influence of each factor in the energy function and the objective function. Secondly,~three groups of experiments for three fusion tasks (monocular-stereo, stereo-ToF, stereo-stereo) show the robustness, accuracy and generality of the proposed algorithm using synthetic datasets (SYNTH3~\cite{DL_Fusion}, Scene Flow~\cite{Dispnet}, our~synthetic garden dataset ({They are not available to the public currently})) and real datasets (Kitti2015~\cite{Kitti2015} dataset, Trimbot2020 Garden datasets ({For more description, see Appendix~\ref{TBDataset}). \textcolor{black}{A brief description of datasets ({\textcolor{black}{In the semi-supervised method, as~for each labelled training sample, we~use it with its ground truth in the supervised part. We~also use it without its ground truth in the unsupervised part}}) in the paper is shown in Table~\ref{Dataset}}. All~the results show the proposed algorithm's superiority compared with the state-of-art or classical depth acquisition algorithms (\cite{PSMNet,Monodepth,FPGA_stereo,Dispnet,SGM_Luis, SGM}), the~state-of-art stereo-stereo fusion algorithms~(\cite{Deep_stere_fusion}), the~state-of-art stereo-ToF fusion algorithm~\cite{Reliable_Fusion,DL_Fusion}, and~the state-of-art image style transfer algorithm~\cite{Pix2pix}. 
	
	{\color{black}
		\begin{table}[h!]
			\centering
			\caption{\textcolor{black}{A Brief Description of Datasets in This Paper.}} \label{Dataset}
			\scalebox{.9}[.9]{\begin{tabular}{ccccc}
					\toprule
					& \multicolumn{2}{c}{\bf Supervised } & \multicolumn{2}{c}{\bf Semi-Supervised } \\ \midrule
					\textbf{Dataset} & \textbf{Labeled Training Samples} & \textbf{Test Samples} & \textbf{Training Samples} & \textbf{Test Samples}\\
					\midrule
					Synthetic Garden & 4600 & 421 & 4600 (labeled) & 421 \\
					Scene Flow & 6000 & 1460 & 600 (labeled) + 5400 (unlabeled) & 1460\\
					SYNTH3 & 40 & 15 & 40 (labeled) & 15\\
					Kitti2015 & 150 & 50 & None & None\\
					Trimbot2020 Garden & 1000 & 250 & 1000 (labeled) & 250 \\
					\bottomrule 
			\end{tabular} }
		\end{table} 
	}
	
	In the following experiments, the~inputs to the neural network were first scaled to $32m \times 32n$ and normalized to [$-$1, 1]. After~that, the~input was flipped vertically with a 50\% chance to double the number of training samples. Weights~of all the neurons were initialized from a Gaussian distribution (standard deviation 0.02, mean 0). We~trained all the models in all the experiments with a batch size of 4 in the supervised and semi-supervised method, using Adam~\cite{Adam} with a momentum of 0.5. The~learning rate is changed from 0.005 to 0.0001 gradually. The~method in~\cite{Standard_GAN} is used to optimize the refiner network and discriminator network by alternating between one step on the discriminator and then one step on the refiner. We~set the parameters $\theta_{1}$, $\theta_{2}$, $\theta_{3}$ in Equation~\eqref{eq5} or Equation~\eqref{eq6} to make those terms contribute differently to the energy function in the training process. We~used the $L_1$ distance between the estimated image and ground truth as the error. The~unit is pixel. For~more details about the network settings and computational complexity, please see Table~\ref{Network_setting}. To~highlight the real test, the~network is so fast that it can run the disparity fusion (e.g., up to 384~$\times$~1248 pixels on Kitti2015 datasets) directly at 90 fps without any cropping (e.g., DSF~\cite{Deep_stere_fusion} used samples with 9~$\times$~9 pixels) or down-sampling.

	\begin{table}[h!]
		\centering
		\caption{Computation Time and Parameter Settings. } \label{Ablation_2} 
		\begin{tabular}{cccccccccccccc} 
			\midrule
			& \multicolumn{13}{c}{Ablation Study with Synthetic Garden Dataset} \\ \midrule
			$Para.$ & Test time & $b$ & 32$m$ & 32$n$ 
			& $c_1$ & $c_2$ & $lg$ & $ld$ & $\theta_{1}$ & $\theta_{2}$ & $\theta_{3}$ & $\alpha$ & $\beta$ \\ 
			$Value$ & 0.007 (s/frame) & 4 & 480 & 640 & 2 & 2 & 12 & 12 & 395 & 5 & 1 & 1 & 650 \\ \midrule
			& \multicolumn{13}{c}{Stereo-Monocular Fusion with Synthetic Scene Flow Dataset~\cite{DL_Fusion} } \\ \midrule 
			$Para.$ & Test time & $b$ & $32m$ & $32n$ & $c_1$ & $c_2$ & $lg$ & $ld$ & $\theta_{1}$ & $\theta_{2}$ & $\theta_{3}$ & $\alpha$ & $\beta$ \\ 
			$Value$ & 0.042 (s/frame) & 4 & 256 & 256 & 2 & 2 & 64 & 64 & 199 & 1 & 1 & 0.5 & 100 \\ \midrule
			
			& \multicolumn{13}{c}{Stereo-ToF Fusion with Synthetic SYNTH3 Dataset~\cite{DL_Fusion} } \\ \midrule
			$Para.$ & Test time & $b$ & $32m$ & $32n$ & $c_1$ & $c_2$ & $lg$ & $ld$ & $\theta_{1}$ & $\theta_{2}$ & $\theta_{3}$ & $\alpha$ & $\beta$ \\ 
			$Value$ & 0.012 (s/frame) & 4 & 544 & 960 & 2 & 2 & 16 & 16 & 395 & 5 & 1 & 1 & 1--1.3K \\ \midrule
			& \multicolumn{13}{c}{Stereo-stereo Fusion with Real Kitti2015 Dataset~\cite{Kitti2015} } \\ \midrule
			$Para.$ & Test time & $b$ & $32m$ & $32n$ & $c_1$ & $c_2$ & $lg$ & $ld$ & $\theta_{1}$ & $\theta_{2}$ & $\theta_{3}$ & $\alpha$ & $\beta$ \\ 
			$Value$ & 0.011 (s/frame) & 4 & 384 & 1280 & 2 & 2 & 16 & 16 & 1 & 1 & 1 & 1 & 1 --2K \\ \midrule
			
			& \multicolumn{13}{c}{Stereo-stereo Fusion with Real Trimbot2020 Garden Dataset } \\ \midrule
			$Para.$ & Test time & $b$ & $32m$ & $32n$ & $c_1$ & $c_2$ & $lg$ & $ld$ & $\theta_{1}$ & $\theta_{2}$ & $\theta_{3}$ & $\alpha$ & $\beta$ \\ 
			$Value$ & 0.008 (s/frame) & 4 & 480 & 768 & 2 & 2 & 12 & 12 & 395 & 5 & 1 & 1 & 1--1.3K \\ \bottomrule 
		\end{tabular}
		\label{Network_setting} 
	\end{table}

	\subsection{Ablation Study}
	This subsection shows the effectiveness of the loss function design in Section~\ref{s4.1.1} 
	and the influence of each factor in the final loss function in Section~\ref{s4.1.2}. All~the experiments in this subsection are conducted on our synthetic garden dataset ({The performance demo on the synthetic garden dataset: \url{https://youtu.be/OqTj6h0QwUw}}). 
	The~synthetic garden dataset contains 4600 training samples and 421 test samples under outdoor environments. Each~sample has one pair of rectified stereo images and dense ground truth with resolution 480~$\times$~640 (height $\times$ width) pixels. The~reason why we use a synthetic dataset is that the real dataset (e.g., Kitti2015) does not have dense ground truth, which will influence the evaluation of the network. We~used Dispnet~\cite{Dispnet} and FPGA-stereo~\cite{FPGA_stereo} to generate the two input disparity images. The~authors of~\cite{Dispnet,FPGA_stereo} helped us get the best performance on the dataset as the input to the network. As~for each model, we~trained it for 100 epochs and it takes 20 h or so. The~inference is fast (about 142 frames per second ) for the 480 $\times$ 640 (Height $\times$ Width) resolution input. One~qualitative example is shown in Figure~\ref{fig:Ablation_study} from Section~\ref{s4.1.1}. 
	
	\begin{figure}[h!]
		\centering
		\subfloat[]{\includegraphics[width = 7.5cm,height=5cm]{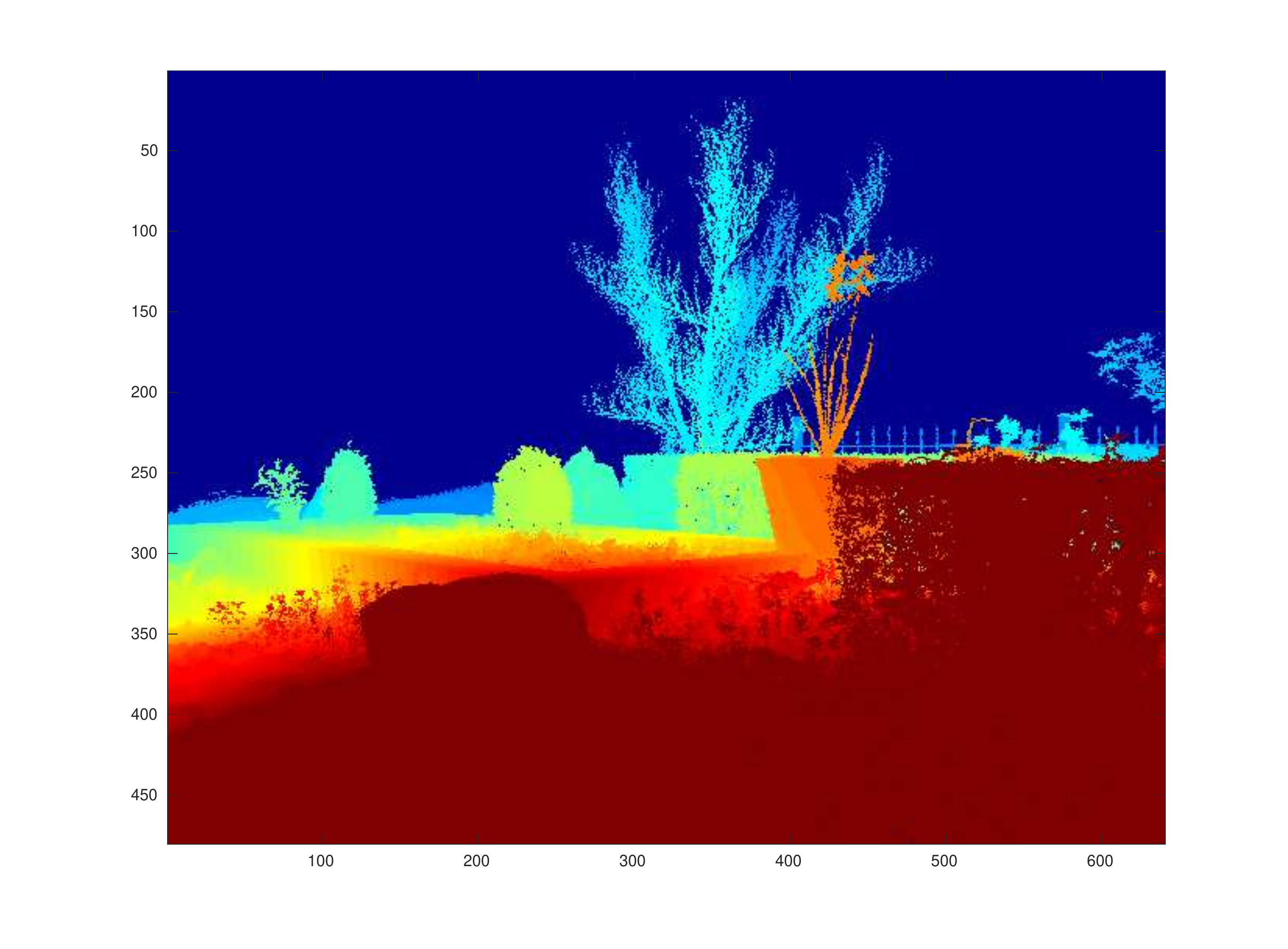}} 
		\vspace{0.001cm} 
		\subfloat[]{\includegraphics[width = 7.5cm,height=5cm]{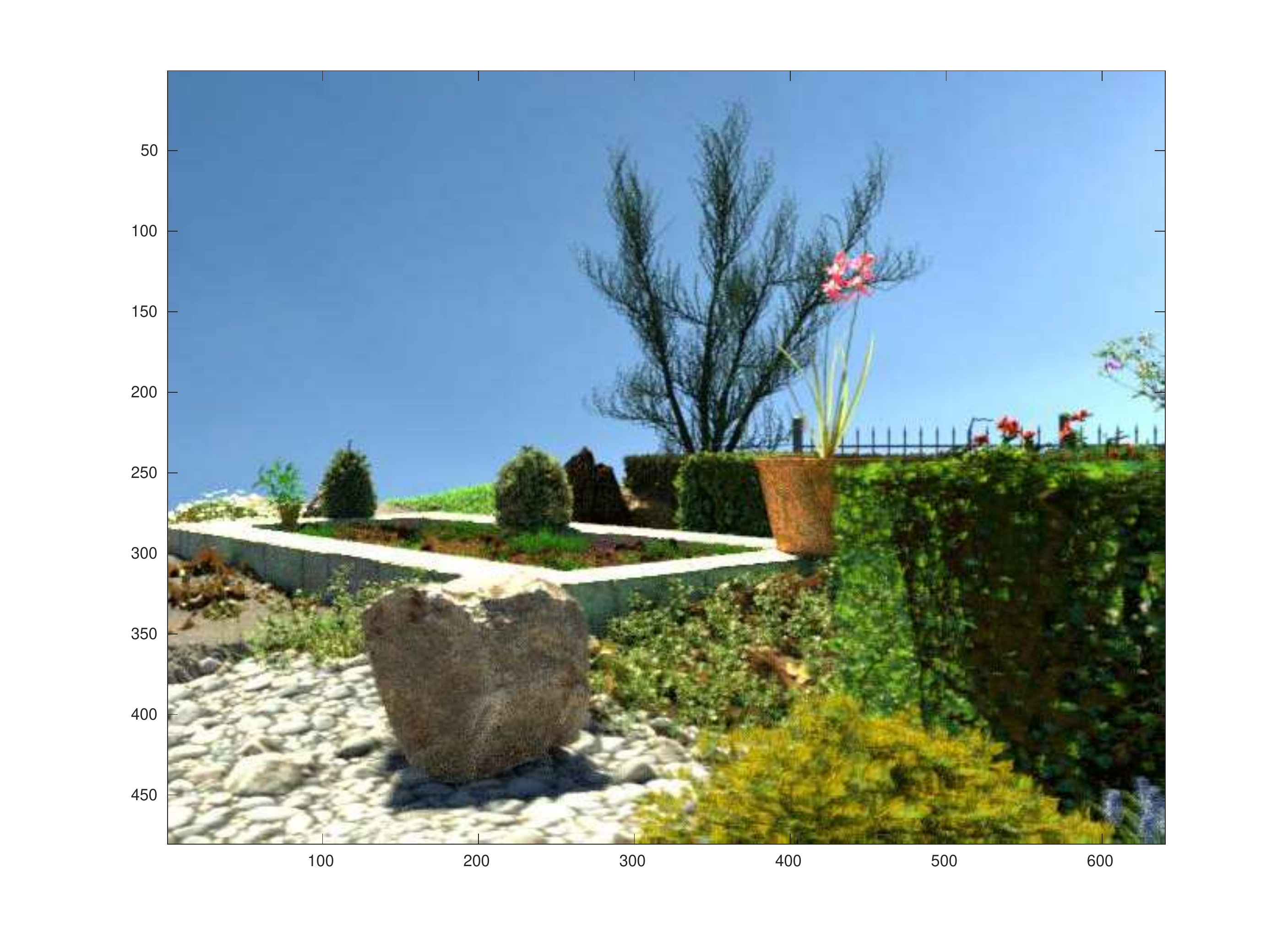}} 
		\vspace{0.001cm}
		
		\subfloat[]{\includegraphics[width = 7.5cm,height=5cm]{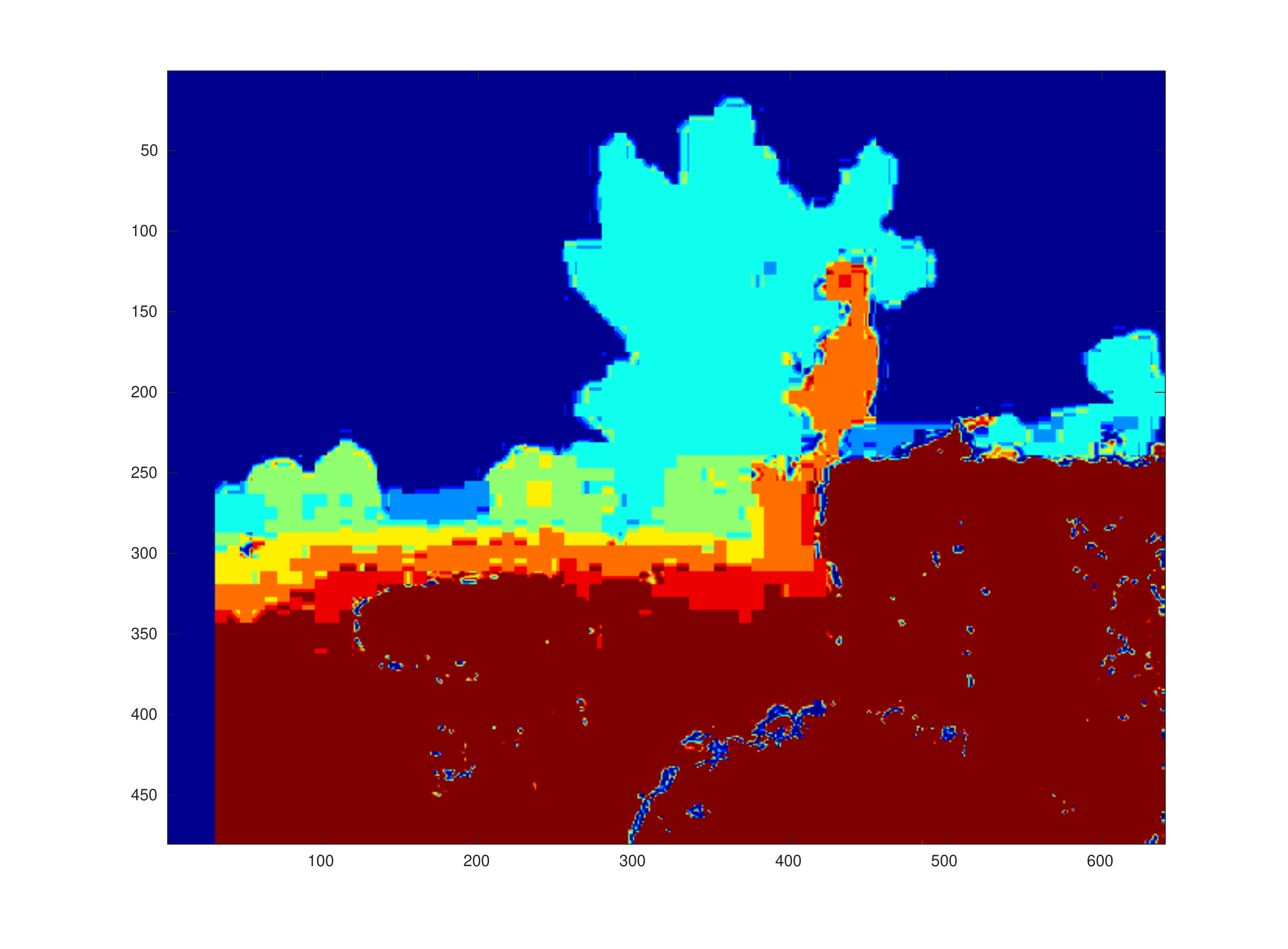}} 
		\vspace{0.001cm}
		\subfloat[]{\includegraphics[width = 7.5cm,height=5cm]{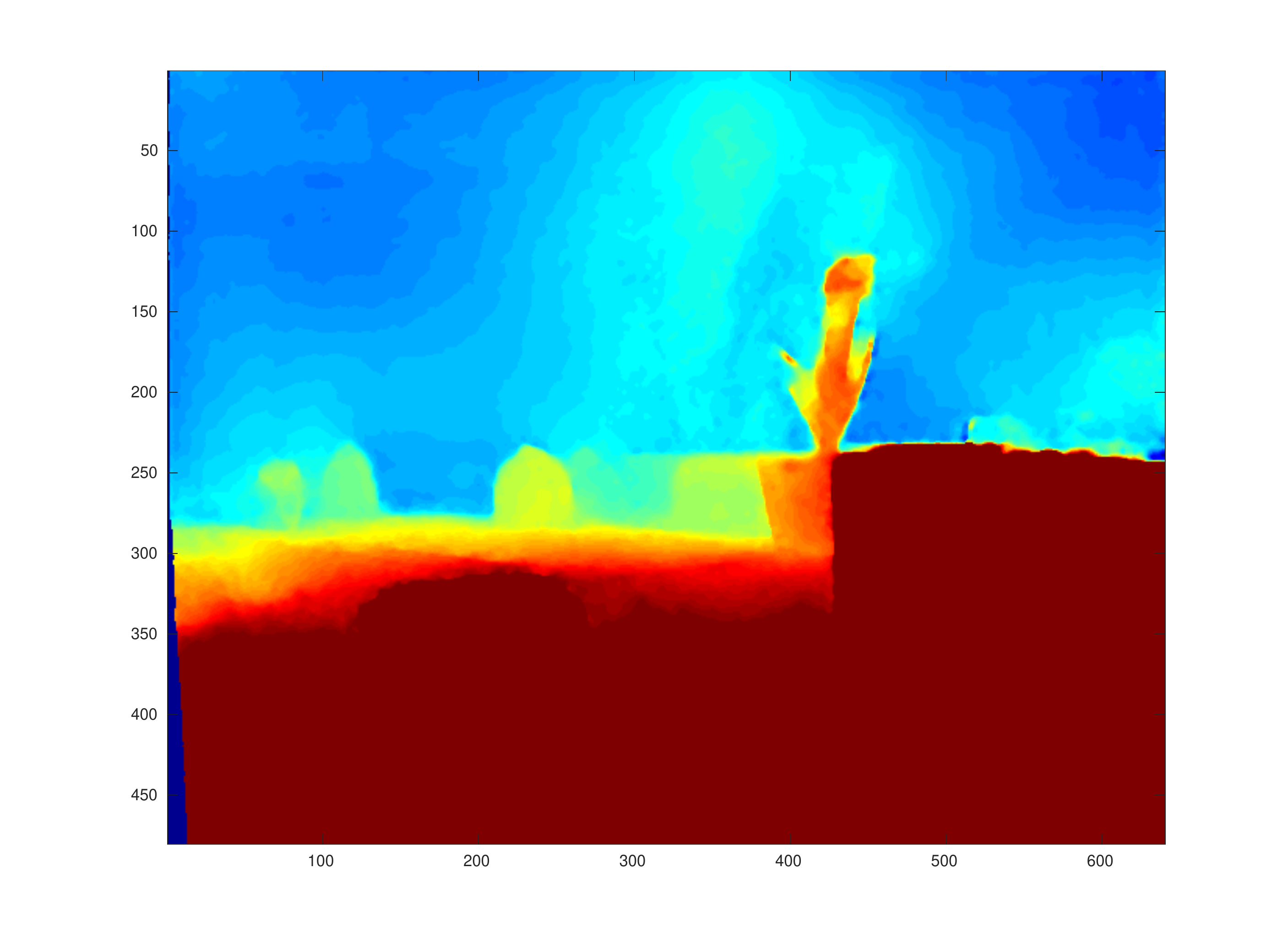}} 
		\vspace{0.001cm}
		
		\subfloat[]{\includegraphics[width = 7.5cm,height=5cm]{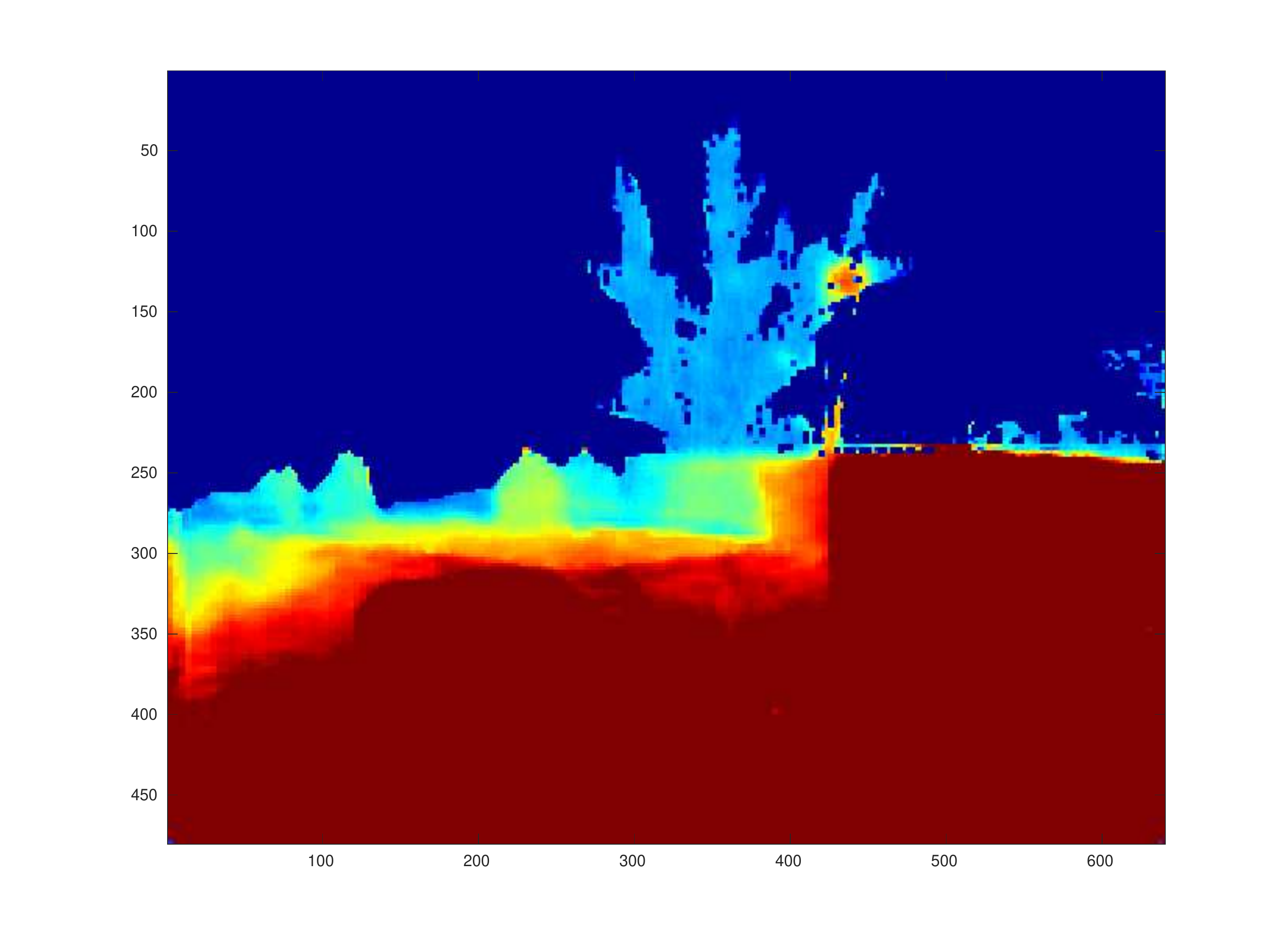}} 
		\vspace{0.001cm}
		\subfloat[]{\includegraphics[width = 7.5cm,height=5cm]{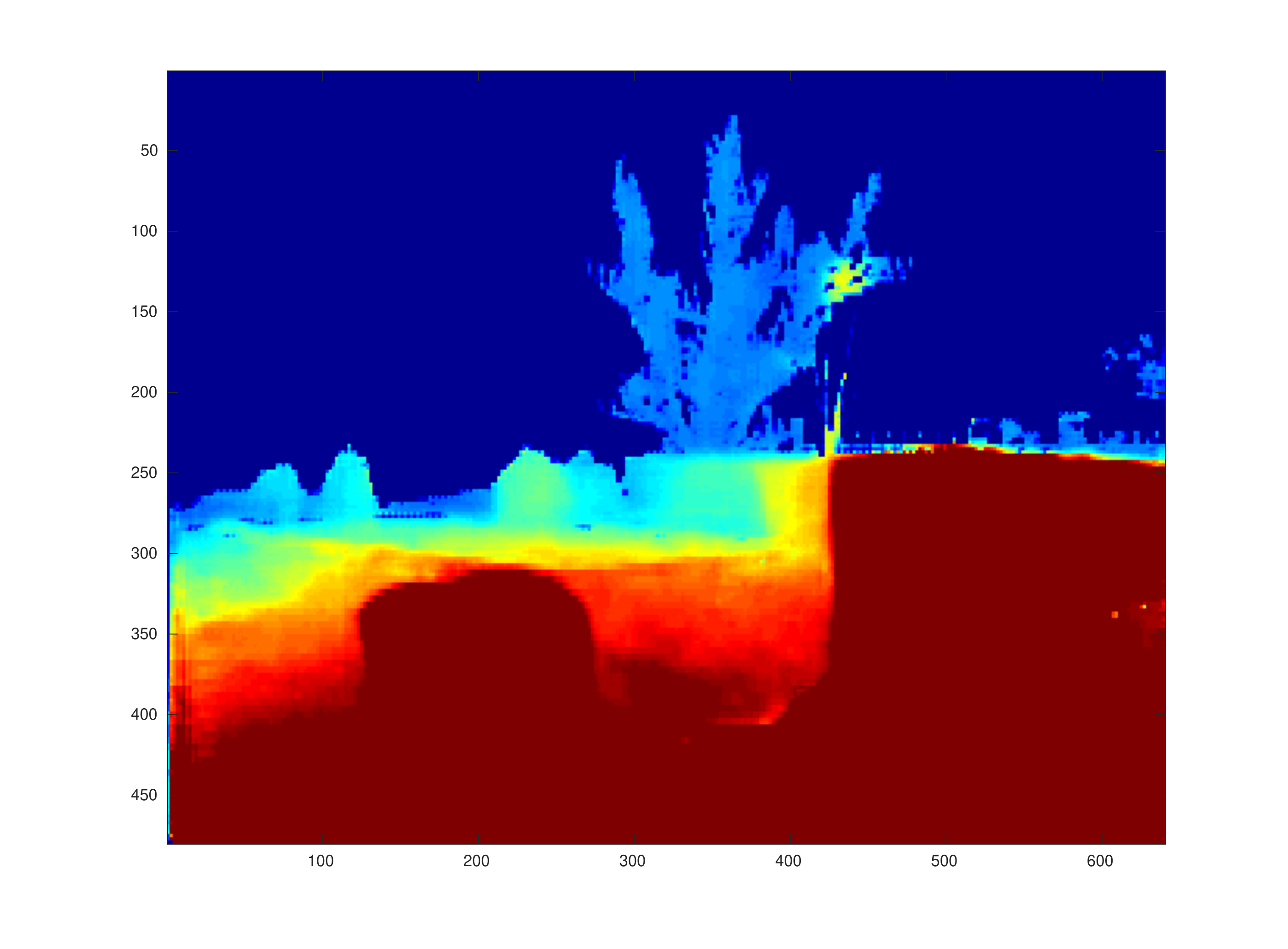}} 
		\vspace{0.001cm} 
		\caption{We fuse two initial raw disparity inputs (\textbf{c},\textbf{d}) to get a refined disparity map (\textbf{e},\textbf{f}) using our Supervised and Semi method on the synthetic garden dataset. (\textbf{a}) is the ground truth and (\textbf{b}) is the corresponding scene. Many,~but~not all, pixels from the fused result are closer to ground truth than the original inputs. (\textbf{a})~Ground Truth; (\textbf{b})~Scene; (\textbf{c})~FPGA Stereo~\cite{FPGA_stereo}; (\textbf{d})~Dispnet~\cite{Dispnet}; (\textbf{e})~Our Supervised; (\textbf{f})~Our Semi.}
		\label{fig:Ablation_study}
	\end{figure}

	\subsubsection{Loss Function Design \label{s4.1.1}}
	
	We aimed at testing the effectiveness of the objective function design from Section~\ref{objective_function}. 
	Table~\ref{Model} defines different combinations of the strategies that were evaluated, based on the objective functions defined in Section~\ref{objective_function}. The~default network settings and some important parameters in this group of experiments, please see ``Ablation Study'' in Table~\ref{Network_setting}.
	\begin{table}[h!] 
		\caption{Model definition.} 
		\label{Model}
		\centering
		\begin{tabular}{cc}
			\toprule
			\textbf{Model Name} & \textbf{Combination} \\
			\midrule
			Supervised & WGAN~\eqref{eq4} + multiscale (M~=~5) + supervised~\eqref{eq5} \\ 
			Semi & WGAN~\eqref{eq4} + multiscale (M~=~5) + semi-supervised~\eqref{eq6} \\
			Monoscale & WGAN~\eqref{eq4} + monoscale (M~=~1) + supervised~\eqref{eq5} \\
			JS-GAN & JS-GAN~\eqref{eq3} + multiscale (M~=~5) + supervised~\eqref{eq5} \\ 
			\bottomrule
		\end{tabular}
	\end{table}
	
	Table~\ref{Simulation} shows the performance of each model. We~used the same amount of data (4600 labeled samples) for the supervised and semi-supervised network training (where the semi-supervised training is augmented with the appropriate number of refined disparity maps and random ground truth). The~test data used 421 samples. The~supervised and semi-supervised methods achieved similar good performance (Semi got the smallest error at 2.84 pixels). The~error of the refined disparity map output by each network is much lower than the error of the input disparity maps. In~the remaining experiments, only the multi-scale supervised and semi-supervised networks are used with WGAN. 
	
	\begin{table}[h!]
		\centering
		\caption{Mean absolute disparity error of each model on Synthetic Garden dataset (421 test samples).} \label{Simulation}
		\begin{tabular}{ccccccc}
			\toprule & \multicolumn{2}{c}{\bf Inputs } & \multicolumn{4}{c}{\bf Experimental Outputs } \\ \midrule
			Experiment & FPGA Stereo~\cite{FPGA_stereo} & DispNet~\cite{Dispnet} & JS-GAN & Monoscale & Supervised & Semi\\
			\midrule 
			Error [px] & 11.41 & 6.28 & 4.40 & 3.40 & 3.10 & 2.84\\
			\bottomrule
		\end{tabular} 
	\end{table}

	\subsubsection{Influence of Each Term in Loss Function  \label{s4.1.2}}
	In this part, we~will change one of the following factors ($\theta_{1}$, $\theta_{2}$, $\theta_{3}$, $\alpha$, $\beta$) in our energy function to see the influence of each cue in Equation~\eqref{eq5}. The~Baseline method in this part is also the Supervised model from Section~\ref{s4.1.1}. The~performance results are listed in Table~\ref{Ablation_study}. We~can see $\mathcal{L}_{L_1}(R)$ in Equation~\eqref{eq1} has the largest influence (corresponding to $\theta_{1}$) and then the gradient information in Equation~\eqref{eq1} (corresponding to $\alpha=0$). After~that, the~smoothing term in Equation~\eqref{eq2} (corresponding to $\theta_{2}$) and $\beta$ have less influence compared with the former factors. The~Loss term in $\mathcal{L}_{GAN}$ (corresponding to $\theta_{3}$) has the least influence. 
	
	\begin{table}[h!]
		\caption{Ablation Study on Each Cue Using the Supervised Model.} 
		\begin{tabular}{ccccccccc} 
			\toprule & \multicolumn{2}{c}{\bf Inputs } & \multicolumn{6}{c}{\bf Experimental Outputs } \\ 
			\midrule
			Experiment & FPGA Stereo~\cite{FPGA_stereo} & DispNet~\cite{Dispnet} & $\theta_{1}=0$ & $\theta_{2}=0$ & $\theta_{3}=0$ & $\alpha=0$ & $\beta=1$ & Baseline \\ \midrule  
			Error [px] & 11.41 & 6.28 & 298.2 & 3.46 & 3.25 & 3.48 & 3.37 &  3.10 \\ \bottomrule 
		\end{tabular}
		\label{Ablation_study} 
	\end{table}

	\subsection{Robustness and Accuracy Test}
	Given that the proposed network does not need confidence values from the specific sensors, the~network architecture can be generalized to fusion tasks using different data sources. Thus,~the~following experiments will input different quality disparity maps from different sources to test the robustness and accuracy of the proposed algorithm.

	\subsubsection{Stereo-Monocular Fusion}

	Monocular depth estimation algorithms are usually less accurate than stereo vision algorithms. Stereo~vision algorithm PLSM
	~\cite{SGM_Luis} and monocular vision algorithm Monodepth~\cite{Monodepth} were used to input the relevant initial disparity maps. Monodepth~was retrained on the Scene Flow dataset (Flying A) with 50 epochs to get its left disparity maps. PLSM~with semi-global matching computed the left disparity map without refinement. The~default network settings and some important parameters of the networks in this part can be seen in ``Stereo-Monocular Fusion'' in Table~\ref{Network_setting}. 6000 labeled samples (80\%) in Scene Flow (Flying A) were used for the supervised training and 600 labeled samples (8\%) + 5400 unlabeled samples for the semi-supervised training. Another~1460 samples (20\%) were used for testing. DSF~\cite{Deep_stere_fusion} is a recent high performance fusion algorithm that we compare with. Pix2pix~\cite{Pix2pix} was set up to use PLSM + Monodepth as inputs and the fused disparity map as output. The~reason to choose Pix2pix as a comparison algorithm is that disparity fusion can be seen as equivalent to an image style transfer and Pix2pix is a famous image style transfer algorithm. DSF~was retrained for 10~epochs (about 5 h per epoch) and Pix2pix~\cite{Pix2pix} was retrained for 100 epochs (0.15 h per epoch).

	The relevant error of each algorithm is shown in Table~\ref{Coarse Input}. The~supervised method (Num~=~6000) and the semi-supervised method (Num~=~600) achieve similar top performances while the semi-supervised method uses much less labeled training data (9 times less than the supervised method). Pix2pix~behaves badly and we neglect it in the following experiments. A~qualitative result comparison can be seen in Figure~\ref{fig:Coarse Input}.
	
	\begin{table}[h!]
		\centering
		\caption{Mean absolute disparity error of stereo-monocular fusion on Scene Flow (1460 test samples).}
		\label{Coarse Input}
		\begin{tabular}{ccccccc}
			\toprule & \multicolumn{2}{c}{\bf Inputs } & \multicolumn{2}{c}{\bf Comparison } & \multicolumn{2}{c}{\bf Our Fused } \\ 
			\midrule
			\textbf{Training Data} & \textbf{PLSM} \newline~\cite{SGM_Luis} & \textbf{Monodepth}~\cite{Monodepth} & \textbf{DSF} \newline~\cite{Deep_stere_fusion} & \textbf{Pix2pix} \newline~\cite{Pix2pix} & \textbf{Supervised} & \textbf{Semi} \\
			\midrule
			Num=600 & 2.41 px & 3.30 px & 2.00 px & 2.91 px & 1.95 px &  1.60 px\\ 
			Num=6000 & 2.41 px & 3.30 px & 1.87 px & 2.65 px &  1.55 px & NA \\ 
			\bottomrule 
		\end{tabular} 
	\end{table} 
	\unskip

	{\begin{figure}[h!]
			\centering
			\subfloat[]{\includegraphics[width = 4.5cm,height=3.5cm]{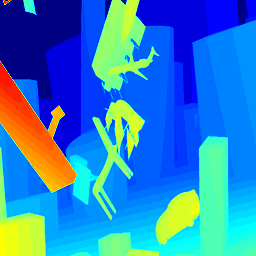}}
			\vspace{0.001cm}
			\hspace{1.5cm} 
			\subfloat[]{\includegraphics[width = 4.5cm,height=3.5cm]{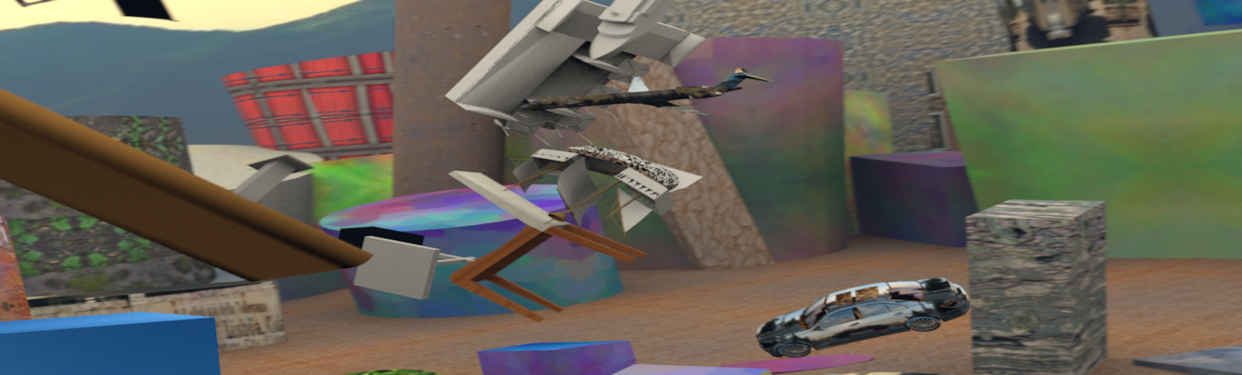}}\\
			\vspace{0.001cm}
			\subfloat[]{\includegraphics[width = 4.5cm,height=3.5cm]{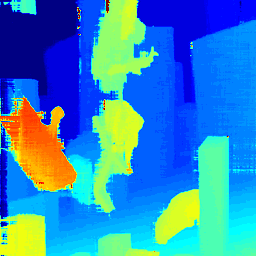}}
			\vspace{0.001cm}
			\hspace{1.5cm}
			\subfloat[]{\includegraphics[width = 4.5cm,height=3.5cm]{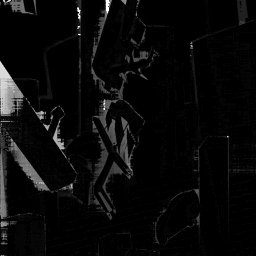}} \\
			\vspace{0.001cm}
			\subfloat[]{\includegraphics[width = 4.5cm,height=3.5cm]{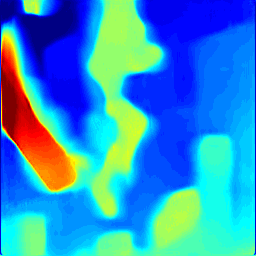}} 
			\vspace{0.001cm}
			\hspace{1.5cm}
			\subfloat[]{\includegraphics[width = 4.5cm,height=3.5cm]{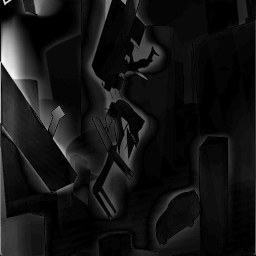}}\\
			\vspace{0.001cm} 
			\subfloat[]{\includegraphics[width = 4.5cm,height=3.5cm]{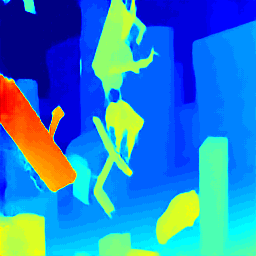}}
			\vspace{0.001cm}
			\hspace{1.5cm}
			\subfloat[]{\includegraphics[width = 4.5cm,height=3.5cm]{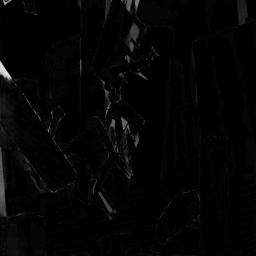}} \\
			\vspace{0.001cm}

			\subfloat[]{\includegraphics[width = 4.5cm,height=3.5cm]{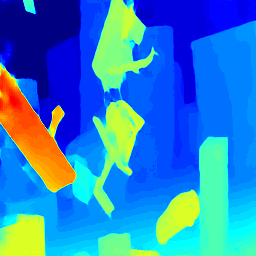}} 
			\vspace{0.001cm}
			\hspace{1.5cm}
			\subfloat[]{\includegraphics[width = 4.5cm,height=3.5cm]{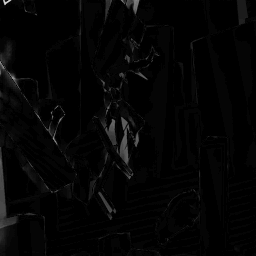}}
			\vspace{0.001cm} 
			\caption{A qualitative result with inputs from PLSM
				~\cite{SGM_Luis} and Monodepth~\cite{Monodepth} in stereo-monocular fusion. The~lighter pixels represent bigger disparity errors in figure (\textbf{d},\textbf{f},\textbf{h},\textbf{j}). (\textbf{a})~Ground Truth; (\textbf{b})~Color image; (\textbf{c})~PLSM~\cite{SGM_Luis}; (\textbf{d})~PLSM error; (\textbf{e})~Monodepth~\cite{Monodepth}; (\textbf{f})~Monodepth error; (\textbf{g})~Supervised 1; (\textbf{h})~Supervised 1 error; (\textbf{i})~semi 2; (\textbf{j})~Semi 2 error.}
			\label{fig:Coarse Input}
	\end{figure}}
	\unskip 
	
	\subsubsection{Stereo-ToF Fusion}

	The default network settings and some important parameters of the networks in this part, can~be seen in ``Stereo-ToF Fusion'' in Table~\ref{Network_setting}. The~network was trained on the SYNTH3 dataset (40 training and 15 test samples with resolution 540 $\times$ 960 pixels). Semi-global matching from OpenCV was used to get the stereo disparity map, with the point-wise Birchfield-Tomasi metric, 7~$\times$~7-pixel window size and 8-path optimization. The~initial ToF depth map was projected onto the right stereo camera image plane and up-sampled and converted to the disparity map. Limited~by the very small number of training samples, the~proposed networks do not reach their best performance. But,~compared with the input disparity maps, the~proposed methods perform slightly better (See Table~\ref{ToF-stereo fusion}). The~experiment results for SGM stereo, ToF, LC~\cite{Reliable_Fusion} and DLF~\cite{DL_Fusion} are from the paper~\cite{DL_Fusion} because we used the same dataset as~\cite{DL_Fusion} from their website ({\url{http://lttm.dei.unipd.it/paper_data/deepfusion/}}). 
	The~proposed Supervised method performs less well because of the insufficient number of training samples. However,~the~proposed Semi method ranks first among all of the stereo-ToF fusion algorithms. One~qualitative result is shown in Figure~\ref{fig:ToF-stereo}.

	\begin{table}[h!]
		\centering
		\caption{Mean absolute disparity error of ToF-stereo fusion on SYNTH3 (15 test samples).} \label{ToF-stereo fusion}
		\begin{tabular}{ccccccc}
			\hline & \multicolumn{2}{c}{\bf Inputs } & \multicolumn{2}{c}{\bf Comparison } & \multicolumn{2}{c}{\bf Our Fused } \\ 
			\hline
			\textbf{Training Data} & \textbf{SGM} \newline \bf Stereo & \textbf{ToF} & \textbf{LC} \newline~\cite{Reliable_Fusion} & \textbf{DLF} \newline~\cite{DL_Fusion} & \textbf{\bf Supervised} & \textbf{Semi} \\
			\hline 
			Num=40 & 3.73 px & 2.19 px & 2.07 px & 2.06 px & 2.18 px &  2.02 px\\
			\hline 
		\end{tabular} 
	\end{table} 
	\unskip

	\begin{figure}[h!]
		\centering
		\subfloat[]{\includegraphics[width = 7.5cm,height=3.5cm]{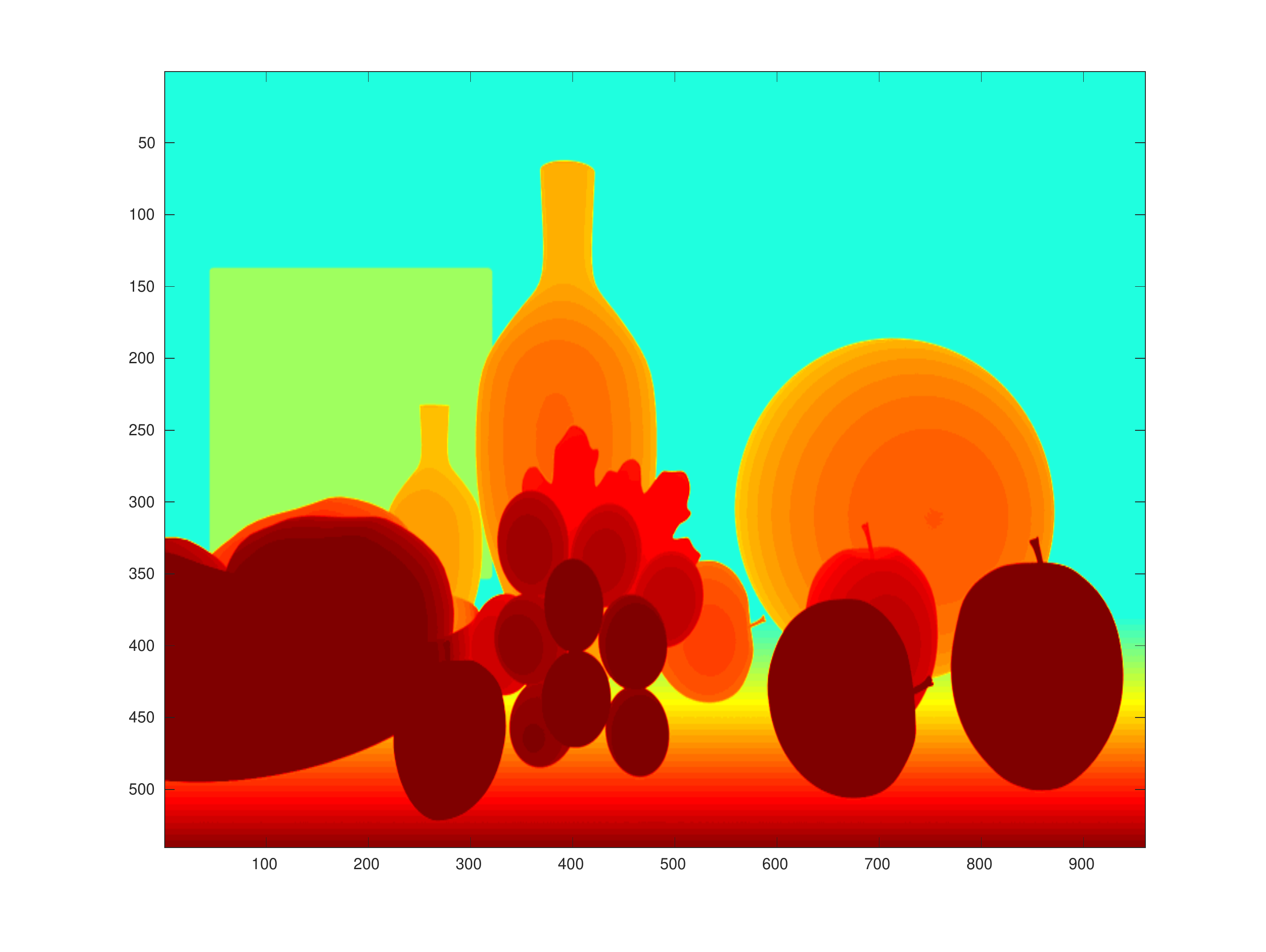}}
		\vspace{0.001cm}
		\hspace{0.1cm}
		\subfloat[]{\includegraphics[width = 7.5cm,height=3.5cm]{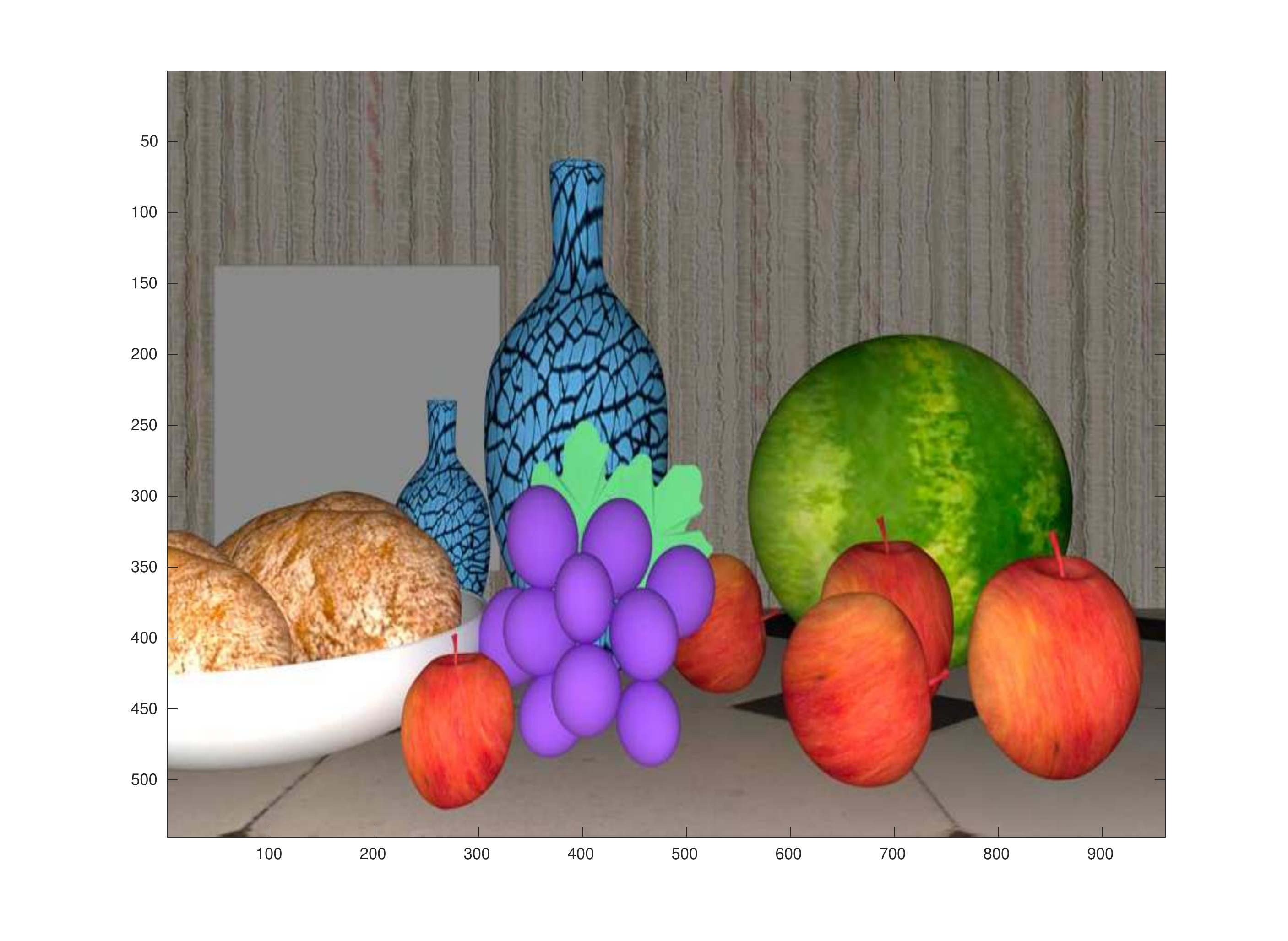}}
		\vspace{0.001cm}

		\subfloat[]{\includegraphics[width = 7.5cm,height=3.5cm]{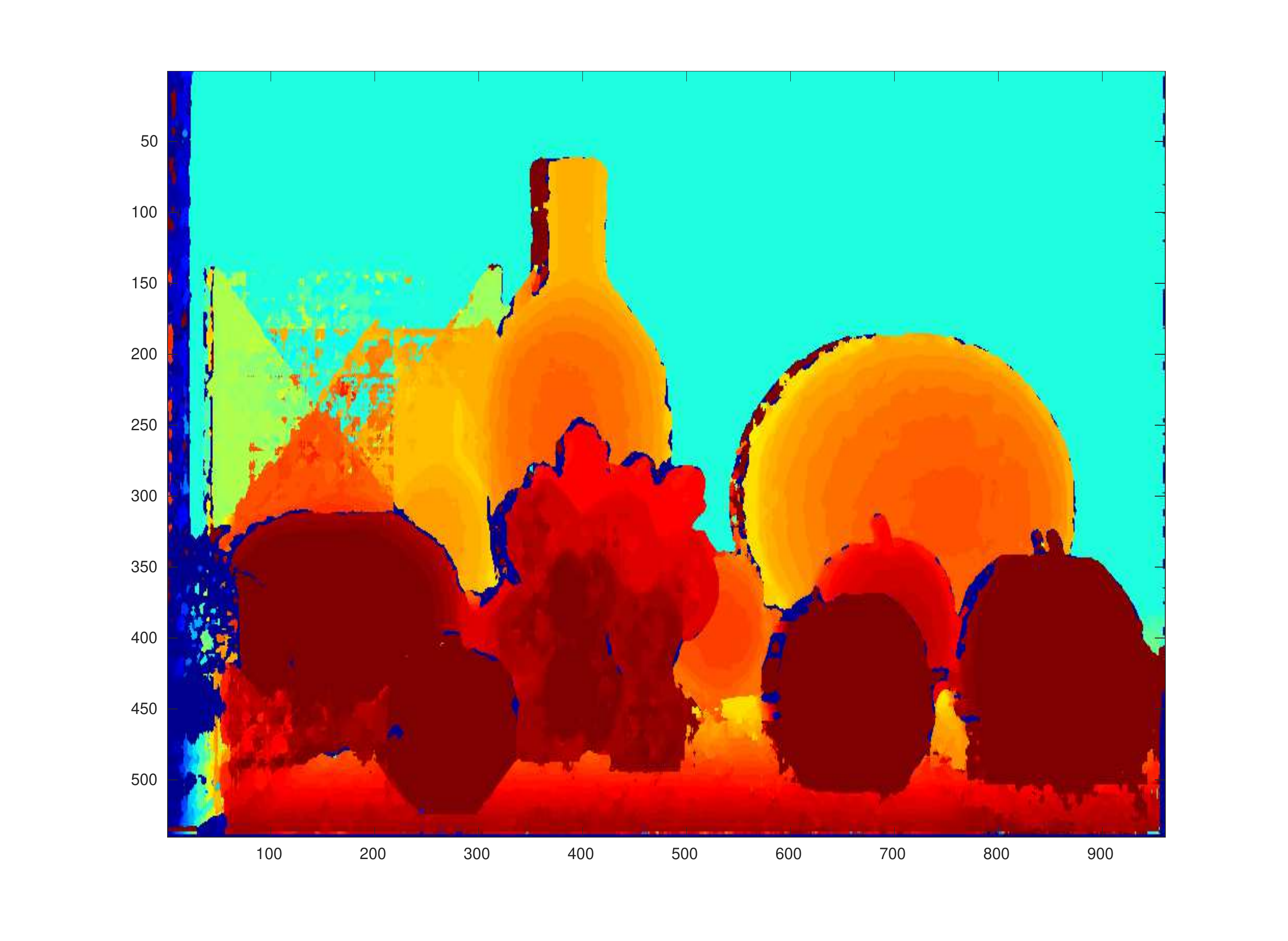}}
		\vspace{0.001cm}
		\hspace{0.1cm} 
		\subfloat[]{\includegraphics[width = 7.5cm,height=3.5cm]{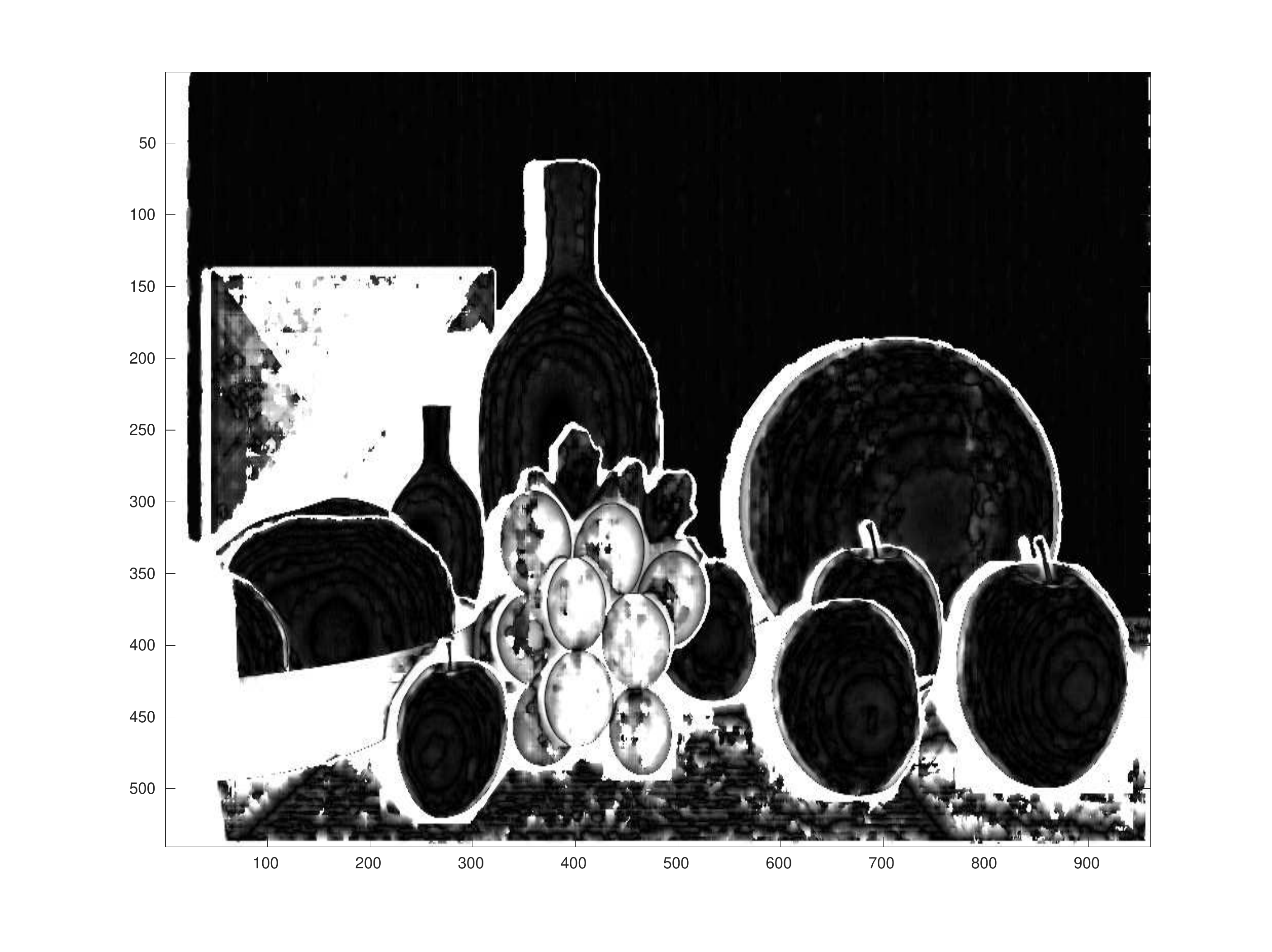}} 
		\vspace{0.001cm}
		\subfloat[]{\includegraphics[width = 7.5cm,height=3.5cm]{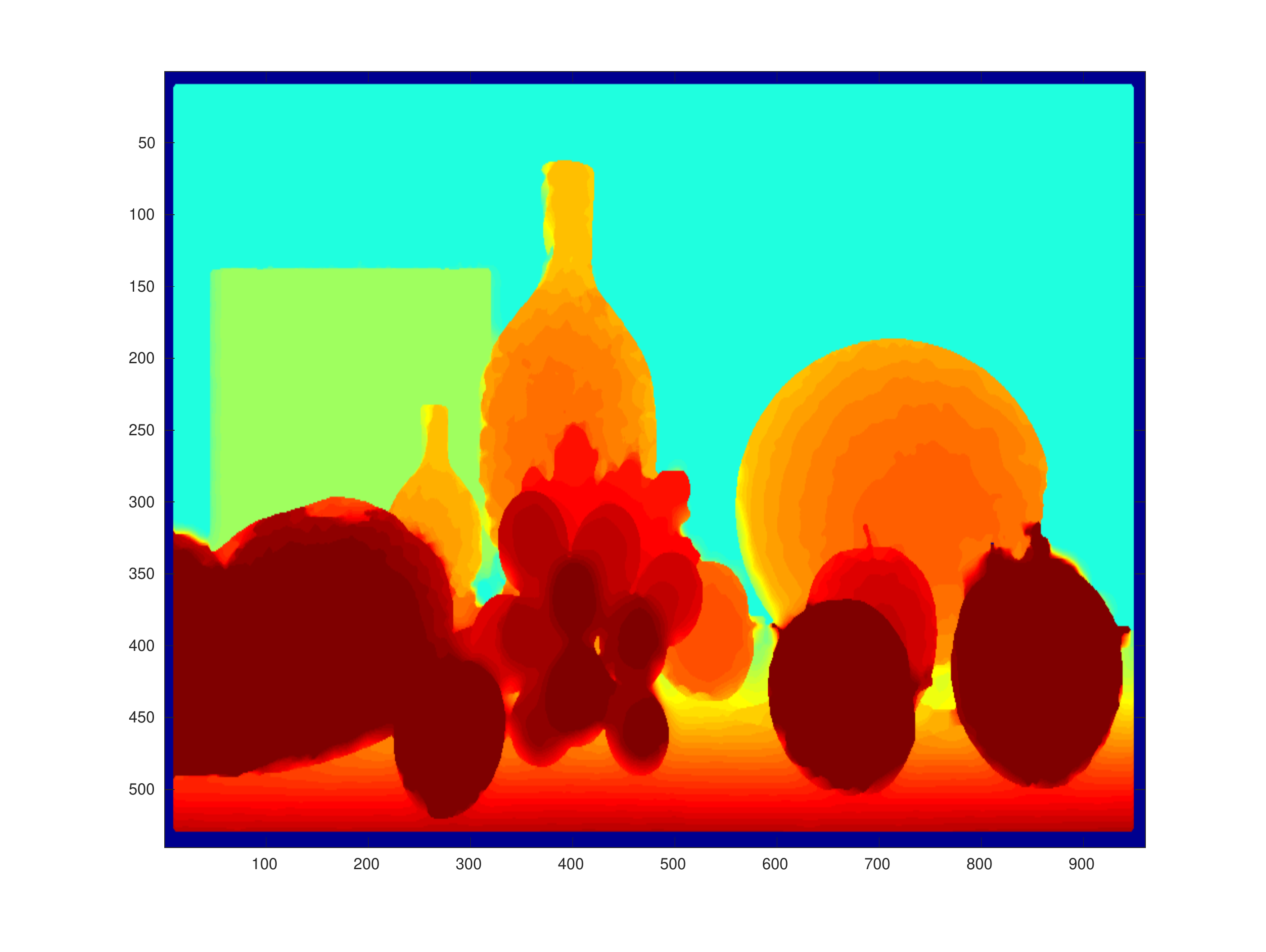}} 
		\vspace{0.001cm}
		\hspace{0.1cm} 
		\subfloat[]{\includegraphics[width = 7.5cm,height=3.5cm]{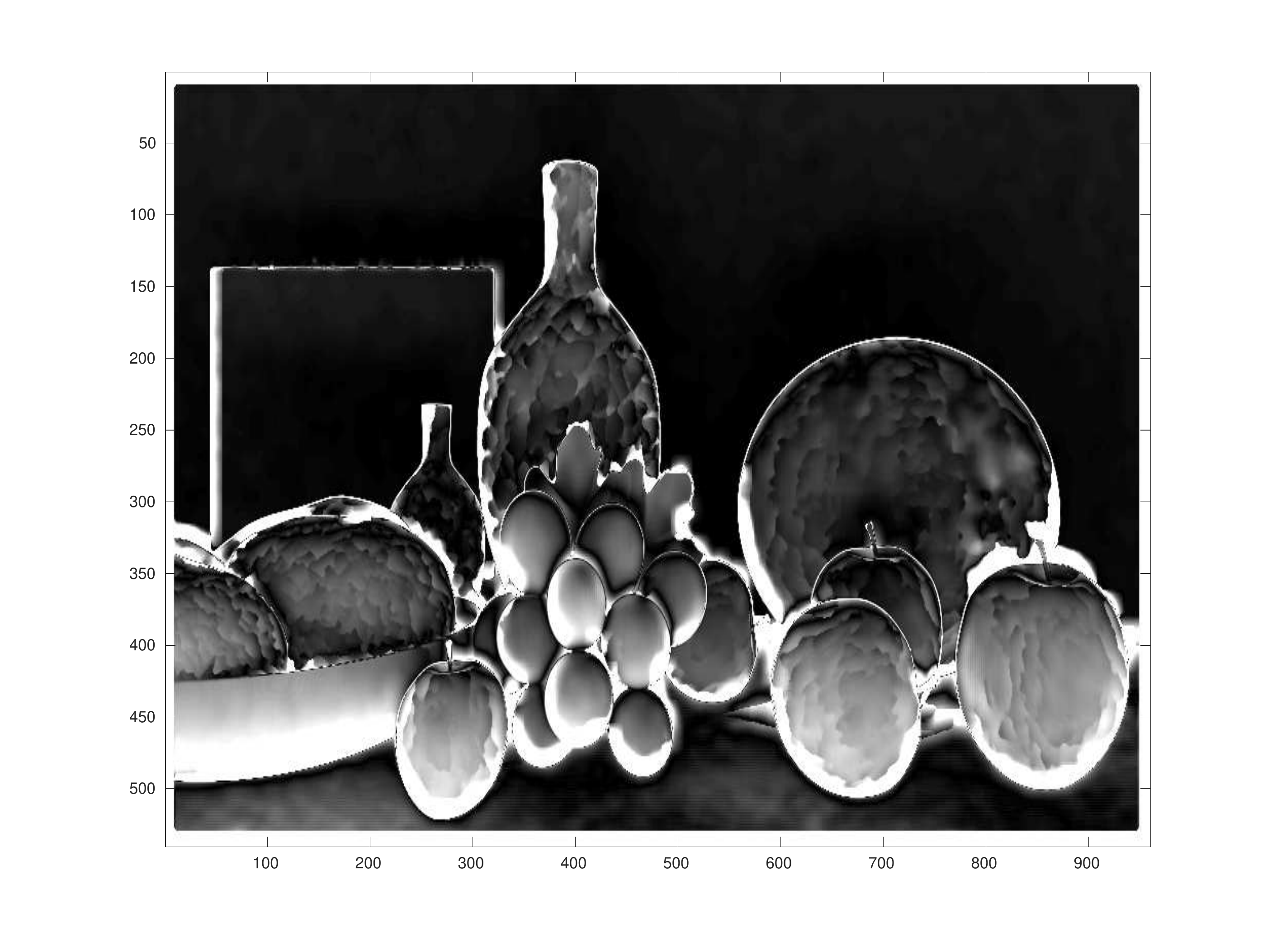}}
		\vspace{0.001cm} 
		\subfloat[]{\includegraphics[width = 7.5cm,height=3.5cm]{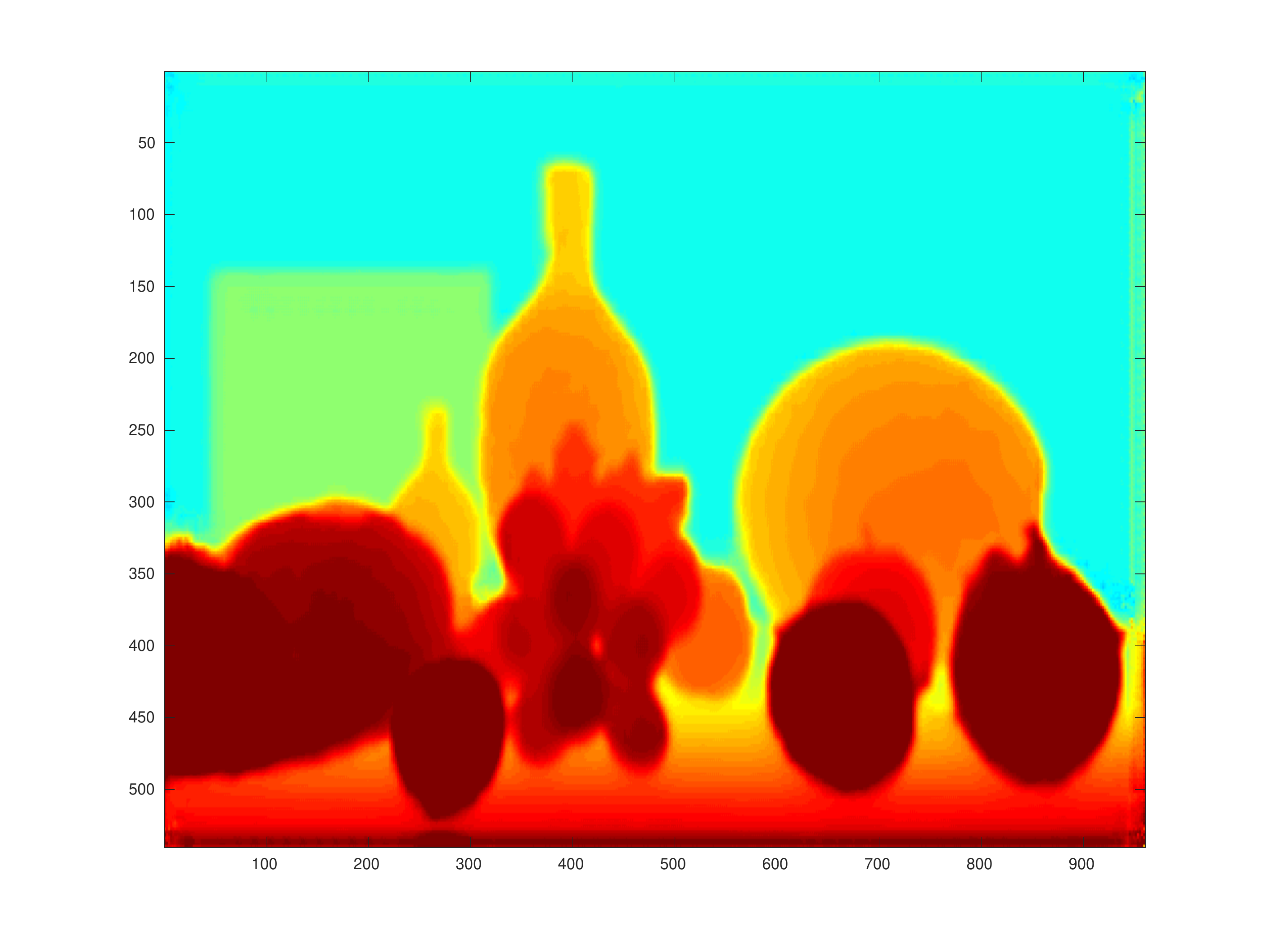}}
		\vspace{0.001cm}
		\hspace{0.1cm} 
		\subfloat[]{\includegraphics[width = 7.5cm,height=3.5cm]{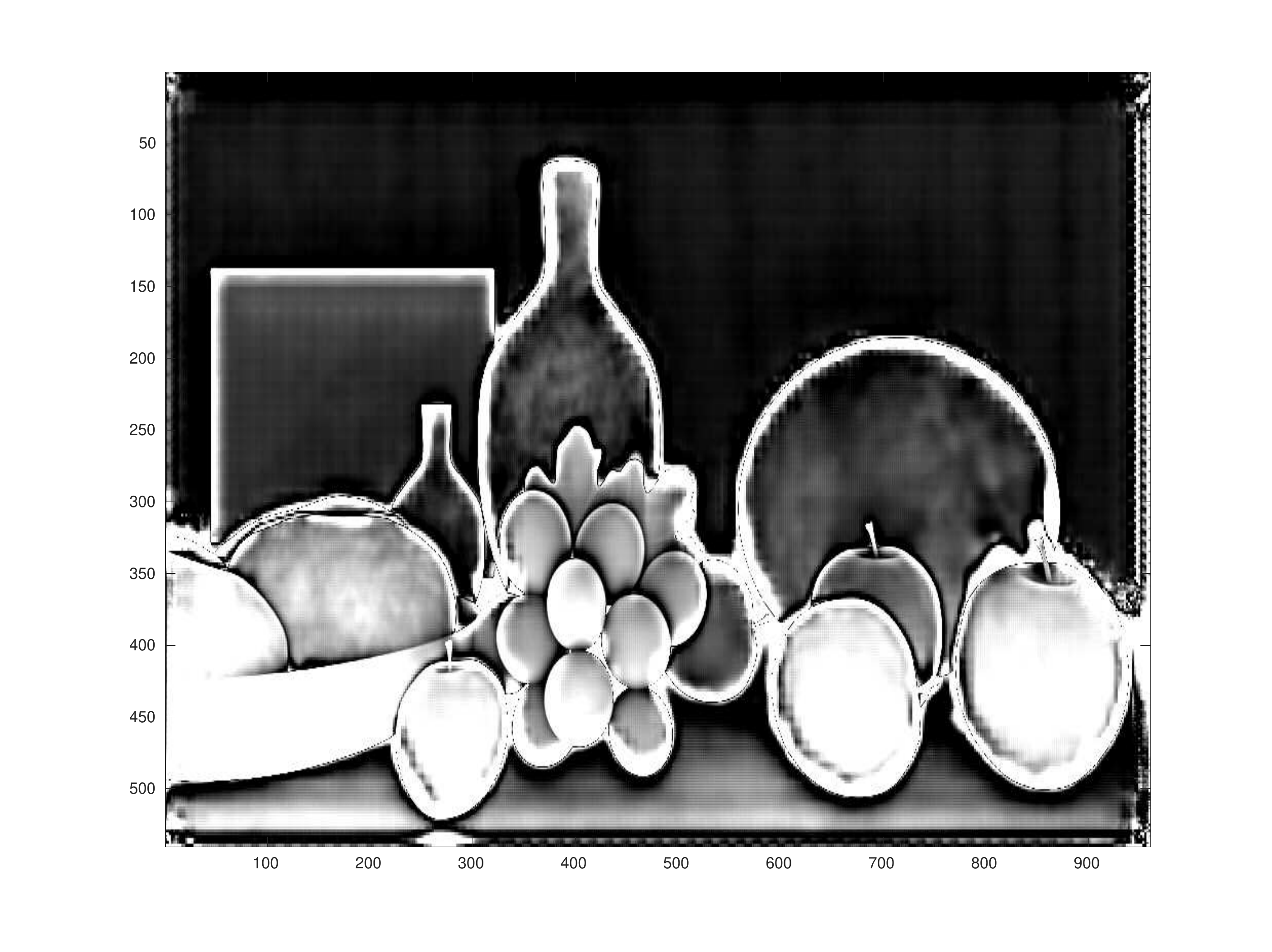}} 
		\vspace{0.001cm}
		\subfloat[]{\includegraphics[width = 7.5cm,height=3.5cm]{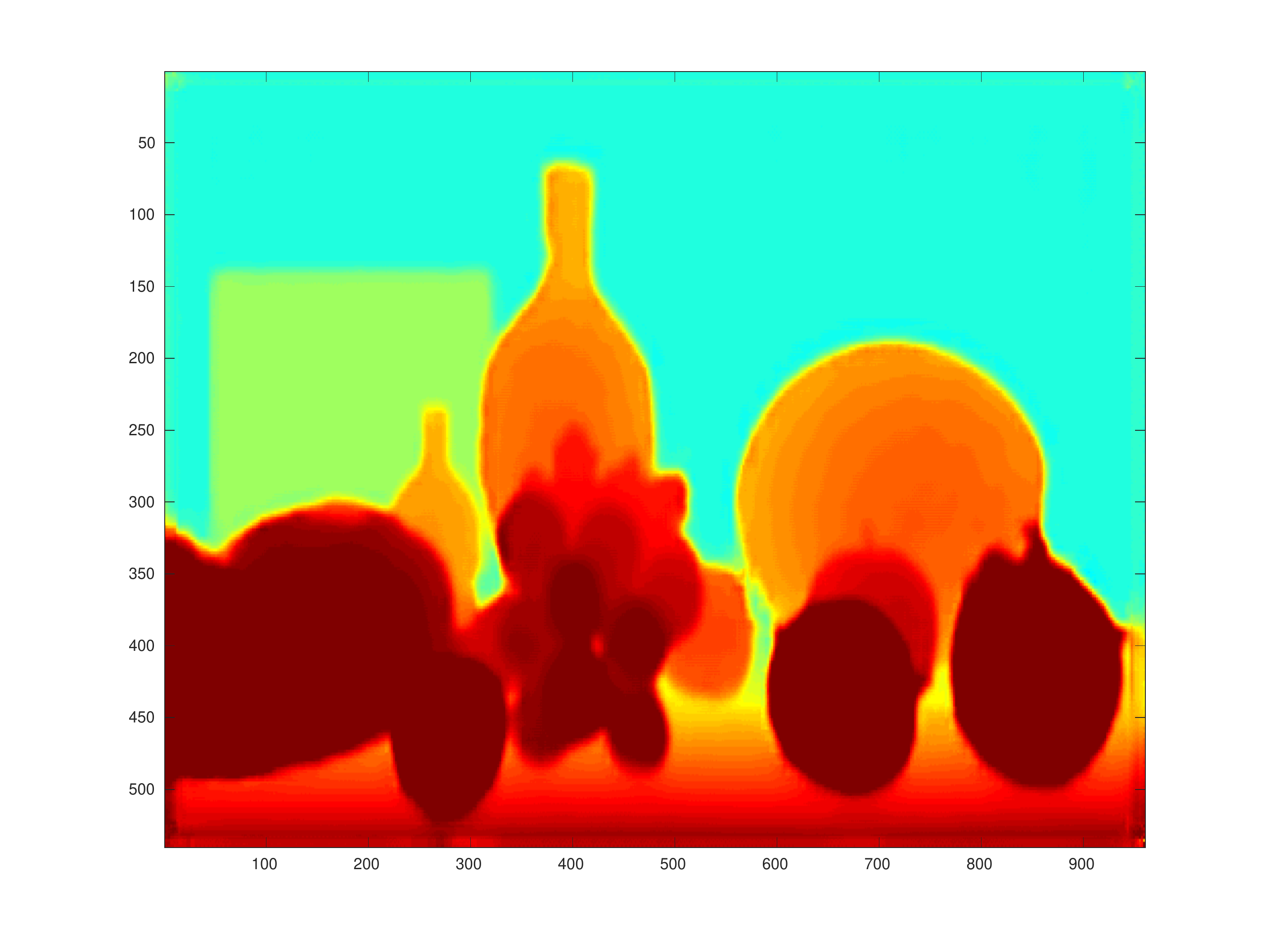}} 
		\vspace{0.001cm}
		\hspace{0.1cm} 
		\subfloat[]{\includegraphics[width = 7.5cm,height=3.5cm]{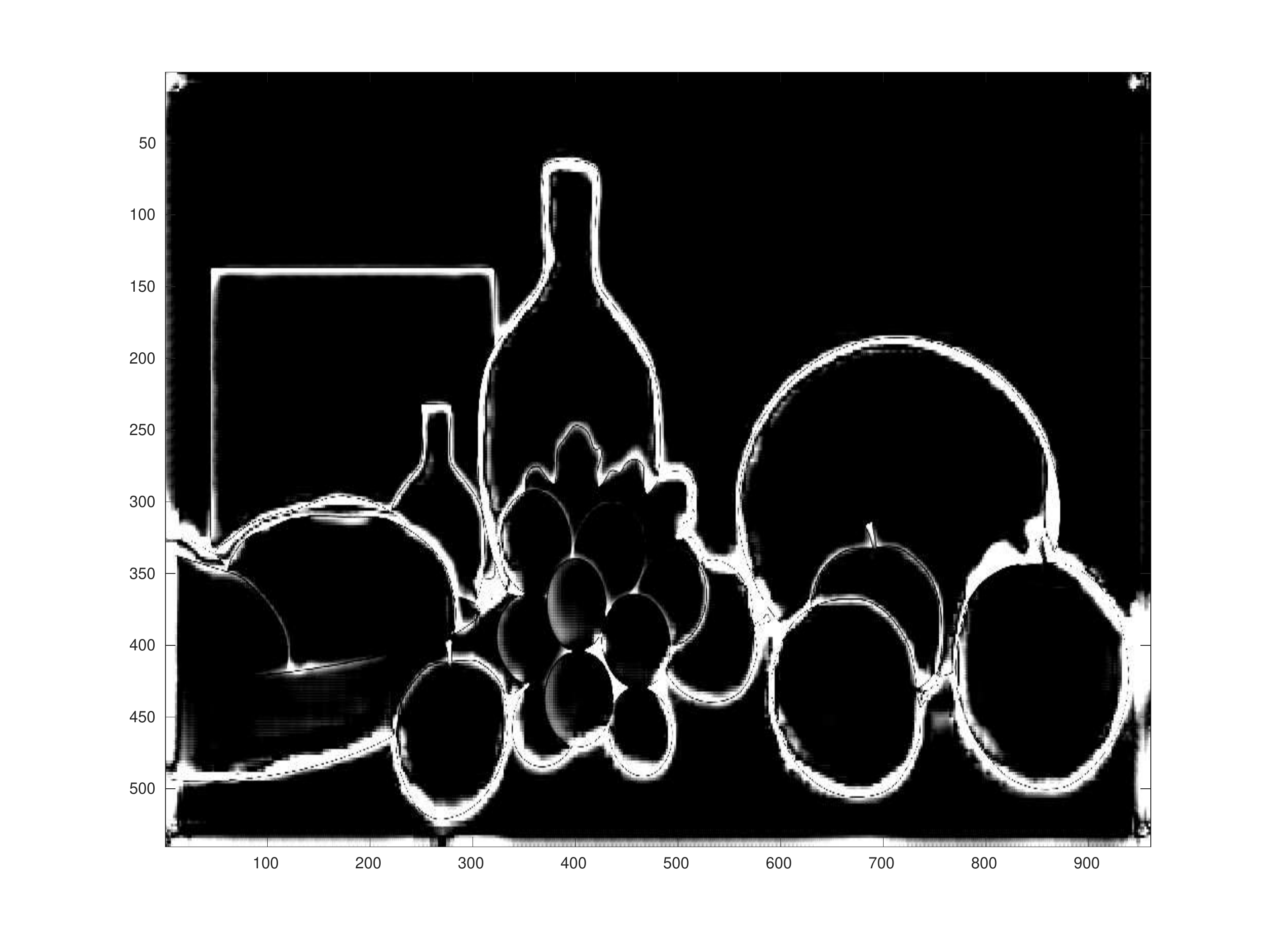}}
		\vspace{0.001cm} 
		\caption{One qualitative result for ToF-stereo fusion with many invalid pixels input. The~inputs are from ToF and disparity calculation algorithm using SGM in OpenCV. The~lighter pixels in (\textbf{d},\textbf{f},\textbf{h},\textbf{j}) represent larger disparity error. (\textbf{a})~Ground Truth; (\textbf{b})~Color image; (\textbf{c})~ToF; (\textbf{d})~ToF error; (\textbf{e})~SGM OpenCV; (\textbf{f})~SGM error; (\textbf{g})~Supervised 1; (\textbf{h})~Supervised error; (\textbf{i})~Semi 2; (\textbf{j})~Semi 2 error.}
		\label{fig:ToF-stereo}
	\end{figure}
	\unskip

	\subsubsection{Stereo-Stereo Fusion \label{s4.2.3}}
	\unskip 
	\paragraph{Performance on Kitti2015 Dataset}

	We tested the proposed network on the real Kitti2015 dataset, which used a Velodyne HDL-64E 
	Lidar scanner to get the sparse ground truth and a 1242~$\times$~375 resolution stereo camera to get stereo image pairs. The~initial training dataset contains 200 labeled samples. We~used 50 samples from `000000\_10.png' to `000049\_10.png' 
	in the Kitti2015 training dataset as our test dataset. We~used the other 150 samples as our training set for fine-tuning. By~flipping the training samples vertically, we~doubled the number of training samples. We~used the state-of-art stereo vision algorithm PSMNet~\cite{PSMNet} as one of our inputs. We~used their released pre-trained model~({PSMNet~\cite{PSMNet}: \url{https://github.com/JiaRenChang/PSMNet}})
	on the Kitti2015 dataset to get the disparity maps. A~traditional stereo vision algorithm SGM~\cite{SGM} is used as the second input to the network. We~set their parameters to produce more reliable but sparse disparity maps. More~specifically, we~used the implementation (`disparity' function) from Matlab2016b. The~relevant parameters are: `DisparityRange' [0, 160], `BlockSize' 5, `ContrastThreshold' 0.99, `UniquenessThreshold' 70, `DistanceThreshold' 2. The~settings of the neural network are shown in ``Stereo-stereo Fusion with Real Kitti2015 Dataset'' in Table~\ref{Network_setting}. We~compared the algorithm with the state-of-art technique~\cite{Deep_stere_fusion} in stereo-stereo fusion and also stereo vision inputs~\cite{PSMNet,SGM}. As~the ground truth of Kitti2015 is sparse, we~do not compare the semi-supervised method (which requires learning the disparity Markov Random Field). We~trained our supervised method on the synthetic garden dataset first and then fine-tuned the pre-trained model on the Kitti2015 dataset. We~used 150 labeled samples from `00050\_10.png' to `000199\_10.png' 
	in the initial training dataset for the supervised method's fine-tuning. The~relevant results are shown in Table~\ref{Medium Input}. The~same conclusion can be made as with the stereo-monocular and stereo-ToF fusion: the proposed method is accurate and robust. An~example result of stereo-stereo fusion is shown in Figure~\ref{fig:example_demo}. We~can see that the proposed method compensates for the weaknesses of the inputs and refines the initial disparity maps effectively. Compared~with SGM~\cite{SGM} (0.78 pixels)~({This is a more accurate disparity but is calculated only using more reliable pixels. On~average only 40\% of the ground truth pixels are used. If~we use all the valid ground truth to calculate its error, it~is 22.13 pixels}), the~fused results are much more dense and accurate. Compared~with PSMNet, the~proposed method preserves the details better (e.g., tree, sky), which are missing in the ground truth though. Our~network can deal with the input (resolution: 384~$\times$~1280) at 0.011 s/frame, which is real-time and very fast.

	\begin{table}[h!]
		\centering
		\caption{Mean absolute disparity error of stereo-stereo fusion on Kitti2015 (50 test samples).} \label{Medium Input}
		\begin{tabular}{ccccc}
			\toprule & \multicolumn{2}{c}{\bf Inputs } & \multicolumn{1}{c}{\bf Comparison } & \multicolumn{1}{c}{\bf Our Fused } \\ 
			\midrule
			\textbf{Training Data} & \textbf{SGM}~\cite{SGM} & \textbf{PSMNet}~\cite{PSMNet} & \textbf{DSF}~\cite{Deep_stere_fusion} & \textbf{Supervised} \\
			\midrule 
			Num=150 & 0.78 px & 1.22 px & 1.20 px & 1.17 px \\
			\bottomrule 
		\end{tabular} 
	\end{table} 
	\unskip 
	
	\begin{figure}[h!]
		\centering
		\subfloat[]{\includegraphics[width=0.5\textwidth]{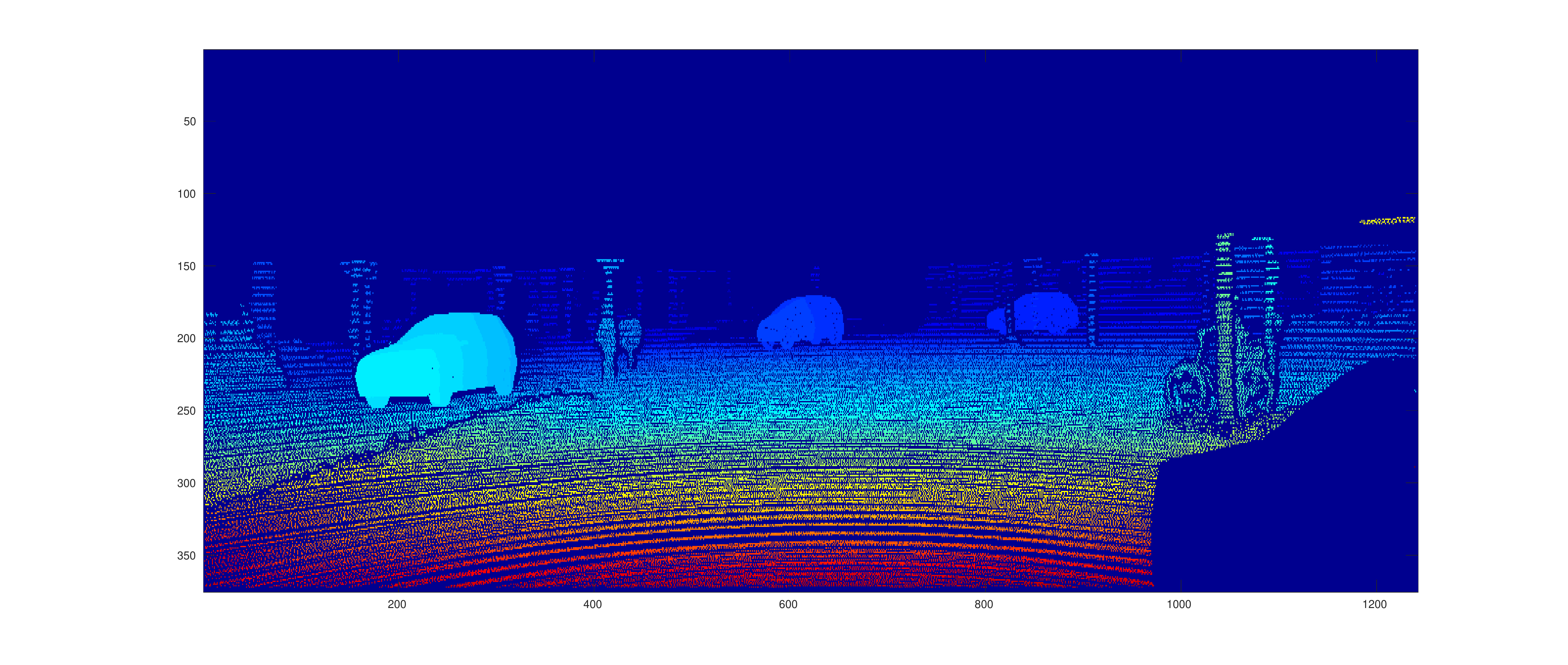}} 
		\subfloat[]{\includegraphics[width=0.5\textwidth]{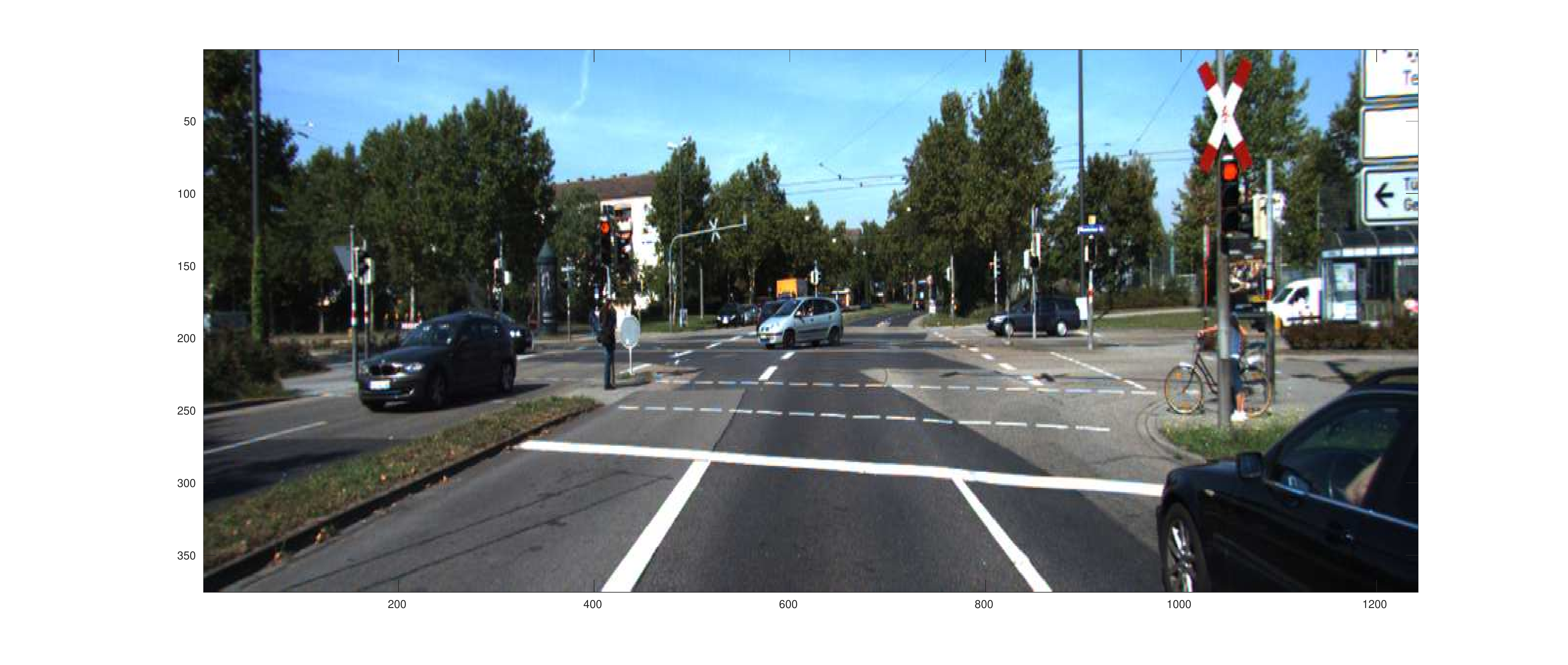}}\\
		\subfloat[]{\includegraphics[width=0.5\textwidth]{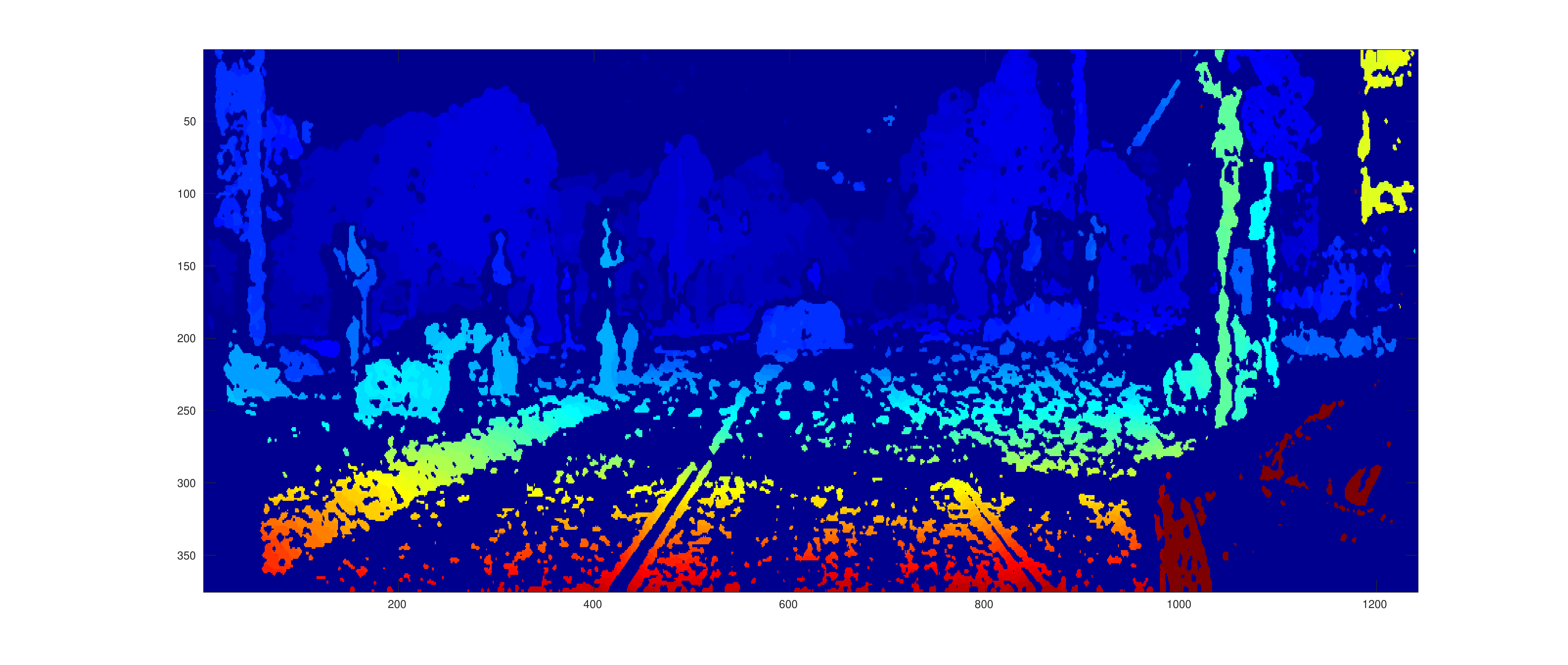}} 
		\subfloat[]{\includegraphics[width =0.5\textwidth]{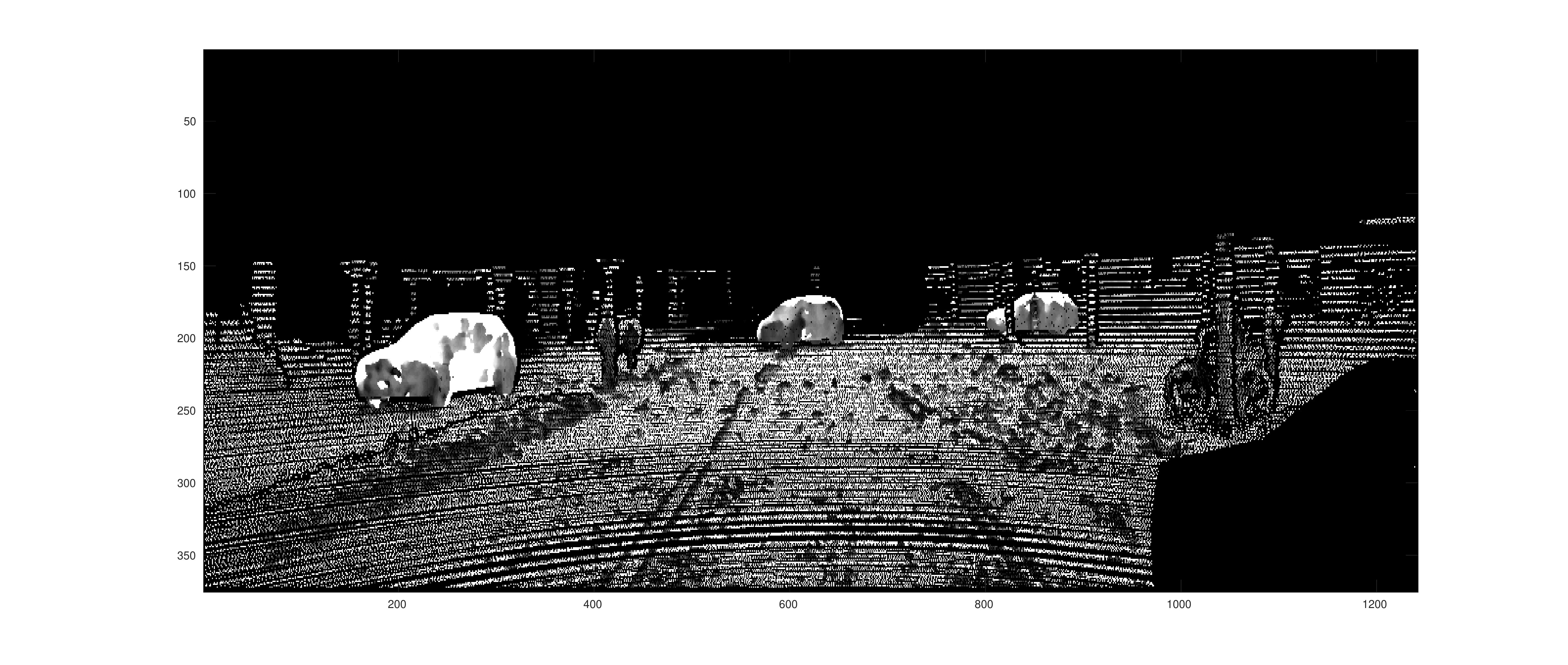}}

		\subfloat[]{\includegraphics[width =0.5\textwidth]{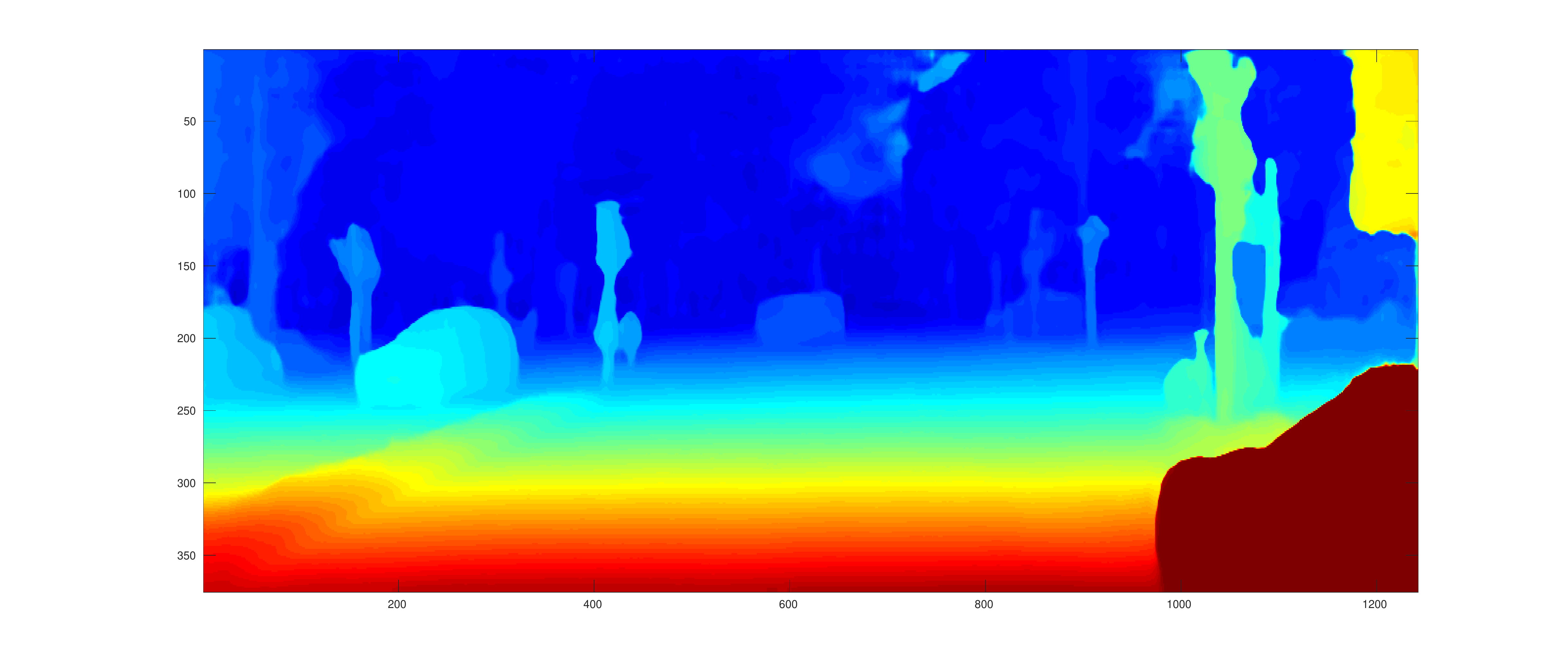}} 
		\subfloat[]{\includegraphics[width =0.5\textwidth]{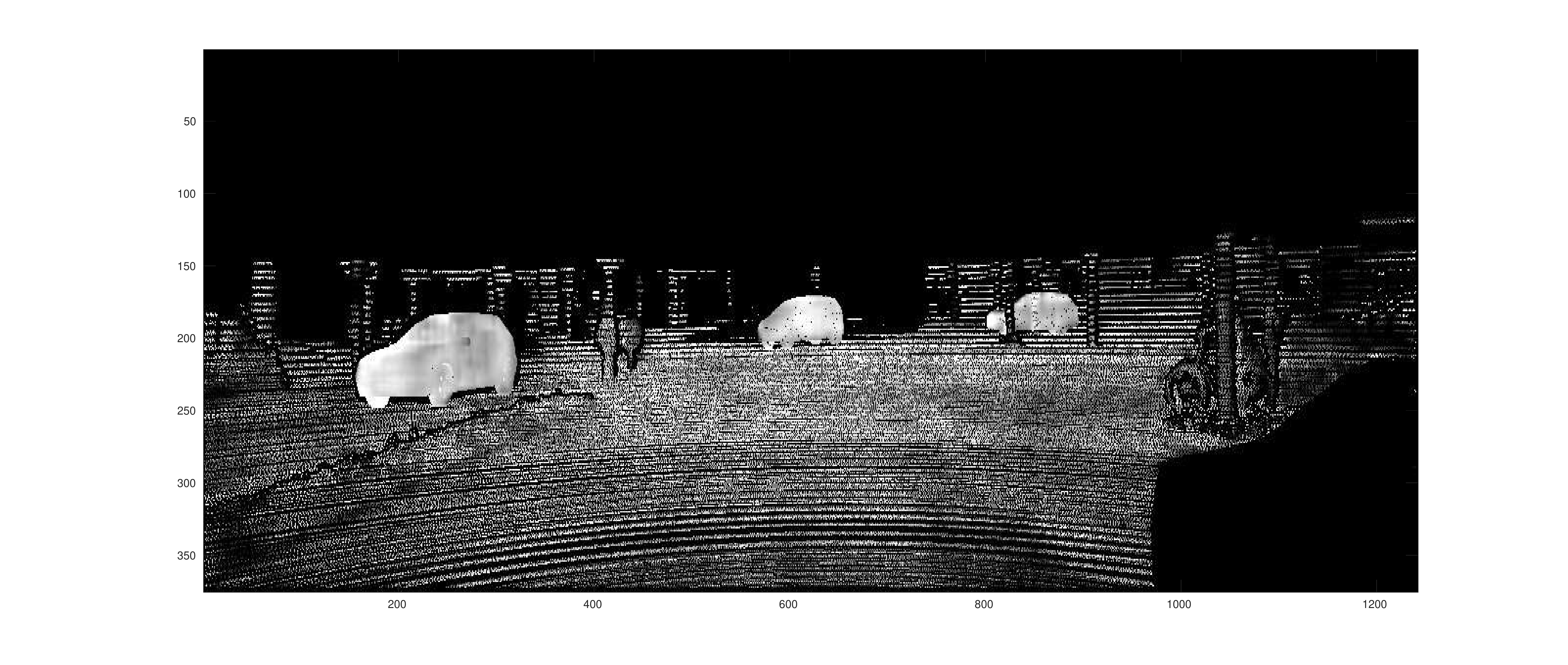}} \\
		\subfloat[]{\includegraphics[width =0.5\textwidth]{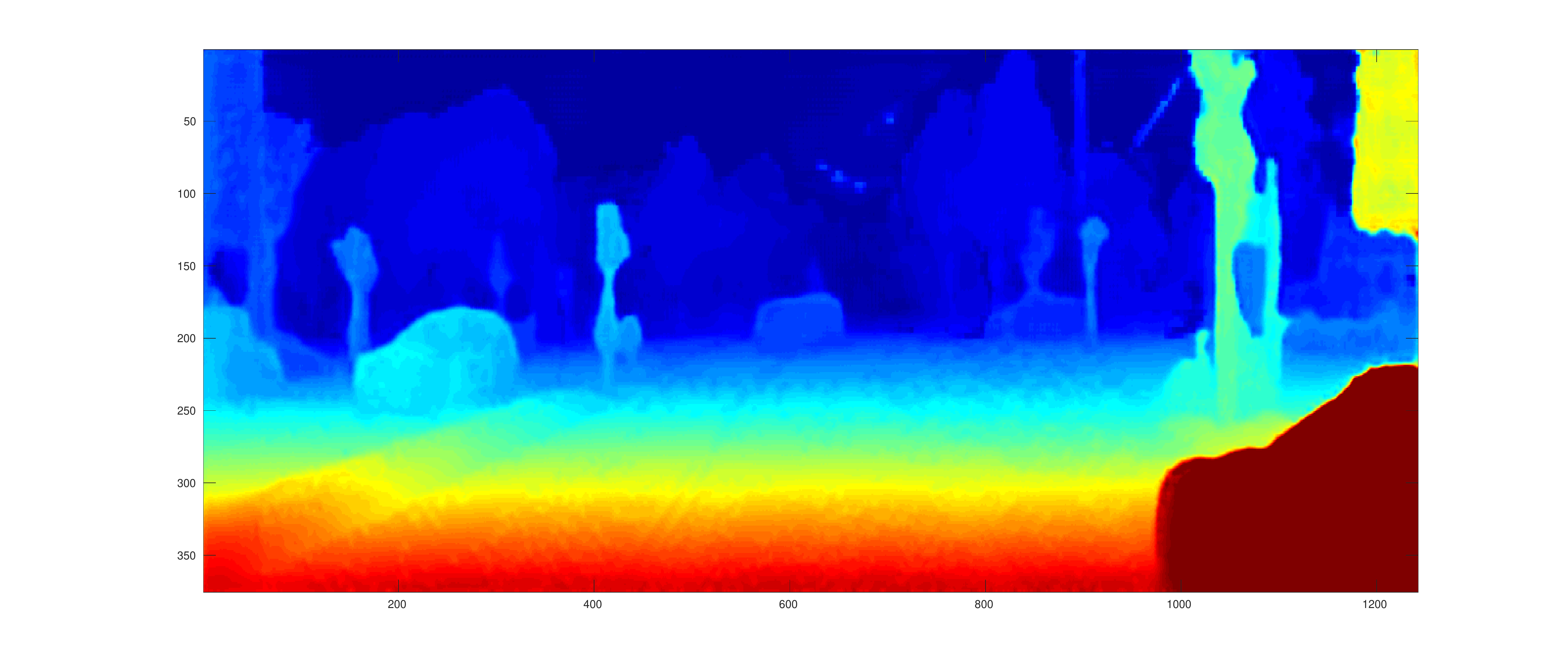}} 
		\subfloat[]{\includegraphics[width =0.5\textwidth]{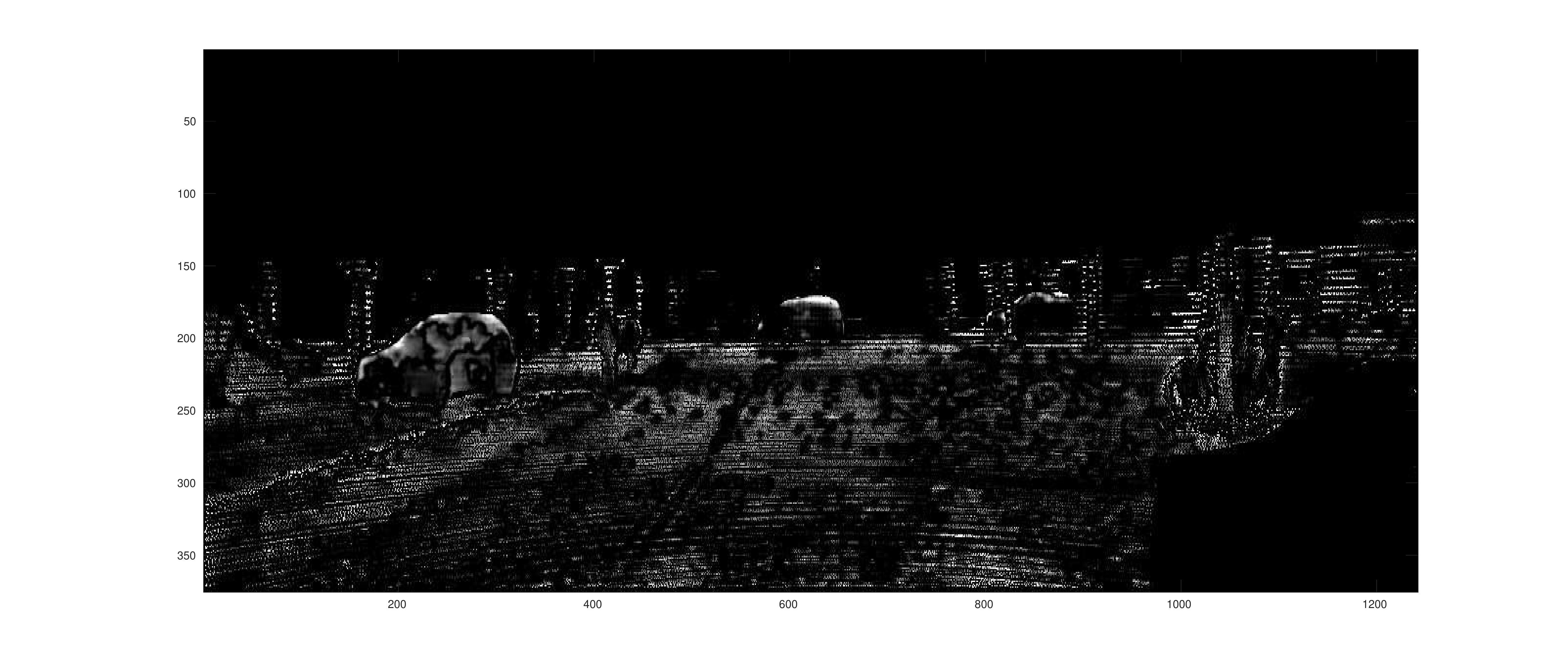}} \\ 
		\caption{We trained our network to fuse the initial disparity maps (\textbf{c},\textbf{e}) into a refined disparity map (\textbf{g}) for the same scene (\textbf{b}) from the Kitti2015 dataset~\cite{Kitti2015} using our supervised method. (\textbf{a}) is the corresponding ground truth. (\textbf{d},\textbf{f},\textbf{h}) are the errors of (\textbf{c},\textbf{e},\textbf{g}). The~lighter pixels have bigger disparity error in (\textbf{d},\textbf{f},\textbf{h}). (\textbf{a})~Ground Truth; (\textbf{b})~Scene; (\textbf{c})~Input Disparity 1: SGM~\cite{SGM}; (\textbf{d})~Input Disparity 1 Error: SGM~\cite{SGM}; (\textbf{e})~Input Disparity 2: PSMNet~\cite{PSMNet}; (\textbf{f})~Input Disparity 2 Error: PSMNet~\cite{PSMNet}; (\textbf{g})~Refined Disparity; (\textbf{h})~Refined Disparity Error. }
		\label{fig:example_demo}
	\end{figure}

	\paragraph{\bf Performance on Trimbot2020 Garden Dataset}
	We tested the proposed network on the real Trimbot2020 Garden dataset, which used a Leica ScanStation P15 
	to capture a dense 3D Lidar point cloud of the whole real garden and then project it to each camera view to get the dense ground truth disparity maps. A~480~$\times$~752 resolution stereo camera was used to get stereo image pairs. The~Trimbot2020 Garden dataset contains 1000 labeled samples for
	training and 250 labeled samples for testing. We~trained the network on the synthetic garden dataset first and fine-tuned the network on the real garden dataset. We~used Dispnet~\cite{Dispnet} and FPGA-stereo~\cite{FPGA_stereo} as inputs. The~authors of~\cite{Dispnet,FPGA_stereo} helped us get the best performance on the real Trimbot2020 Garden dataset as the input to the network. The~settings of the network are shown in ``Stereo-stereo Fusion with Real Trimbot Garden Dataset'' in Table~\ref{Network_setting}. The~demo in the real outdoors garden is available from \url{https://youtu.be/2yyoXSwCSeM}. 
	
	The relevant error of each algorithm on valid pixels is shown in Table~\ref{Trimbot}. The~supervised method and the semi-supervised method have achieved similar top performances compared with the rest. 
	A~qualitative result comparison can be seen in Figure~\ref{fig:Trimbot_Input}.
	The proposed network can deal with the input (resolution: 480~$\times$~768) at 125 fps, which is faster than real-time.
	\begin{table}[h!]
		\centering
		\caption{Mean absolute disparity error of stereo-stereo fusion on Trimbot Garden Dataset (270 test~samples).} \label{Trimbot}
		\begin{tabular}{cccccc}
			\toprule & \multicolumn{2}{c}{\bf Inputs } & \multicolumn{1}{c}{\bf Comparison } & \multicolumn{2}{c}{\bf Our Fused } \\ 
			\midrule
			\textbf{Training Data} & \textbf{FPGA Stereo~\cite{FPGA_stereo}} & \textbf{Dispnet~\cite{Dispnet}} & \textbf{DSF~\cite{Deep_stere_fusion}} & \textbf{Supervised} & \textbf{Semi}\\
			\midrule 
			Num=1000 & 2.94 px & 1.35 px & 0.83 px & 0.67 px &  0.66 px \\ 
			\bottomrule 
		\end{tabular} 
	\end{table} 
	\unskip

	\begin{figure}[h!]
		\centering	
		\setcounter{subfigure}{0}
		\subfloat[]{\includegraphics[width = 6.5cm,height=4cm]{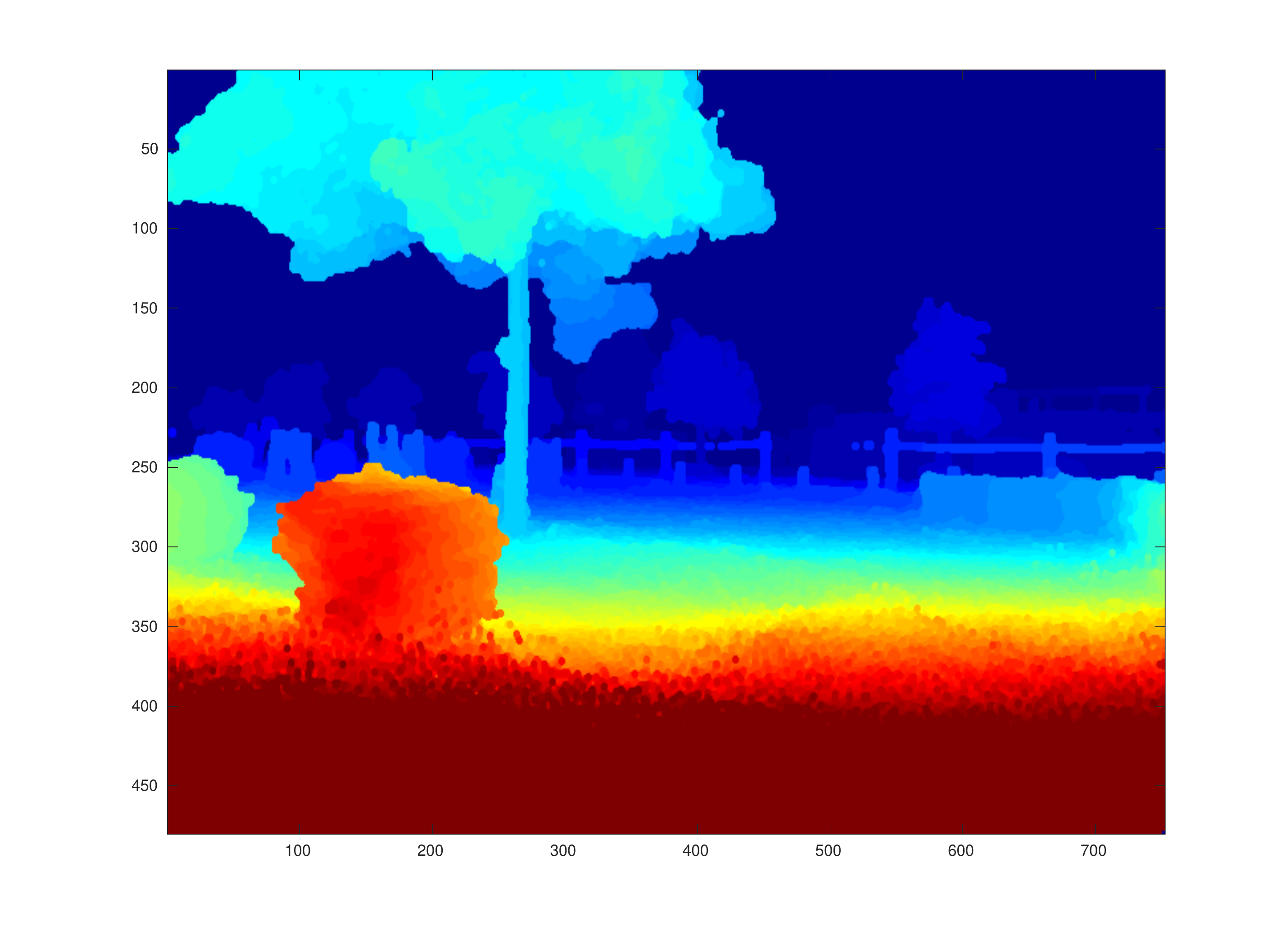}}
		\vspace{0.001cm}
		\hspace{0.1cm}
		\subfloat[]{\includegraphics[width = 6.5cm,height=4cm]{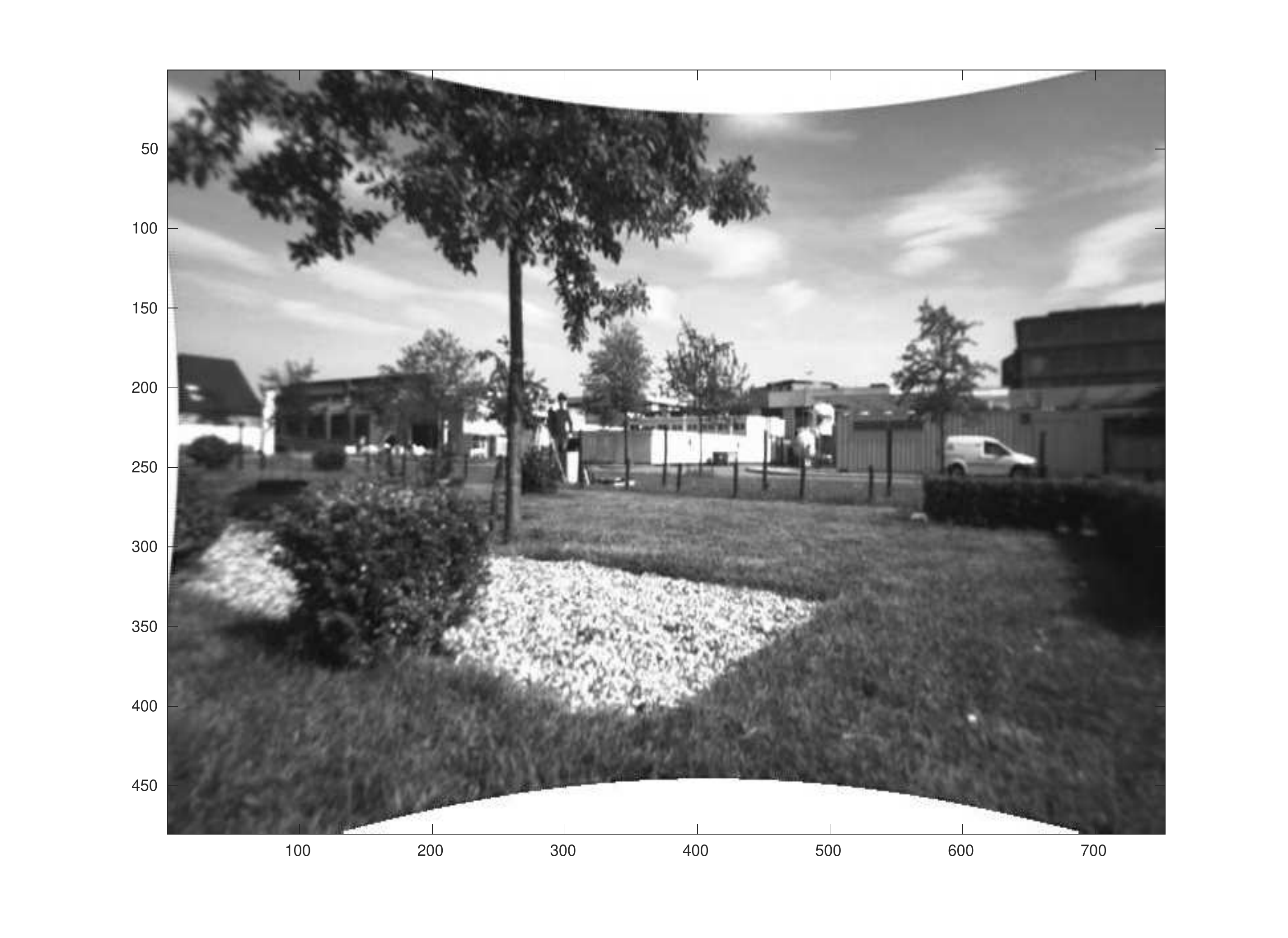}}
		\vspace{0.001cm}
		\subfloat[]{\includegraphics[width = 6.5cm,height=4cm]{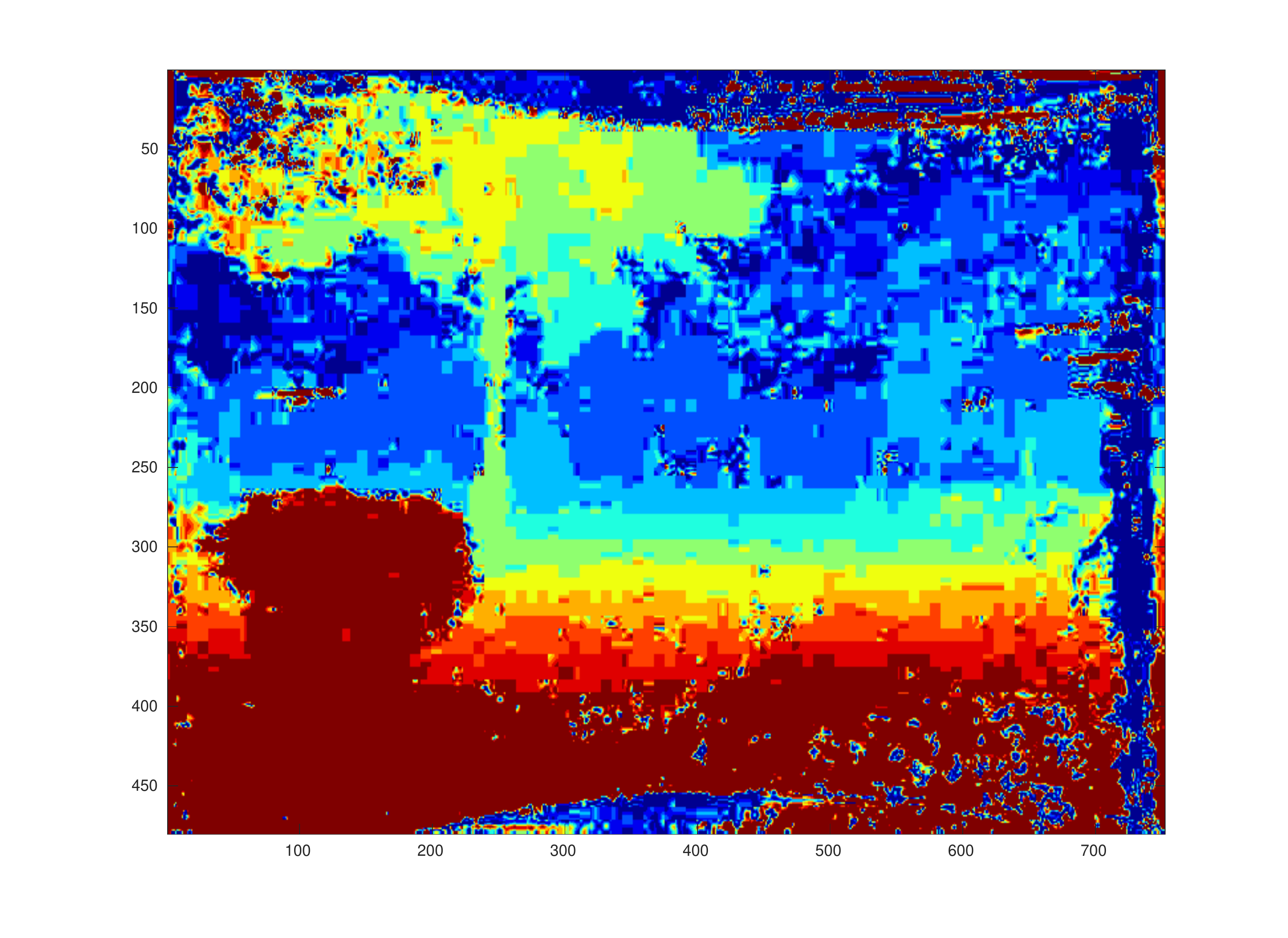}}
		\vspace{0.001cm}
		\hspace{0.1cm} 
		\subfloat[]{\includegraphics[width = 6.5cm,height=4cm]{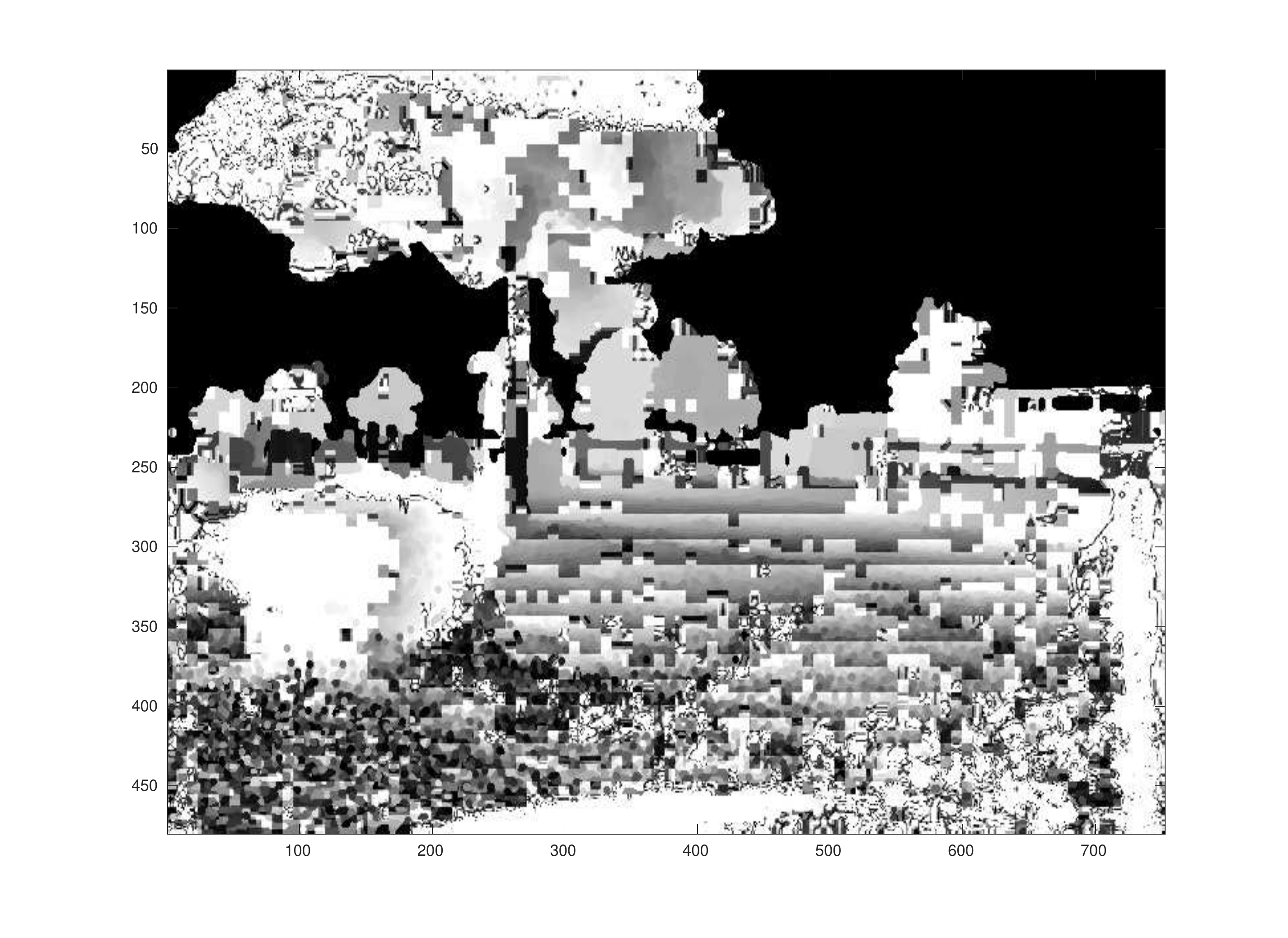}} 
		\vspace{0.001cm}
		\subfloat[]{\includegraphics[width = 6.5cm,height=4cm]{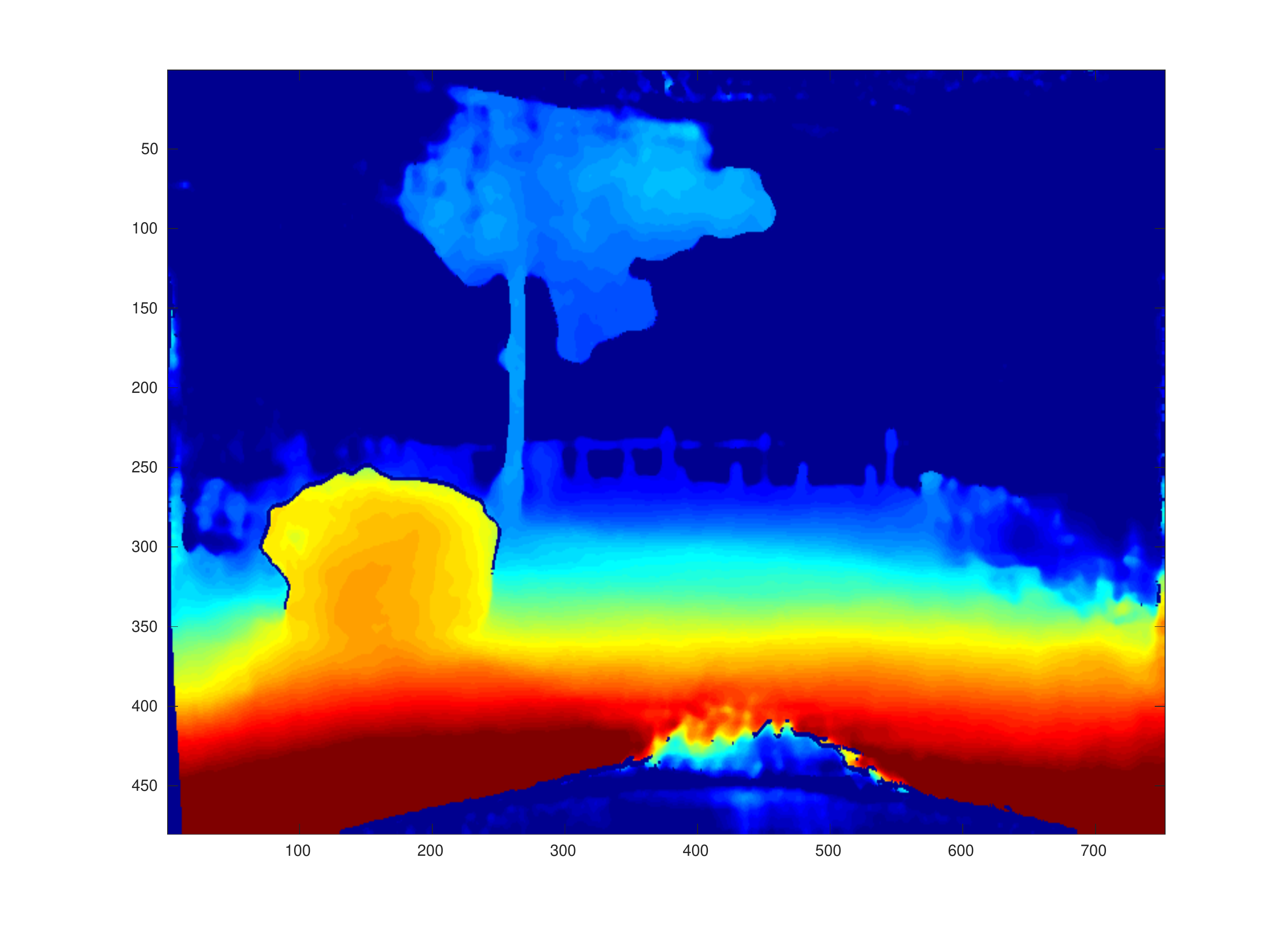}} 
		\vspace{0.001cm}
		\hspace{0.1cm} 
		\subfloat[]{\includegraphics[width = 6.5cm,height=4cm]{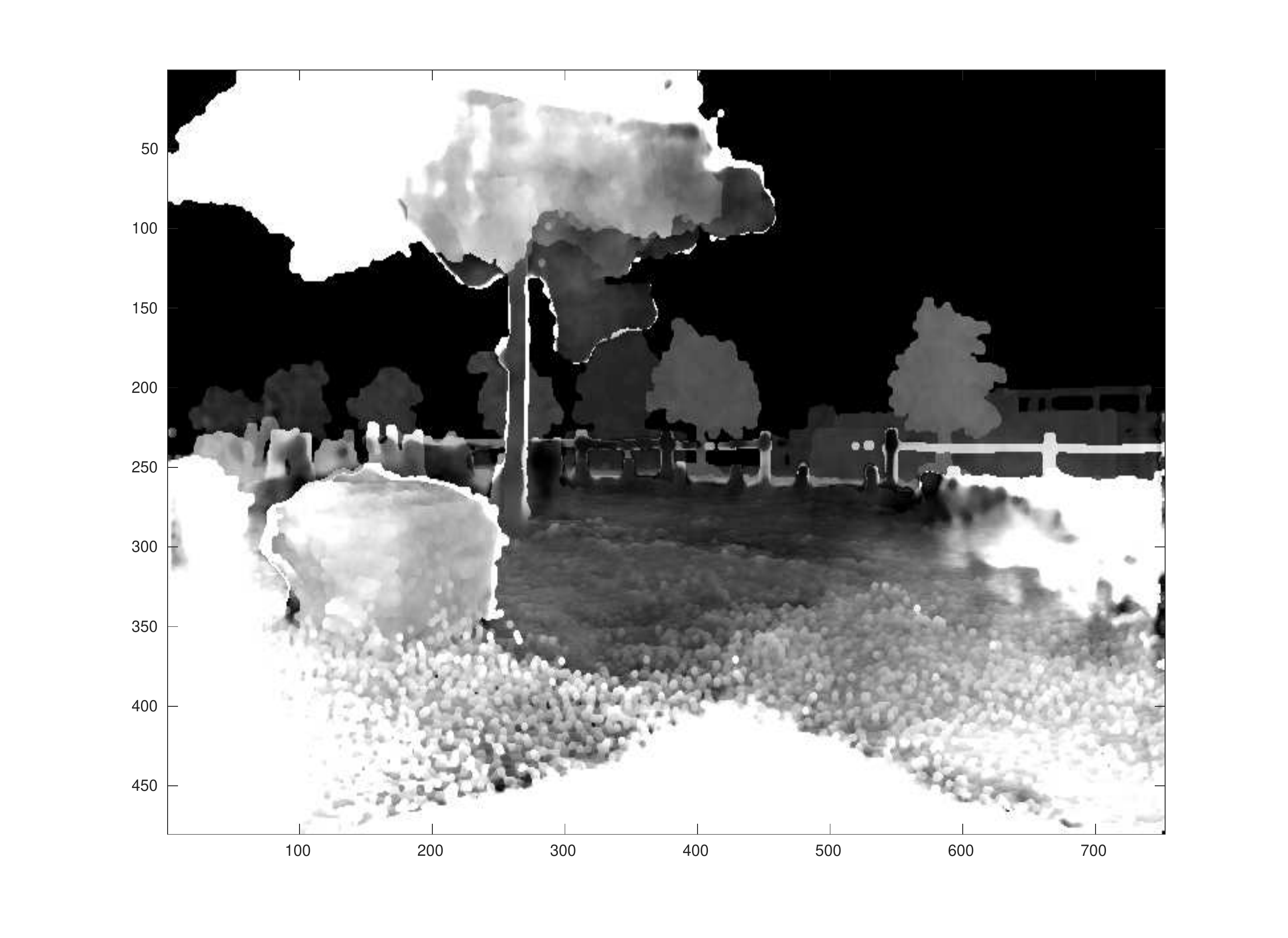}}
		\vspace{0.001cm} 
		\subfloat[]{\includegraphics[width = 6.5cm,height=4cm]{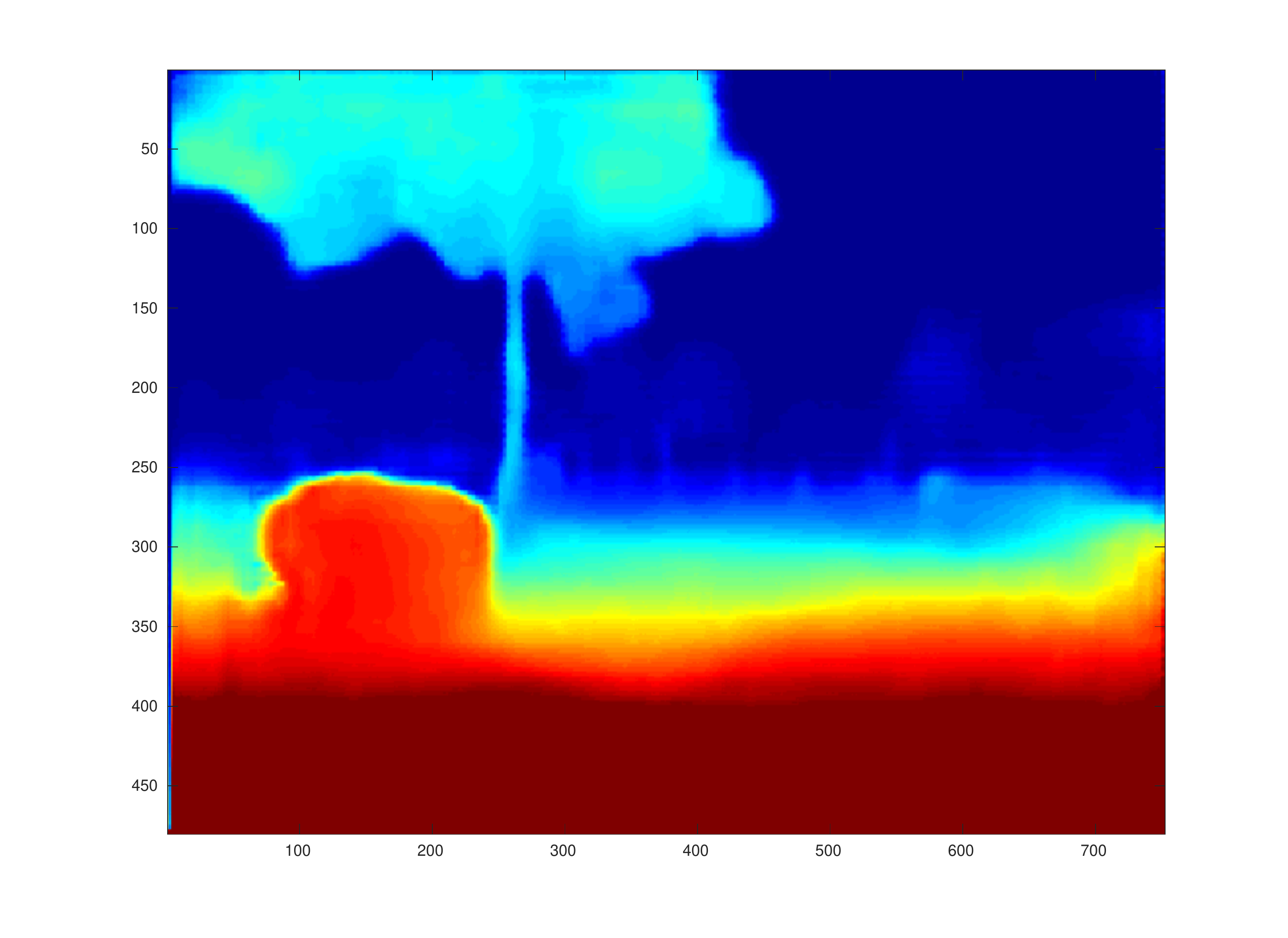}}
		\vspace{0.001cm}
		\hspace{0.1cm} 
		\subfloat[]{\includegraphics[width = 6.5cm,height=4cm]{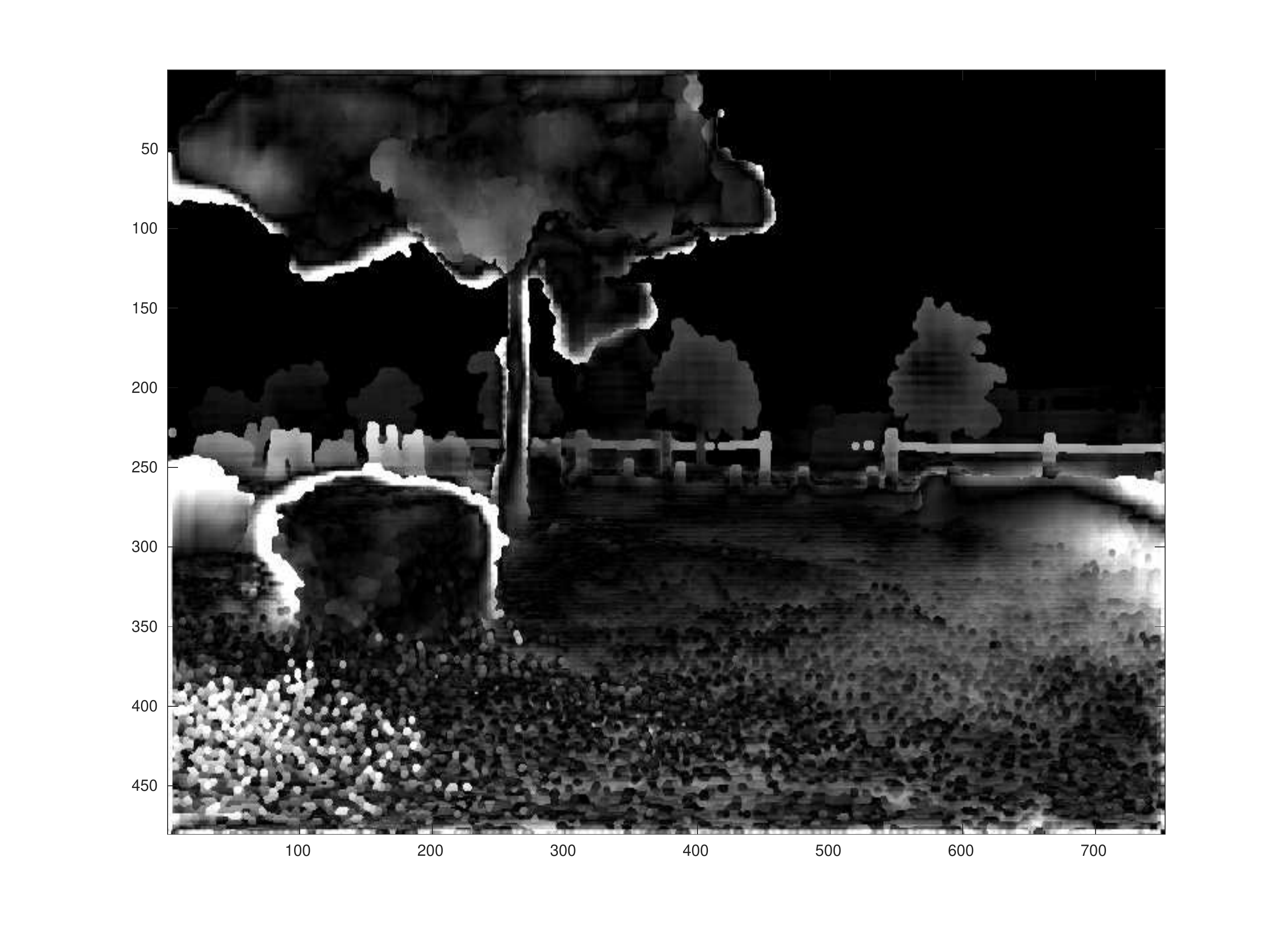}} 
		\vspace{0.001cm}
		\caption{\textit{Cont.}}
	\end{figure}
	
	\begin{figure}[h!]\ContinuedFloat
		\centering
		\subfloat[]{\includegraphics[width = 6.5cm,height=4cm]{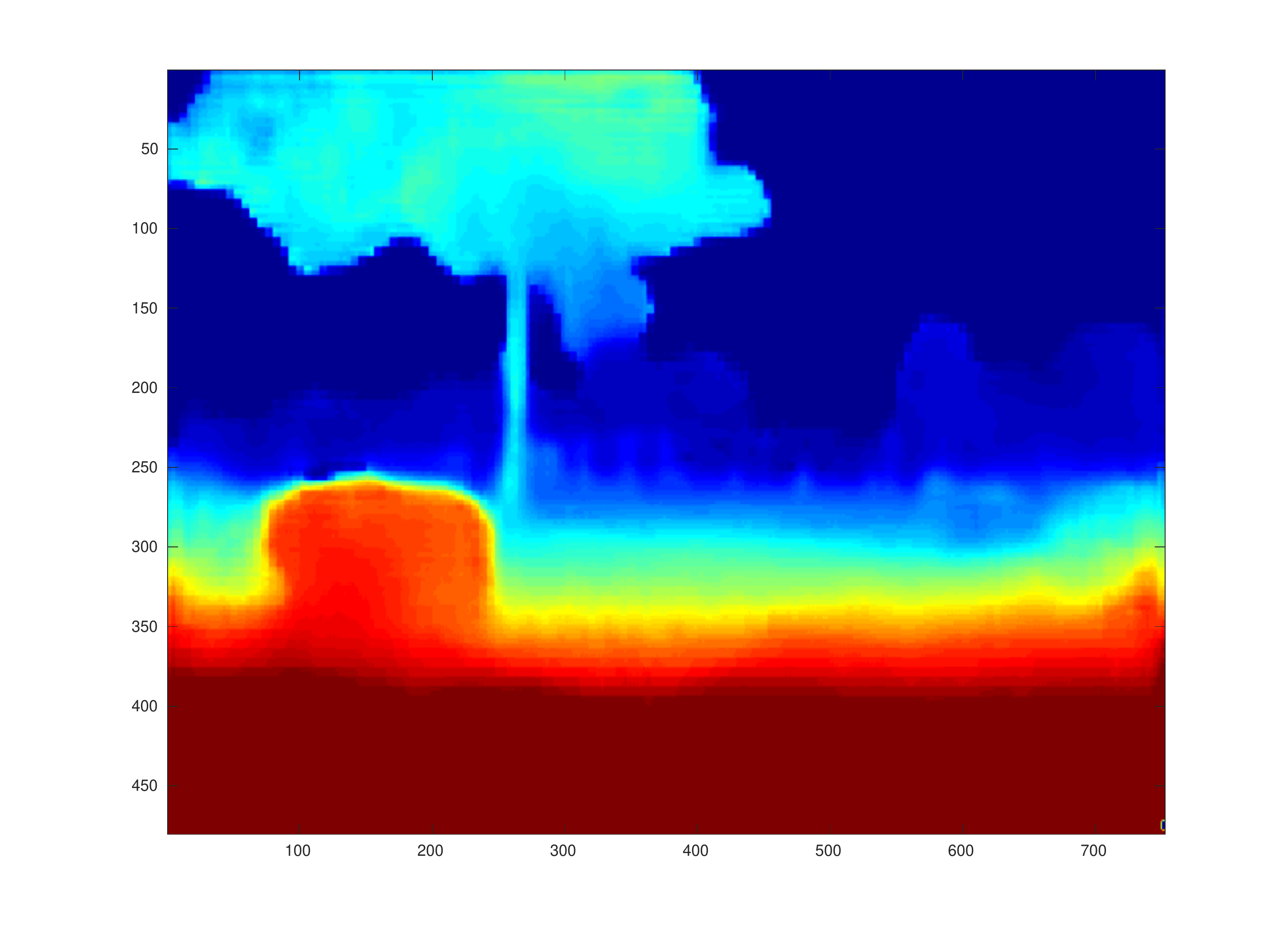}} 
		\vspace{0.001cm}
		\hspace{0.1cm} 
		\subfloat[]{\includegraphics[width = 6.5cm,height=4cm]{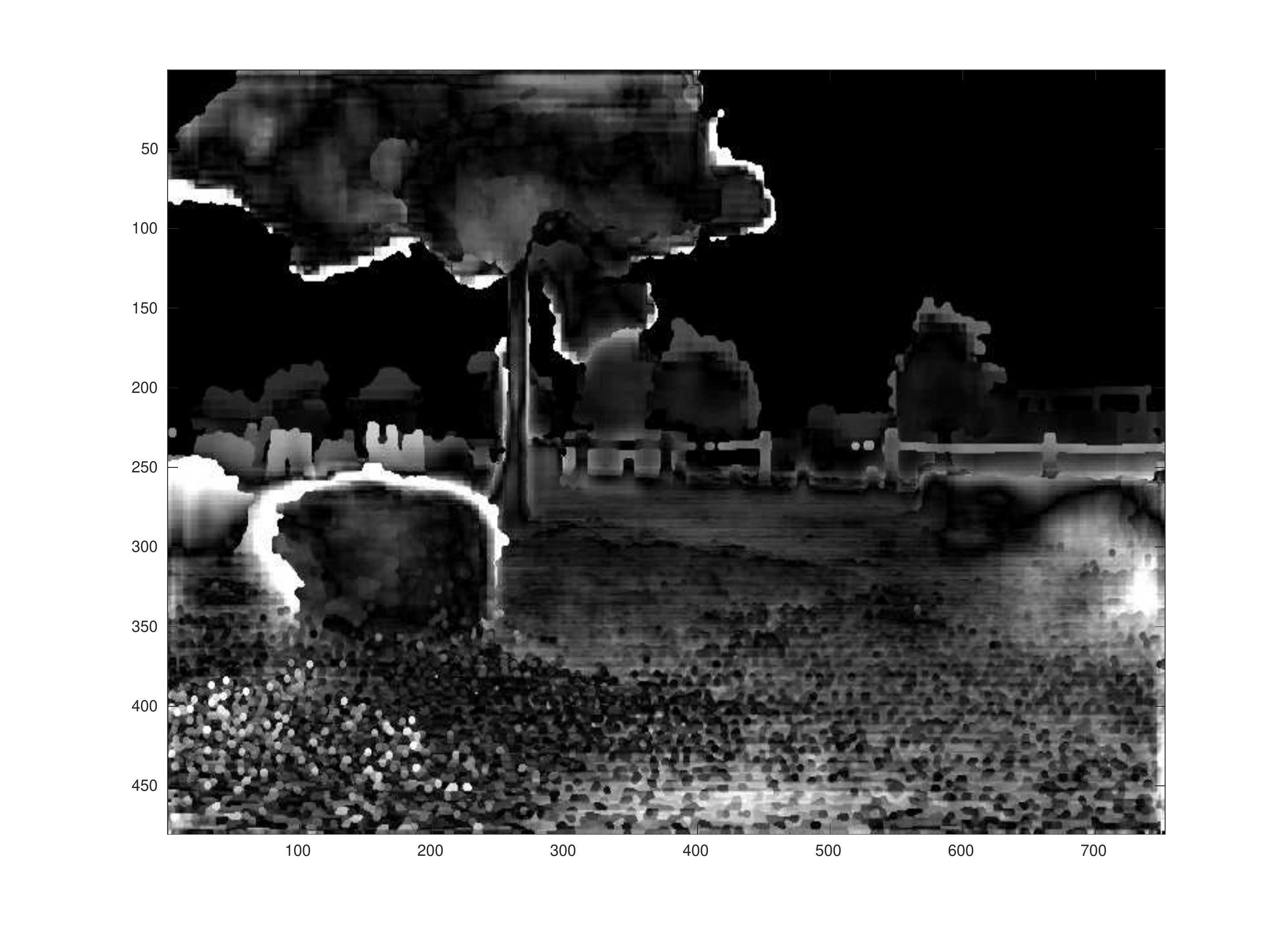}}
		\vspace{0.001cm} 
		\caption{One qualitative result for stereo-stereo fusion in real Trimbot2020 Garden Dataset. The~lighter pixels in (\textbf{d},\textbf{f},\textbf{h},\textbf{j}) represent larger disparity error. (\textbf{a})~ground truth; (\textbf{b})~intensity image; (\textbf{c})~FPGA SGM; (\textbf{d})~FPGA SGM error; (\textbf{e})~DispNet; (\textbf{f})~DispNet error; (\textbf{g})~Supervised 1; (\textbf{h})~error 1 (\textbf{i})~Semi 2; (\textbf{j})~error 2.}
		\label{fig:Trimbot_Input}
	\end{figure}
	\unskip 
	
	{\color{black}
		\subsection{\textcolor{black}{Sensitivity Analysis}}
		All the following experiments are conducted on the Trimbot2020 Garden dataset using the same settings with Performance on Trimbot2020 Garden Dataset 
		in Section~\ref{s4.2.3} except the control variables. The~sensitivity analysis is done for the parameter alpha in Equation~\eqref{eq1}, the~number of scales M in Equation~\eqref{eq5} and Equation~\eqref{eq6}, the~number of feature maps for the refiner network and discriminator network architectures $lg=ld=L$, and~also the parameter momentum in the optimization algorithm Adam. 
		
		\subsubsection{Alpha}
		Table~\ref{Alpha} (corresponding to Figure~\ref{fig:alpha}) shows the performance change when alpha varies from 0.5 to 1.5 with an interval 0.25. Figure~\ref{fig:alpha} shows the robustness of the proposed algorithm. When~alpha~=~1, it~achieves its best performance.
		
		\begin{table}[h!]
			\centering
			\caption{Sensitivity Analysis (Alpha).} \label{Alpha}
			\begin{tabular}{cccccc}
				\toprule
				\textbf{Alpha} & \textbf{0.5} & \textbf{0.75} & \textbf{1} & \textbf{1.25} & \textbf{1.5}\\ 
				\midrule
				Supervised & 0.75 & 0.69 & 0.67 & 0.86 & 0.72\\
				Semi & 0.71 & 0.69 & 0.66 & 0.74 & 0.85\\
				\bottomrule
			\end{tabular} 
		\end{table} 
		\unskip 
		
		\begin{figure}[h!]
			\centering
			\includegraphics[width = 9cm]{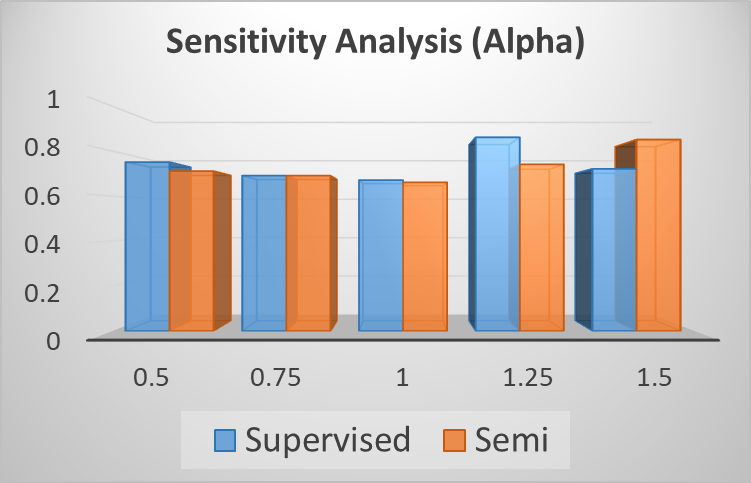} 
			\caption{Sensitivity Analysis (Alpha).}
			\label{fig:alpha}
		\end{figure}
		
		\subsubsection{The Number of Scales}
		Table~\ref{Scales} (corresponding to Figure~\ref{fig:M}) shows the performance change when the number of scales M varies from 1 to 5 with an interval 1. Figure~\ref{fig:M} shows that with the increment of the number of scales, the~error decrease gradually. Therefore~we chose M~=~5.
		
		\begin{table}[h!]
			\centering
			\caption{Sensitivity Analysis (M).} \label{Scales}
			\begin{tabular}{cccccc}
				\toprule
				\textbf{M} & \textbf{1} & \textbf{2} & \textbf{3} & \textbf{4} & \textbf{5}\\
				\midrule
				Supervised & 0.87 & 0.81 & 0.74 & 0.69 & 0.67\\
				Semi & 0.80 & 0.80 & 0.79 & 0.74 & 0.66\\
				\bottomrule 
			\end{tabular} 
		\end{table} 
		\unskip 
		
		\begin{figure}[h!]
			\centering
			\includegraphics[width = 9cm]{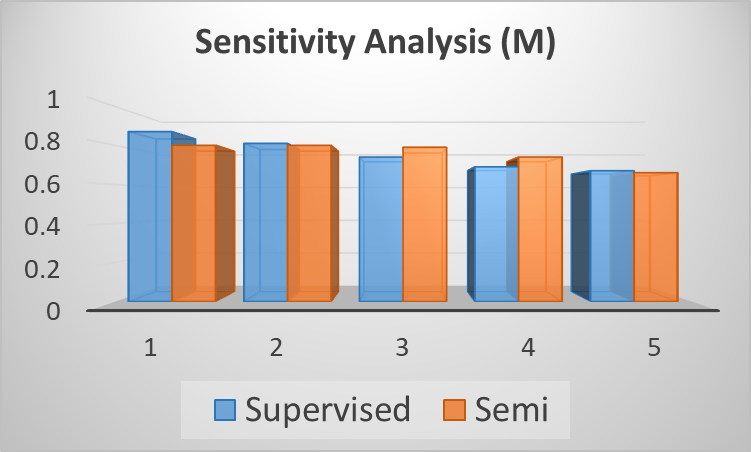} 
			\caption{Sensitivity Analysis (M).}
			\label{fig:M}
		\end{figure}
		\unskip 
		
		\subsubsection{The Number of Feature Maps }
		Table~\ref{Feature_maps} (corresponding to Figure~\ref{fig:L}) shows the performance change when $L$ 
		varies from 6 to 18 with an interval 3. Figure~\ref{fig:L} shows that with the increment of the number of feature maps’ channels, the~overall performance does not change too much but when L~=~12 it performs best.
		
		\begin{table}[h!]
			\centering
			\caption{Sensitivity Analysis (L).} \label{Feature_maps}
			\begin{tabular}{cccccc}
				\toprule
				\textbf{$L$} & \textbf{6} & \textbf{9} & \textbf{12} & \textbf{15} & \textbf{18}\\
				\midrule 
				Supervised & 0.75 & 0.85 & 0.67 & 0.81 & 0.78\\
				Semi & 0.76 & 0.77 & 0.66 & 0.73 & 0.72\\
				\bottomrule 
			\end{tabular} 
		\end{table} 
		\unskip 
		
		\begin{figure}[h!]
			\centering
			\includegraphics[width = 9cm]{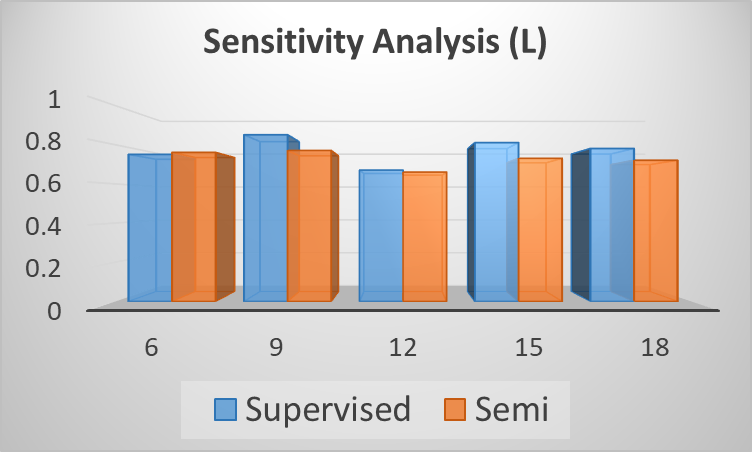} 
			\caption{Sensitivity Analysis (L).}
			\label{fig:L}
		\end{figure} 
		
		\subsubsection{Momentum}
		As for the momentum in the Adam optimization algorithm, different momentum values are used to redo the experiments again. The~experimental results are shown in Table~\ref{Momentum} and Figure~\ref{fig:Momentum}. Table~\ref{Momentum} (corresponding to Figure~\ref{fig:Momentum}) shows the performance change when momentum varies from 0.1 to 0.9 with an interval 0.1. Figure~\ref{fig:Momentum} shows that when momentum is bigger than 0.5, it~achieves better performance compared with below 0.5. When~it is equal to 0.5, both the supervised and semi-supervised methods achieve the best performance simultaneously. 
		
		\begin{table}[h!]
			\centering
			\caption{Sensitivity Analysis (Momentum).} \label{Momentum}
			\begin{tabular}{cccccc}
				\toprule
				\textbf{Momentum} & \textbf{0.1} & \textbf{0.3} & \textbf{0.5} & \textbf{0.7} & \textbf{0.9}\\
				\midrule 
				Supervised & 0.81 & 0.83 & 0.67 & 0.76 & 0.67\\
				Semi & 0.87 & 0.90 & 0.66 & 0.66 & 0.70\\
				\bottomrule
			\end{tabular} 
		\end{table} 
		\unskip 
		
		\begin{figure}[h!]
			\centering
			\includegraphics[width = 9cm]{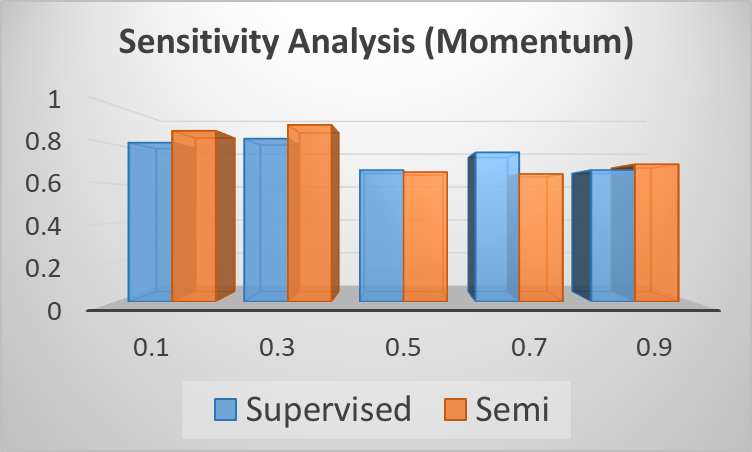} 
			\caption{Sensitivity Analysis (Momentum).}
			\label{fig:Momentum}
		\end{figure} 
		\unskip 
		
		\subsubsection{Statistical Analysis}
		In the paper, for~the experimental results above, we~have done them once, because of the costs of deep net retraining. To~show the robustness and accuracy of the proposed method, we~repeated the experiments on the real Trimbot2020 datasets five times using the same settings. The~corresponding results are shown in Table~\ref{Repeated_experiment}. As~can be seen, the~proposed algorithms are significantly better than the DSF algorithm, however it is less clear if there is a significant difference between the supervised and semi-supervised performances.
		
		\begin{table}[h!]
			\centering
			\caption{\textcolor{black}{Statistical Analysis}.} \label{Repeated_experiment}
			\begin{tabular}{cccccccc}
				\toprule & \multicolumn{5}{c}{\bf Repeated Experiment } & \multicolumn{2}{c}{\bf Statical Result } \\ \midrule
				\textbf{Experiment} & \textbf{1} & \textbf{2} & \textbf{3} & \textbf{4} & \textbf{5 }& \textbf{Mean} & \textbf{Std.}\\
				\midrule 
				DSF & 0.83 & 0.89 & 0.86 & 0.87 & 0.85 & 0.86 & 0.02\\ 
				Supervised & 0.73 & 0.77 & 0.67 & 0.70 & 0.72 & 0.72 & 0.04\\
				Semi & 0.67 & 0.71 & 0.66 & 0.76 & 0.71 & 0.70 & 0.04\\
				\bottomrule 
			\end{tabular} 
	\end{table}}
	\vspace{-12pt}

	\section{Conclusions and Discussion \label{s5}}
	The paper has presented a method to refine disparity maps based on fusing the results from multiple disparity calculation algorithms and other supplementary image information (e.g., intensity, gradient). The~proposed method can generalize to perform different fusion tasks and achieves better accuracy compared with several recent fusion algorithms. It~could potentially fuse multiple algorithms (not only 2 algorithms as shown in this paper) by concatenating more initial disparity maps in the network's input but this has not been explored. The~objective function and network architecture are novel and effective. In~addition, the~proposed semi-supervised method greatly reduces the amount of ground truth training data needed, while achieving comparable performance with the proposed supervised method. The~proposed semi-supervised method can achieve better performance when using the same amount of labeled data as the supervised method plus the additional unlabeled data. \textcolor{black}{In the future, we~plan to explore unsupervised disparity fusion with adversarial neural networks using left-right intensity consistency between the two stereo vision cameras. Meanwhile,~future exploration on disparity fusion in object space (e.g.,~\cite{Reconstruct_past}) is considered. It~will be interesting to compare the disparity fusion in image space versus object space. Additionally,~more datasets will be used to explore the generalization of the proposed method, such as using remote sensing datasets that are acquired by Satellite or UAV sensors.}

\section{Acknowledgement}
The research is fully funded by the TrimBot2020 project [Grant Agreement No. 688007, URL: \url{http://trimbot2020.webhosting.rug.nl/}] from the European Union Horizon 2020 programme. We thank Chengyang Zhao, Marija Jegorova and Timothy Hospedales for giving us good advice on this paper. We thank our partners (the authors of ~\cite{Dispnet,FPGA_stereo}) for helping us get their best performance as the input to the network.

\appendix
\section{Description of Trimbot2020 Garden Dataset \label{TBDataset}}
\unskip


We make use of the Trimbot Garden 2017 dataset used for the semantic reconstruction challenge of the ICCV 2017 workshop `3D Reconstruction meets Semantics'~\cite{Sattler2017rms}. 
The dataset consists of a 3D laser scan of the garden as well as multiple traversals of the robot through the garden (see Figure~\ref{fig:eval-gt-dataset}). In~addition to the challenge dataset (2 camera pairs), we~included all 5 camera pairs (Figure~\ref{fig:cam-setup}) , obtaining total 1250 sample pairs.
Robot poses for the traversals were recorded in the coordinate system of the laser scanner using a Topcon laser tracker. 
The results were subsequently refined using Structure-from-Motion~\cite{Schoenberger2016CVPR}. 
The quantitative evaluation is performed only on a subset of pixels which correspond to static non-ground areas (the grass on the ground surface yields noisy GT measurements as well as other moving parts like tree branches).

The accuracy of stereo depth map estimates depends on the distance of the cameras to the scene, with the uncertainty growing quadratically with the distance. 
In contrast, the~uncertainty grows only linearly in the disparity space (measured in pixels). 
As is common~\cite{Geiger2012CVPR,Schoeps2017CVPR}, we~thus measured the accuracy of the stereo algorithms by comparing their estimated disparity values with the ground truth disparity values provided by the laser scanner.

\begin{figure}[h!] \centering
	\includegraphics[width=\textwidth]{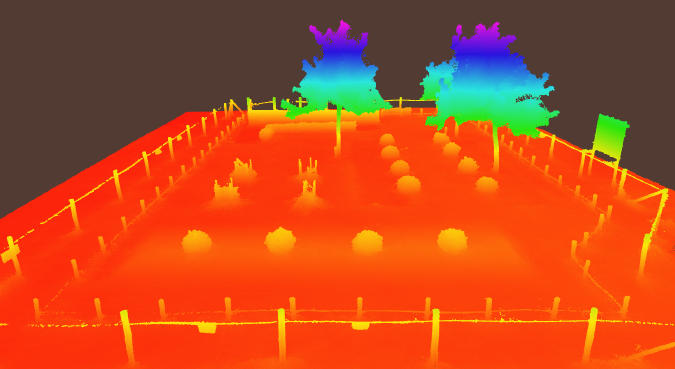}
	\vspace{3mm}
	\includegraphics[width=\textwidth]{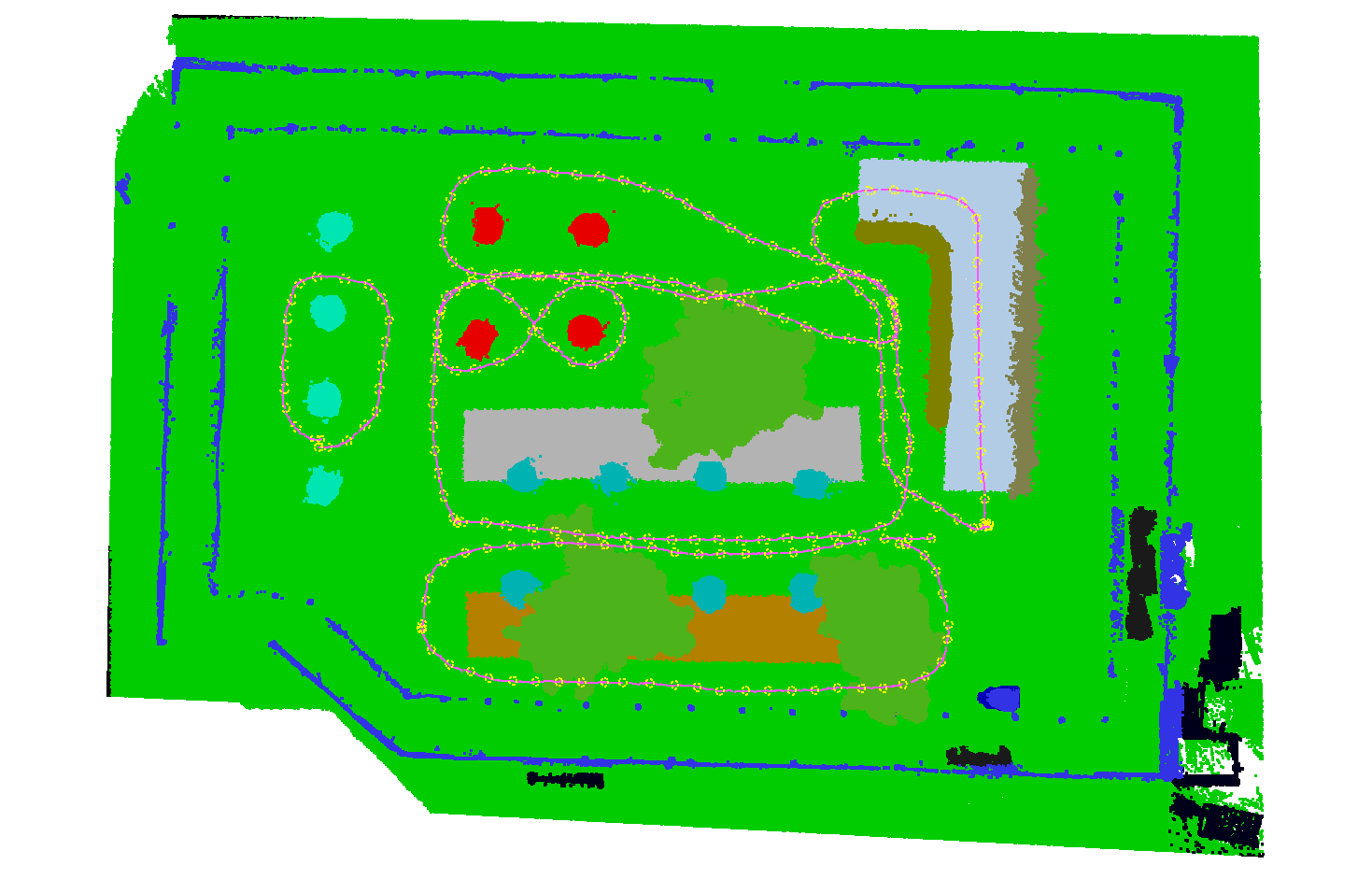}
	\caption{Trimbot Garden 2017 GT dataset~\cite{Sattler2017rms}. {\textbf{Above}:}~point cloud with color-encoded height. {\textbf{Below}:}~semantic point cloud with trajectories (magenta line) and camera centers (yellow). } 
	\label{fig:eval-gt-dataset}
\end{figure} 
\unskip 

\begin{figure}[h!] 
	\centering
	\includegraphics[width=0.45\textwidth]{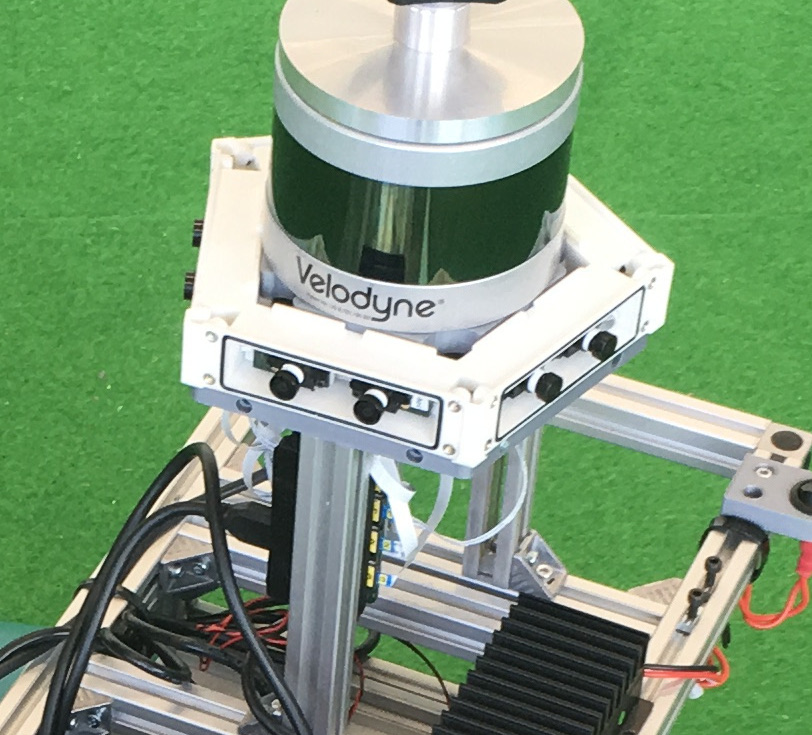}\hspace{5mm}\includegraphics[width=0.45\textwidth]{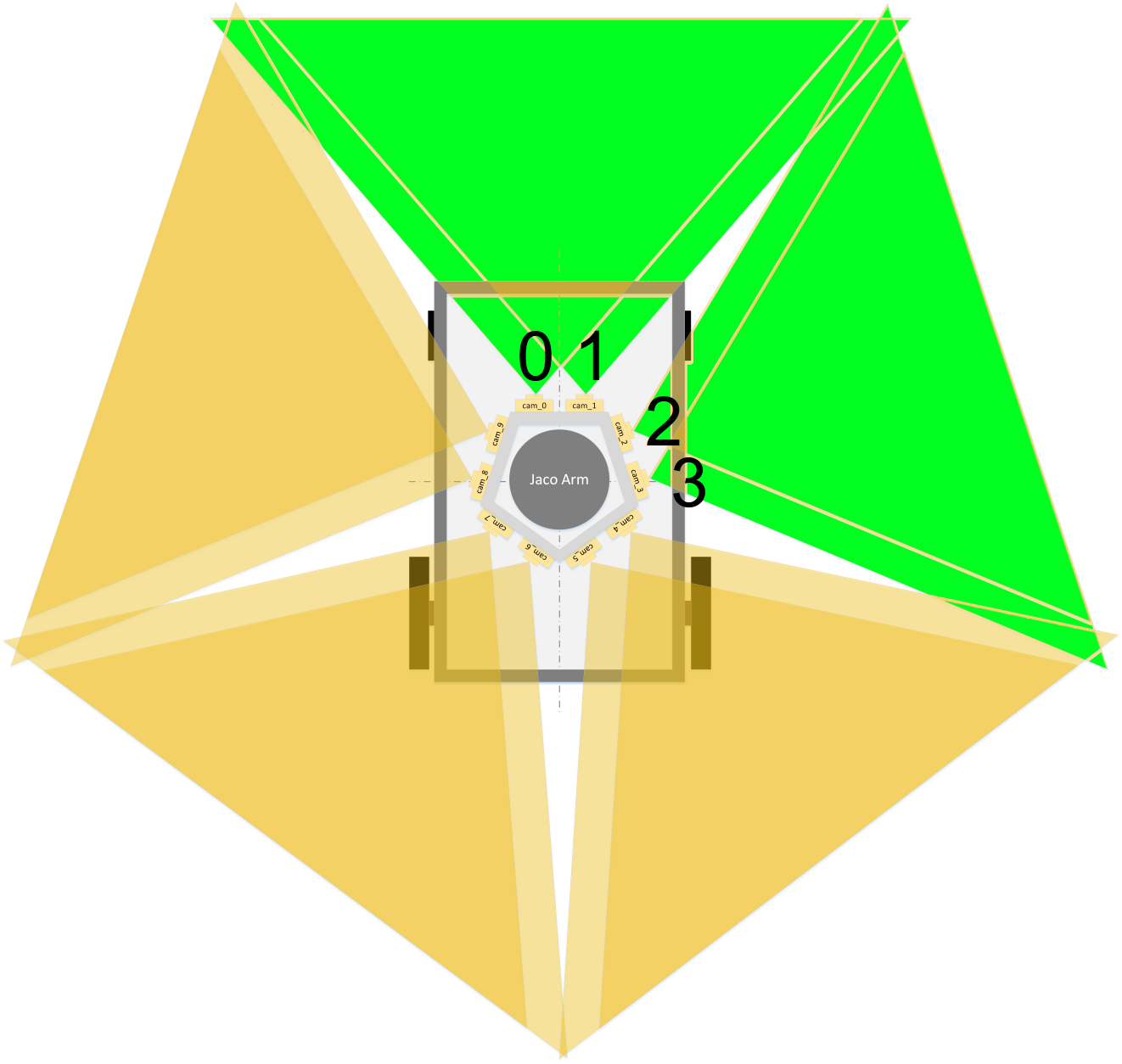} 
	\par
	\caption{Trimbot Garden 2017 GT dataset~\cite{Sattler2017rms}. {\textbf{Left}:}~Pentagonal camera rig mounted on the robot with five stereo pairs. 
		{\textbf{Right}:}~Top view of camera rig with test set pairs (green field of veiw) and training set pairs (yellow field of view).}
	\label{fig:cam-setup}
\end{figure}

\bibliographystyle{unsrt}  

\end{document}